%% file: main.tex
\documentclass{article}

\usepackage{microtype}
\usepackage{graphicx}
\usepackage{booktabs} %

\usepackage{hyperref}

\usepackage[accepted]{icml2025}

\usepackage{amsmath}
\usepackage{amssymb}
\usepackage{mathtools}
\usepackage{amsthm}

\usepackage[capitalize,noabbrev]{cleveref}

\theoremstyle{plain}

\theoremstyle{definition}

\theoremstyle{remark}

\icmltitlerunning{Scaling Value Iteration Networks to 5000 Layers for Extreme Long-Term Planning}

\input{lib_my/A_ALL.tex}

\input{lib_my/A_CUR_HighwayVIN.tex}

\input{lib_my/A_CUR.tex}

\input{lib_my/math_commands}
\usepackage{wrapfig}
\usepackage{varwidth}

\ifshowtmp
    \newcommand{\dylan}[1]{\textcolor{blue}{(\textbf{Dylan}: #1)}}
    \newcommand{\francesco}[1]{\textcolor{orange}{(\textbf{Francesco:} #1)}}
    \newcommand{\juergen}[1]{\textcolor{blue}{(\textbf{Juergen}: #1)}}
    \newcommand{\qingyuan}[1]{\textcolor{purple}{(\textbf{Qingyuan}: #1)}}
    \newcommand{\weida}[1]{\textcolor{green}{(\textbf{Weida}: #1)}}
    \newcommand{\yuhui}[1]{\textcolor{teal}{(\textbf{Yuhui}: #1)}}
\else
    \newcommand{\dylan}[1]{}
    \newcommand{\francesco}[1]{}
    \newcommand{\juergen}[1]{}
    \newcommand{\qingyuan}[1]{}
    \newcommand{\weida}[1]{}
    \newcommand{\yuhui}[1]{}
\fi

\begin{document}

\twocolumn[
\icmltitle{Scaling Value Iteration Networks to 5000 Layers \\ for Extreme Long-Term Planning}

\icmlsetsymbol{equal}{*}

\begin{icmlauthorlist}
\icmlauthor{Yuhui Wang}{equal,kaust}
\icmlauthor{Qingyuan Wu}{equal,Southampton}
\icmlauthor{Dylan R.\ Ashley}{kaust,idsia}
\icmlauthor{Francesco Faccio}{kaust,idsia}
\icmlauthor{Weida Li}{nus}\\
\icmlauthor{Chao Huang}{Southampton}
\icmlauthor{J\"{u}rgen Schmidhuber}{kaust,idsia}
\end{icmlauthorlist}

\icmlaffiliation{kaust}{Center of Excellence for Generative AI, King Abdullah University of Science and Technology}
\icmlaffiliation{idsia}{The Swiss AI Lab IDSIA/USI/SUPSI, Switzerland}
\icmlaffiliation{Southampton}{The University of Southampton}
\icmlaffiliation{nus}{National University of Singapore}

\icmlcorrespondingauthor{Yuhui Wang}{yuhui.wang@kaust.edu.sa}

\icmlkeywords{Machine Learning, ICML}

\vskip 0.3in
]

\begin{abstract}

The Value Iteration Network (VIN) is an end-to-end differentiable neural network architecture for planning. It exhibits strong generalization to unseen domains by incorporating a differentiable planning module that operates on a latent Markov Decision Process (MDP).
However, VINs struggle to scale to long-term and large-scale planning tasks, such as navigating a $100\times 100$ maze---a task that typically requires thousands of planning steps to solve.
We observe that this deficiency is due to two issues:
the representation capacity of the latent MDP and
the planning module's depth.
We address these by augmenting the latent MDP with a dynamic transition kernel, dramatically improving its representational capacity,
and, to mitigate the vanishing gradient problem, introduce an ``adaptive highway loss'' that constructs skip connections to improve gradient flow.
We evaluate our method on 2D/3D maze navigation environments, continuous control, and the real-world Lunar rover navigation task.
We find that our new method, named \emph{Dynamic Transition VIN (DT-VIN)}, scales to 5000 layers and solves challenging versions of the above tasks.
Altogether, we believe that DT-VIN represents a concrete step forward in performing long-term large-scale planning in complex environments.

\end{abstract}

\printAffiliationsAndNotice{\icmlEqualContribution} %

\section{Introduction}

Planning is the problem of finding a sequence of actions that achieves a specific pre-defined goal.
As the aim of both some older algorithms (e.g., Dyna~\citep{sutton1991dyna}, A*~\citep{hart1968formal}, and others \citep{Schmidhuber:90diffgenau,Schmidhuber:90sandiego}) and many recent ones (e.g., the Predictron~\citep{silver2017predictron}, the Dreamer family of algorithms~\citep{hafner2020dream,hafner2021mastering,hafner2023mastering}, SoRB~\citep{eysenbach2019search}, SA-CADRL~\citep{chen2017socially}, and the LLM-planner~\citep{song2023llm}), effective planning is a long-standing and important challenge in artificial intelligence (AI).

\begin{figure*}
    \def\height{0.232}
    \centering
    \captionsetup[subfloat]{font=tiny}
    \subfloat[\tiny $100\times 100$ Maze]{
        \includegraphics[height=\height\linewidth]{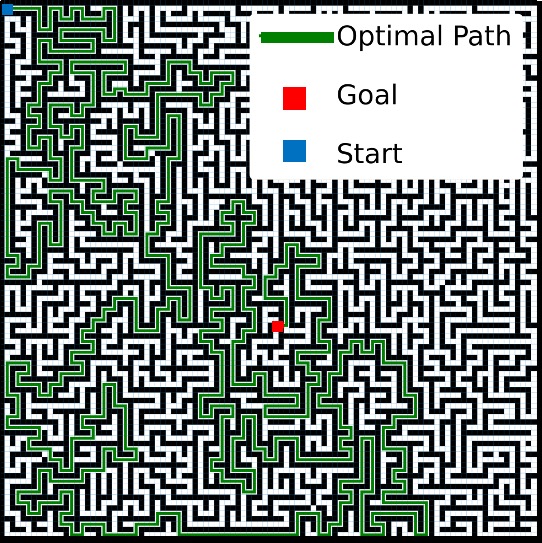}
        \label{fig_maze_100}
    }
    \subfloat[\tiny{Performance by Domain Scales}]{
        \includegraphics[height=\height\linewidth]{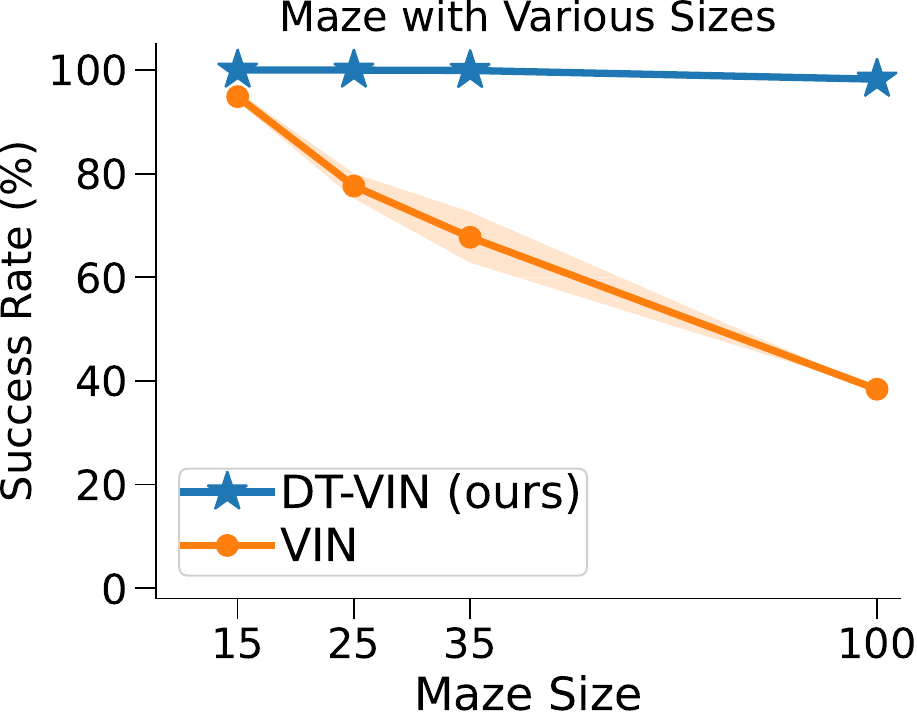}
        \label{fig_performance__vs__doman_scales}
    }
    \subfloat[\tiny{Performance by Planning Steps}]{
        \includegraphics[height=\height\linewidth]{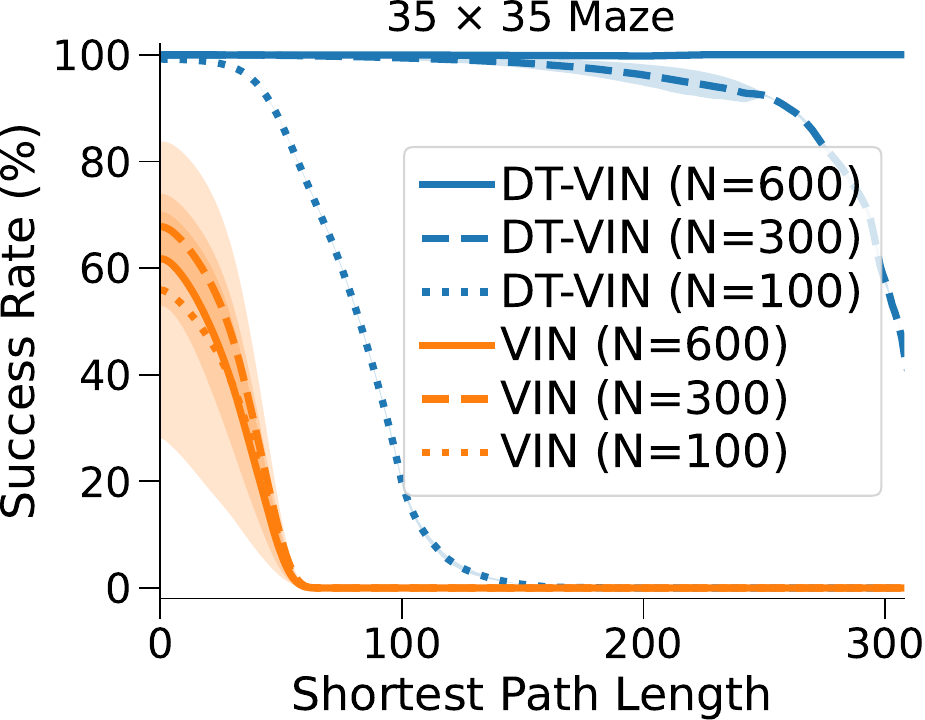}
        \label{fig_performance__vs__planning_steps}
    }
    \caption{
        \subref{fig_maze_100} shows an example of $100\times 100$ maze navigation task, where the green line shows the optimal path from the start position (blue) to the goal position (red).
        See Appendix~\cref{fig_maze_various_size} for more examples of mazes with other sizes.
        \subref{fig_performance__vs__doman_scales} shows the success rate of VIN and DT-VIN on the maze navigation tasks as a function of maze size. The reported results are computed in expectation over different shortest path lengths for each maze size.
        \subref{fig_performance__vs__planning_steps} shows the success rate of VIN~\citep{tamar2016value} and our DT-VIN as a function of planning steps on the $35 \times 35$ maze benchmark.
    }
    \label{fig_introduction}
\end{figure*}

Traditional search-based planning algorithms like A* require an accurate environmental model.
Thus, these algorithms are less effective in scenarios with unknown environmental models or when the state and action spaces are large or continuous.
In such scenarios, a policy can be learned either through imitation learning (IL), which leverages expert demonstrations, or through trial and error with reinforcement learning (RL).
Within RL and IL, the Value Iteration Network (VIN)~\citep{tamar2016value} stands out as quite unique due to its distinctive architecture that integrates a differentiable latent ``planning module'' into the deep neural network, rather than maintaining an explicit learned environment model like Dreamer~\citep{hafner2020dream} or MuZero~\citep{schrittwieser2020mastering}.
This integrated planning structure of VINs endows them with powerful generalization capabilities for unseen planning tasks.
VINs have been shown to perform well in some small-scale short-term planning situations, like path planning~\citep{pflueger2019rover,jin2021value}, autonomous navigation~\citep{wohlke2021hierarchies}, and complex decision-making in dynamic environments~\citep{li2021dynamic}.
However, they still struggle to solve larger-scale and longer-term planning problems.
We refer to \emph{large-scale planning tasks} as those with high-dimensional observation space (e.g., the maze size), and \emph{long-term planning tasks} as those necessitating extended planning horizons to achieve the goal.
For example, in a $100 \times 100$ maze navigation task, the success rate of VINs in reaching the goal drops to well below 40\% (see \cref{fig_performance__vs__doman_scales}).
Even in smaller $35 \times 35$ mazes, the success rate of VINs drops to 0\% when the required planning steps exceed 60 (see \cref{fig_performance__vs__planning_steps}).

Our work identifies that the principal deficiency causing this is the mismatch between the complexity of planning and the comparatively weak representational capacity of the relatively shallow networks that it uses.
And while there has been moderate success in learning more complicated networks (e.g., GPPN~\citep{lee2018gated} and Highway VINs~\citep{wang2024highway}), until now, VINs of a scale capable of long-term or large-scale planning have not been computationally tractable due to persistent issues with vanishing and exploding gradients---a fundamental problem of deep learning~\citep{Hochreiter:91}.

In this work, we aim to surgically correct deficiencies in VIN-based architectures to enable large-scale long-term planning.
Specifically, we first identify the limitations of the latent MDP in the planning module of VIN and propose a dynamic transition kernel to dramatically increase the representational capacity of the network.
We then build on existing work that identifies the connection between network depth and long-term planning~\citep{wang2024highway} and propose an ``adaptive highway loss'' that selectively constructs skip connections to the final loss according to the actual number of planning steps.
This approach helps mitigate the vanishing gradient problem and enables the training of very deep networks.
With these changes, we find that our new \emph{Dynamic Transition Value Iteration Network} (\emph{DT-VIN}), is able to be trained with $5000$ layers and scale to $1800$ planning steps in a $100 \times 100$ maze navigation task (compared to the original VIN, which only scaled to 120 planning steps in a $25\times 25$ maze).
We apply our method to various challenging tasks, including 2D/3D maze navigation tasks~\citep{Wydmuch2019ViZDoom}, continuous control~\cite{gymnasium,fu2020d4rl}, and real-world Lunar rover navigation tasks~\citep{apollo17orthomosaic}.
We find that DT-VINs can solve both despite these problems requiring hundreds to thousands of planning steps.
Together, these demonstrate the practical utility of our method on vision-based tasks that previous methods are simply unable to solve.
This also serves to highlight the potential of our method to scale to increasingly complex planning tasks alongside the increasing availability of computing power.

\section{Preliminaries}\label{sec_Preliminaries}
\paragraph{Reinforcement Learning (RL) and Imitation Learning (IL).}
The most common formalism used for RL is that of the Markov Decision Process (MDP)~\citep{bellman1957markovian}.
We consider an MDP---as per \citet{puterman2014markov}---to be the $6$-tuple ($\mathcal{S}, \mathcal{A}, {\cal T}, \mathcal{R}, \gamma, \mu$), where $\mathcal{S}$ is a countable state space, $\mathcal{A}$ is a finite action space, ${\cal T}(s'|s,a)$ represents the transition probability to state $s' \in \mathcal{S}$ from state $s \in \mathcal{S}$ and taking action $a \in \mathcal{A}$, $\mathcal{R}(s,a,s')$ is the reward function, $\gamma \in [0,1)$ is a discount factor, and $\mu$ is a distribution over initial states. The behaviour of an artificial agent in an MDP is defined by its policy $\pi(a|s)$, which specifies the probability of taking action $a$ in state $s$. The state value function $V^{\pi}(s)$ is the expected discounted sum of rewards from state $s$ and following policy $\pi$, i.e., $V^\pi (s) \triangleq \E \left[ \sum_{t=0}^{\infty} \gamma^t \mathcal{R}(s_t,a_t,s_{t+1})  | s_0=s; \pi \right]$.
The goal of RL is usually to find an optimal policy $\pi^*$ that achieves the highest expected discounted sum of rewards. The value function of an optimal policy is denoted by $V^*(s)= \max_{\pi} V^\pi(s)$, and satisfies $V^{\pi^*}(s) = V^{*}(s)  \forall s$. The Value Iteration (VI) algorithm iteratively applies the following update to all states to obtain the optimal value function: $V^{(n+1)} (s) = \max_a \sum_{s^{\prime}}{\mathcal{T} \left( s^{\prime}|s,a \right)}\left[ \mathcal{R} \left( s,a,s^{\prime} \right) +\gamma V^{(n)}\left( s^{\prime} \right) \right]$, where $n$ is the iteration number. 
In scenarios where designing a comprehensive reward function is difficult, IL offers a practical alternative. IL enables agents to learn from human or algorithmic demonstrations, with approaches like Behavioral Cloning directly mimicking expert actions in similar states \citep{bain1995framework, schaal1996learning, ross2011reduction}.

\begin{figure*}[bt]
    \centering
    \includegraphics[width=.75\linewidth]{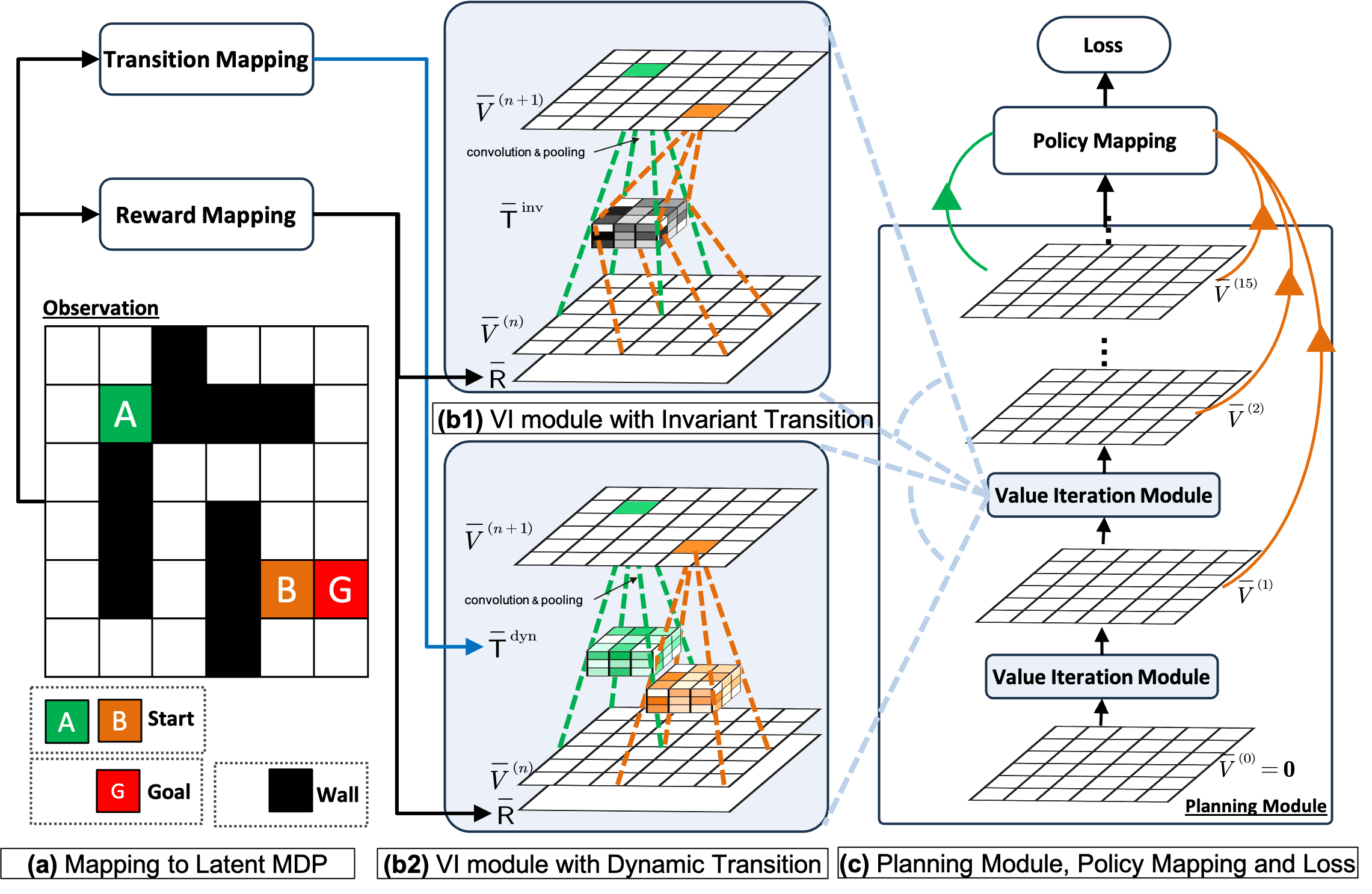}
    \caption{
        The architecture of VIN and DT-VIN in the maze navigation task.
        (a) shows the observation of the maze, which is mapped to the latent reward/transition matrix of the latent MDP through the reward/transition mapping module.
        (c) shows the ``planning module'', the policy mapping module and the loss.
        The ``planning module'' contains numerous stacked Value Iteration (VI) modules.
        The green and orange connections show an example of adaptive highway loss for planning tasks starting from A and B, respectively.
        (b1) shows the VI module of the original VIN with invariant transition $\latentTransition^{\rm inv} \in \mathbb{R}^{|\mathcal{A}| \times F \times F} $.
        (b2) shows the VI module of DT-VIN with dynamic transition kernel $\latentTransition^{\rm dyn} \in \mathbb{R}^{M \times M \times |\mathcal{A}| \times F \times F} $.
    }
    \label{fig_overview}
\end{figure*}

\paragraph{Value Iteration Networks (VINs).}
VIN is an end-to-end differentiable architecture that conducts planning on a latent MDP $\overline{\mathcal{M}}$~\citep{tamar2016value}.
Below, we use $\overline{\:\cdot\:}$ to denote all the terms associated with the latent MDP $\overline{\mathcal{M}}$.
For each decision, VIN first maps an observation $x$,
e.g., an image of a maze and the agent's position, to $\overline{\mathcal{M}}$.
$\overline{\mathcal{M}}$ is described by the latent state space $\latentSSpace=\{(i,j)\}_{i,j \in [M]}$, where $M$ denotes its size; a fixed discrete latent action space $\latentASpace$; a latent reward matrix $\latentReward=f^{ \latentReward  }( x ) \in \mathbb{R}^{M \times M}$, where $f^{\latentReward}$ is a learnable NN called a \emph{reward mapping module};
and a latent transition kernel $\latentTransition^{\rm inv} \in \R^{ |\overline{\mathcal{A}}| \times F \times F } $ with $F$ representing the dimension of the kernel.
The latent transition kernel is a learnable parameter matrix that is invariant of both the latent state and the observation $x$.
Next, VIN conducts VI on the latent MDP $\overline{\mathcal{M}}$ to approximate the latent optimal value function $\overline{V}^*$.
To ensure the differentiability, a differentiable VI module is proposed, simulating VI computation using CNN operations, i.e., convolutional and max-pooling operations:
\begin{equation}\label{eq_VIN}
\overline{V}_{i,j}^{(n)}=\max_{\overline{a}} \sum_{i^{\prime},j^{\prime}}{\overline{\mathsf{T}}_{\overline{a},i^{\prime},j^{\prime}}^{\rm inv}\left( \overline{\mathsf{R}}_{i-i^{\prime},j-j^{\prime}}+\overline{V}_{i-i^{\prime},j-j^{\prime}}^{(n-1)} \right)}
\end{equation}
This equation sums over a matrix patch centered around position $(i,j)$.
After the above, by stacking the VI module for $N$ layers, the latent value function is then fed to a policy mapping module by $f^{\pi}$ to represent a policy that is applicable to the actual MDP $\mathcal{M}$.
Here, $ f^{\pi} \left( \overline{V}^{(n)}(x), a \right) $ represents the probability of taking action $a$ given observation $x$.
Finally, the model can be trained by standard RL and IL algorithms with the following general loss:
$\mathcal{L} \left( \theta \right) =\frac{1}{|\mathcal{D} |}\sum_{\left( x,y \right) \in \mathcal{D}}^{}{ \ell \left( f^{\pi}\left( \overline{V}_{}^{(N) }(x), \cdot \right) ,y \right)}$,
where $\mathcal{D}=\{(x,y)\}$ is the training data, $x$ is the observation, $y$ is the label, and $\ell$ is the sample-wise loss function.
The specific meaning of these items varies depending on the task.
For example, in imitation learning, where the expert data is provided, the label $y$ is the expert action and $\ell$ is the cross-entropy loss, i.e., $
\ell \left( f^{\pi}\left( \overline{V}_{}^{(N) }(x) \right) ,y \right)=-\sum_{a \in \mathcal{A}}{
        \indicatorFunctionSymbol_{ \lbrace a = y \rbrace } \log f^{\pi} \left( \overline{V}^{(n)}(x), a \right)
    }
$, where $\indicatorFunctionSymbol$ is the indicator function.

\section{Method}

In this section, we discuss how to train scalable VINs for long-term large-scale planning tasks.
Our method addresses the two key issues with VIN that are identified as hampering its scalability: the capacity of the latent MDP representation and the depth of the planning module.

\subsection{Improving Latent MDP's Representation Capacity}\label{sec_method_dynamic_transition}

VIN utilizes CNNs to simulate the VI computation process, where the latent transition kernel is implemented as a learnable parameter $\latentTransition^{\rm inv} \in \R^{ |\overline{\mathcal{A}}| \times F \times F }$, as described in \cref{sec_Preliminaries}.
However, there is a discrepancy between the computation process of CNNs and the general VI.

\emph{First, the latent transition kernel of VIN is {invariant} for each latent state $\latentS=(i,j)$.}
This severely limits the latent MDP’s capacity to represent the complex, state-dependent transitions of the real MDP--an ability that is essential for its effectiveness.
For example, in the maze navigation problem shown in \cref{fig_overview} (a), the transition probabilities are quite different if the adjacent cell is a wall versus an empty cell.

Therefore, we propose to use a \emph{latent state-dynamic transition kernel} $\overline{\mathsf{T}}^{\rm dyn} \in \R^{
M \times M \times  |\overline{ \mathcal{A}} | \times F \times F }$
and the augmented VI module is computed as follows:
\begin{fontsize}{9.4}{1}
\begin{equation}\label{eq_DT_VIN}
\overline{V}_{i,j}^{(n)}=\max_{\overline{a}} \sum_{i^{\prime},j^{\prime}}{\overline{\mathsf{T}}_{i,j,\overline{a},i^{\prime},j^{\prime}}^{\rm dyn}\left( \overline{\mathsf{R}}_{i-i^{\prime},j-j^{\prime}}+\overline{V}_{i-i^{\prime},j-j^{\prime}}^{(n-1)} \right)}.
\end{equation}
\end{fontsize}

Different from VIN's VI module in \Cref{eq_VIN}, which uses an invariant kernel $\overline{\mathsf{T}}^{\rm inv} \in \R^{ |\overline{\mathcal{A}}| \times F \times F } $ that is same across all latent states $(i,j)$, here \Cref{eq_DT_VIN} utilizes a dynamic kernel $\overline{\mathsf{T}}^{\rm dyn}_{i,j}$ that adjusts dynamically to each latent state $\overline{s}=(i,j)$.

Second, although the original VIN paper proposes a general framework where the latent transition kernel depends on the observation, i.e., $\latentTransition^{\rm inv}=f^{\latentTransition}\left(x\right)$, with $f^{\latentTransition}$ as a learnable {transition mapping module}, \emph{the latent transition kernel in all VIN's experiments is implemented as a learnable parameter independent of the observation}.
Intuitively, ignoring observations prevents VIN from exploiting any information in the observation to model varying transition dynamics across different environments.
While this limitation does not impact VIN's performance in small-scale, short-term planning tasks (as were tested on in the original work) where the state space is limited and only a few steps are needed to reach the goal, we found it to be a major barrier to VIN's effectiveness when employed to plan on large-scale, long-term planning tasks with completely different observations.
Therefore, we propose using observation-dynamic transition kernels, which employ a learnable transition mapping module $f^{\latentTransition}$ to dynamically generate the latent transition kernel for each observation.

The notions of latent state-dynamic and observation-dynamic features are orthogonal, each capturing an independent aspect of variation in the latent transition kernel.
Their combination yields four distinct configurations, along with corresponding implementations, as summarized in \Cref{table_latent_transition_kernel_types}.
Our experiment in \Cref{sec_abaltion_study} shows that the fully dynamic kernel—which varies with both the latent state and the observation—achieves the best performance.
To implement this kernel, we design a CNN-based architecture for the transition mapping module $f^{\mathsf{T}}$, which maps each local $F' \times F'$ maze patch to a $|\latentASpace| \times F \times F$ latent transition kernel for each latent state,
where $F'$ is the convolutional kernel size of the transition mapping module, and $F$ is the size of the latent transition kernel.
This architecture only requires $ |\latentASpace| {F'}^2 F^2 $ additional parameters.
Note that in practice, setting both $F'$ and $F$ to the small value $3$ suffices to achieve strong performance.
Thus, this alternative module greatly improves the representation capacity of VIN, but typically does not introduce a significant change in training cost.

Lastly, we further enforce a softmax operation on the values of the latent transition kernel for each latent state $\latentS$ to avoid the gradient exploding problem.
This not only provides a statistical semantic meaning to the kernels but also plays a crucial role in ensuring the training stability, as will be demonstrated in \cref{sec_abaltion_study}.

Altogether, these above components form our proposed method, \emph{Dynamic Transition VINs (DT-VINs)}, which employs latent transition kernels that dynamically adapt based on both the latent state and observation, along with a softmax normalization operation.

\begin{table}[t]
\centering
\caption{
Types of latent transition kernels, categorized by whether they are \emph{latent state-dynamic} (i.e., vary with the latent state) and \emph{observation-dynamic} (i.e., vary with the observation).
\label{table_latent_transition_kernel_types}
}
\begin{adjustbox}{max width=\linewidth}
\begin{tabular}{|
>{\centering\arraybackslash}p{2.1cm}|
>{\centering\arraybackslash}p{1.85cm}|
>{\centering\arraybackslash}p{1.8cm}|
>{\centering\arraybackslash}p{4.8cm}|
}
\hline
    \textbf{Types}
    & \textbf{Latent state-dynamic?}
    & \textbf{Observation-dynamic?}
    & \textbf{Implementation}
    \\
\hline
    \textbf{fully invariant} 
    & \xmark 
    & \xmark 
    & \makecell{
        learnable parameter  \\
        $ \overline{\mathsf{T}} \in \R^{ |\overline{\mathcal{A}}| \times F \times F } $ 
    }
\\
    \textbf{latent state- dynamic only}
    & \cmark 
    & \xmark 
    & \makecell{
     learnable parameter \\
     $ \overline{\mathsf{T}} \in \R^{ M \times M \times  |\overline{ \mathcal{A}} | \times F \times F } $ 
    }
\\
    \textbf{observation-dynamic only} 
    & \xmark 
    & \cmark 
    & \makecell{
    learnable model $f^{ \overline{ \mathsf{T} } }(x)$ which \\ 
    output
    $\overline{ \mathsf{T}  } \in  \in \mathbb{R}^{|\bar{\mathcal{A}}| \times F \times F}$ 
    }
\\
    \textbf{fully dynamic} 
    & \cmark 
    & \cmark 
    & \makecell{ 
    learnable model $f^{ \overline{ \mathsf{T} } }(x)$ which \\ 
    output
    $\overline{ \mathsf{T} } \in \mathbb{R}^{M \times M \times |\bar{\mathcal{A}}| \times F \times F}$ } 
    \\
\hline
\end{tabular}
\end{adjustbox}
\end{table}

\subsection{{Increasing Depth of Planning Module}}\label{sec_method_depth}

Recent work on Highway VIN has demonstrated the relationship between the depth of VIN's planning module and its planning ability~\citep{wang2024highway}.
A deeper planning module implies more iterations of the value iterations process, which is proven to yield a more accurate estimation of the optimal value function (see Theorem 1.12)~\citep{agarwal2019reinforcement}.
However, training very deep neural networks is challenging due to the vanishing or exploding gradient problem~\citep{Hochreiter:91}.
Highway VINs address this issue by incorporating skip connections within the context of reinforcement learning, showing similarities to existing works for classification tasks~\citep{srivastava2015training, he2016deep}.
Although Highway VINs can be trained with up to 300 layers, they still fail to achieve perfect scores in larger-scale and longer-term planning tasks and necessitate a more intricate implementation.
Here, we present a simpler, easy-to-implement method for training very deep VINs.

To facilitate the training of very deep VINs, we also adopt the skip connections structure but implement it differently.
Our central insight is that short-term planning tasks generally require fewer iterations of value iteration compared to long-term planning tasks. This is because the information from the goal position propagates to the start position in fewer steps when their distance is short.
Therefore, we propose adding additional loss to shallower layers directly when the task requires only a few steps during the training process.
We achieve this by introducing the following \emph{adaptive highway loss}:
\begin{fontsize}{7.8}{1}
\begin{equation}\label{eq_loss_highway}
\begin{aligned}
&\mathcal{L}\left( \theta \right) =
\frac{1}{K  |\mathcal{D}| } \cdot \\
& \sum_{\left( x,y,l \right) \in \mathcal{D}}^{}
{\sum_{1\le n\le N}{
\indicatorFunction[ n \geq l ]
\indicatorFunction[ n \ \mathrm{ mod }\ J =0 ]
\ell
\Bigg( f^{\pi} \left( \overline{V}^{(n)}(x), \cdot \right) ,y \Bigg)}}
\end{aligned}
\end{equation}
\end{fontsize}Here,
$
K=\sum\limits_{\left( x,y,l \right) \in \mathcal{D}}^{}{
\sum\limits_{1\le n\le N}{
\indicatorFunction[ n \geq l ]
\indicatorFunction[ n \ \mathrm{ mod }\ J =0 ]
}}
$,
$
\indicatorFunctionSymbol
$ is the indicator function,
and $l$ is the length of the planning path or trajectory, which can be computed from the training data.
For example, in the imitation learning of the maze navigation task, the length $l$ of the provided expert path from start to goal is inherently known for each maze in the dataset.
Here, \Cref{eq_loss_highway} presents a general form of $\mathcal{L}$ with an arbitrary single-term loss function $\ell$ (e.g., cross-entropy loss); see \Cref{sec_app_loss_function} for its specific form in different cases.

As \cref{eq_loss_highway} implies, it constructs skip connections for the hidden layers to improve information flow, similar to existing works such as Highway Nets and subsequently, Residual Nets~\citep{srivastava2015training,he2016deep}.
However, we connect hidden layers directly to the final loss,
while existing works typically connect skip connections between the intermediate layers.
Besides, we construct skip connections for each layer $n\geq l$ rather than at the specific layer $n=l$, as the value iteration process should converge to the correct solutions even with additional iterations.
Note that this additional loss will not alter the inherent structure of the value iteration process and will be removed during the execution phase.
Moreover, to decrease computational complexity, we only apply adaptive highway loss to the layers that satisfy the condition $n\ \mathrm{mod}\ J=0$, where $J\geq 1$ is a hyperparameter.

\section{Experiments}\label{sec_experiment}

We perform several experiments to test if our modifications to VIN's planning module allow training very deep DT-VINs for large-scale long-term planning tasks.
Following previous work \citep{tamar2016value,lee2018gated}, we focus on the imitation learning scenario, where we leverage expert demonstrations to evaluate planning capabilities. 

In line with previous works (e.g.,~\citep{lee2018gated}), we assess our planning algorithms on maze navigation tasks, encompassing both 2D and 3D environments \citep{Wydmuch2019ViZDoom}, as detailed in \Cref{sec_2dmaze}. To demonstrate the potential for complex action spaces, we further evaluate the generality of DT-VIN on continuous control tasks (\Cref{sec_robotic_navigation}), which present more challenges than the continuous control tasks described in the original VIN paper due to increased complexity of the maze and the physics \citep{fu2020d4rl}. Additionally, we assess DT-VIN in rover navigation tasks (\Cref{sec_rover}), where the agent must perform planning based on orthomosaic images \citep{apollo17orthomosaic}.

Each task includes a start position and a goal position.
We say that an agent has succeeded in a task if it can navigate from the start position to the goal position within a predetermined number of steps ($M^2$ in our paper).
To assess planning capabilities across tasks of varying complexities, our experiments evaluate various tasks distinguished by their \textit{shortest path lengths (SPLs)}.
The SPL represents the length of the expert path, derived from either human or algorithmic experts.
We further define a path as relatively optimal if it has the shortest length among all solutions provided by various models, including the expert solution.
We follow GPPN and use these for the \textit{success rate (SR)}, which is the rate at which the algorithm succeeds in the task, and the \textit{optimality rate (OR)}, which is the rate at which the algorithm provides a relatively optimal path.

On the above tasks, we compare our DT-VIN method with several advanced neural networks designed for planning tasks, including the original VIN~\citep{tamar2016value}, GPPNs~\citep{lee2018gated}, and Highway VIN~\citep{wang2024highway}.
The models are trained through imitation learning using a labeled dataset.
We then identify the best-performing model based on its results on a validation dataset and evaluate it on a separate test dataset.
Following the methodology from the GPPN paper, we conduct evaluations using three different random seeds for each algorithm.
This is sufficient to provide a reliable performance estimate here due to the low standard deviation we observe in the tasks.
All figures that show learning curves report the mean and standard deviation on the test set.

\subsection{Maze Navigation}\label{sec_2dmaze}

\paragraph{Setting.}
We first evaluate 2D maze navigation tasks with sizes $M$ set to $15$, $35$, and $100$, where the agent moves one step at a time to any of the four adjacent cells.
Each task is defined by a starting position, a visual representation of the $M \times M$ \emph{map design} matrix (with 0 and 1 representing obstacles and roads, respectively), and an $M \times M$ goal matrix indicating the position of the goal.
These mazes require hundreds to thousands of planning steps, as shown in \Cref{fig_maze_100} and further detailed in Appendix \Cref{fig_maze_various_size}.
For each maze size, we generate a dataset following the methodology in GPPN~\citep{lee2018gated} and ensure no overlap between training and test datasets.
Specifically, each unique obstacle arrangement---with its varying start positions and goals---is exclusive to either the training or the test dataset, preventing information leakage.
To assess the performance of each algorithm, we test various neural network depths $N$.
For the largest mazes ($M=100$), we specifically examine $N=600$ and $N=5000$, while for $M=35$, we test multiple depths: $N=30$, $100$, $300$, and $600$.
\removeForSavingSpace{Specifically, for mazes of size $M=15$, we examine depths in $N={30,100,200}$. }
For more details, see \cref{sec_experiment_detail_app_2D_maze}.

\begin{figure*}[t]
    \def\height{0.125}
    \centering
    \includegraphics[width=0.4\linewidth]{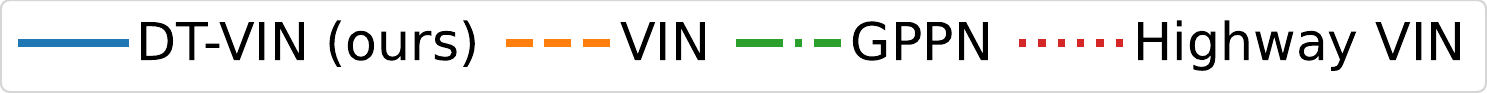}
    \centerline{
        \subfloat[Success Rate \vspace{1em}]{
            \label{fig__success_rate__algorithm__best_depth}
            \includegraphics[height=\height\linewidth]{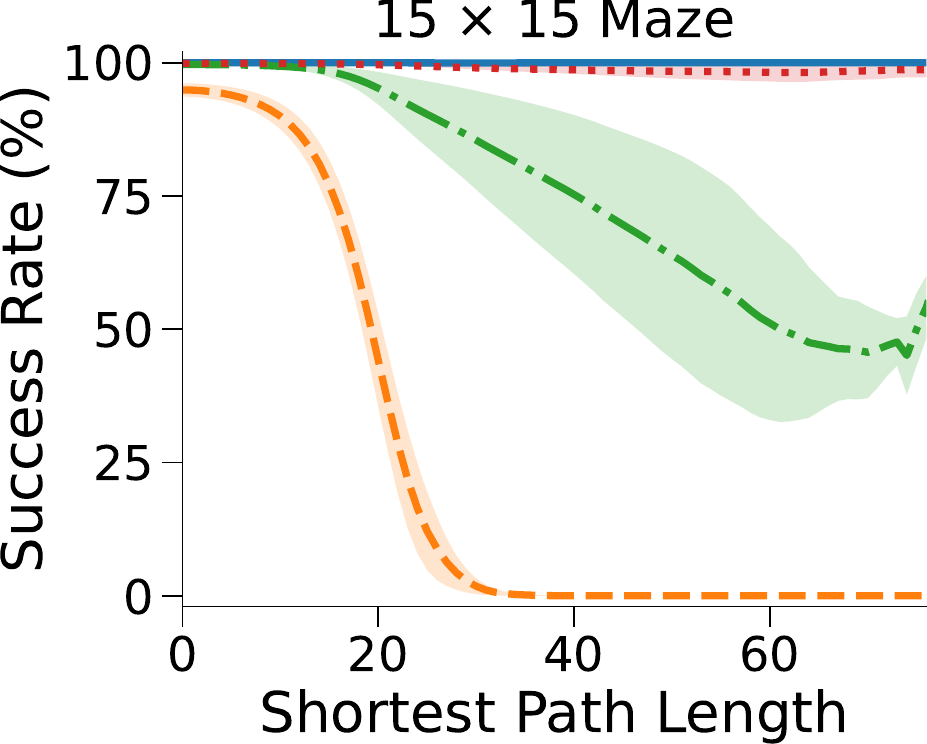}
            \includegraphics[height=\height\linewidth]{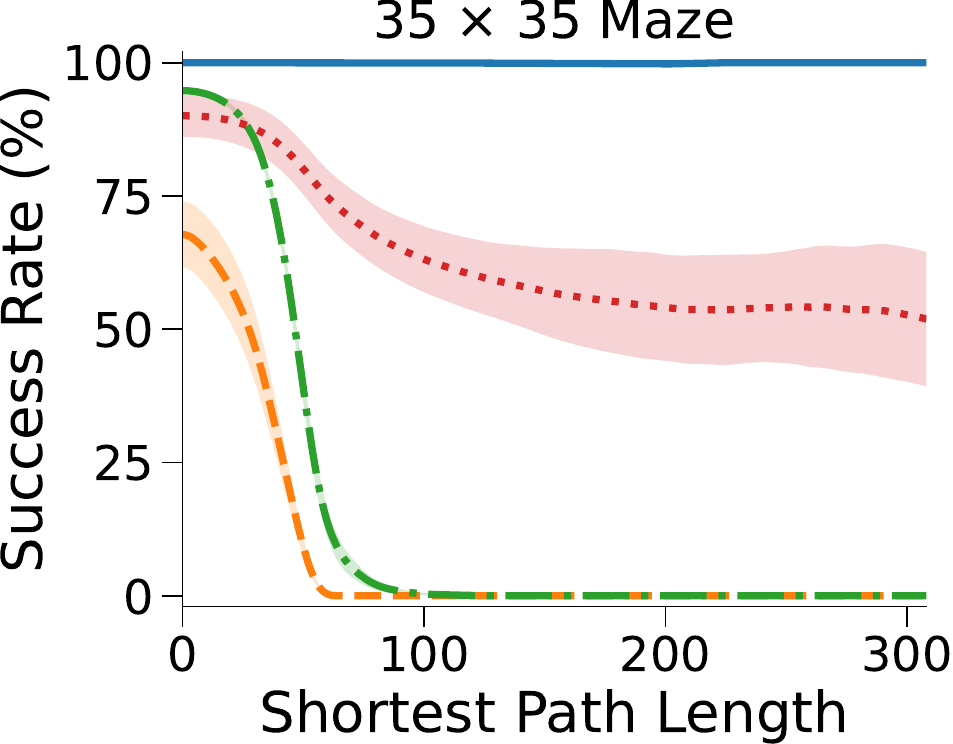}
            \includegraphics[height=\height\linewidth]{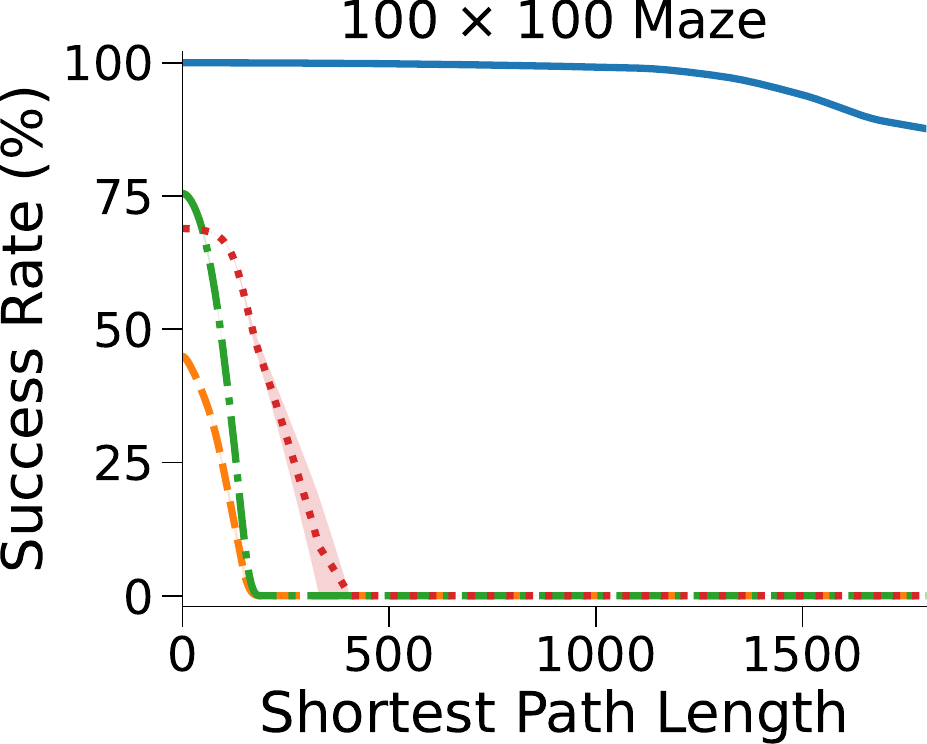}
            }
    \hspace{0.2pt}
    {\color{gray}\rule{0.2pt}{\height\linewidth}}
        \subfloat[Optimality Rate ]{
    \label{fig__optimality_rate__algorithm__best_depth}
            \includegraphics[height=\height\linewidth]{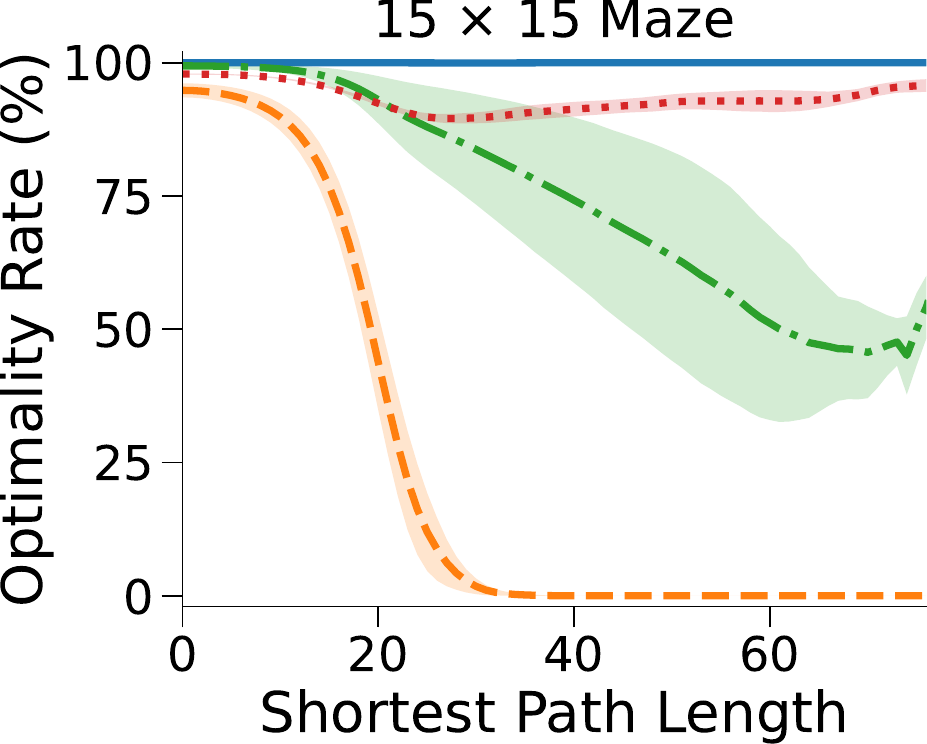}
            \includegraphics[height=\height\linewidth]{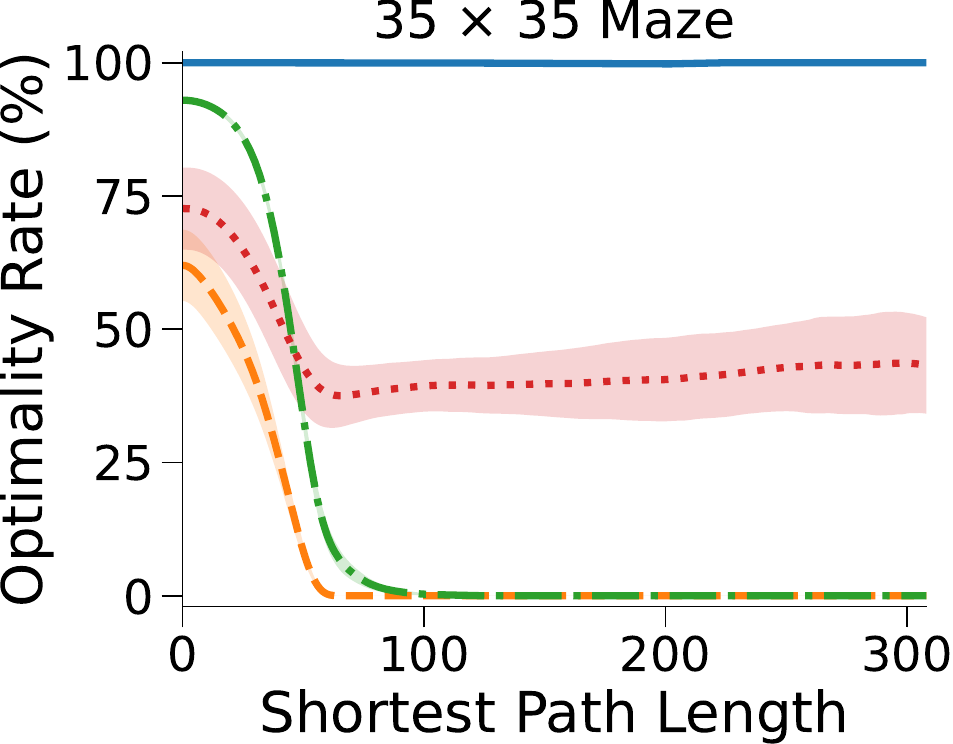}
            \includegraphics[height=\height\linewidth]{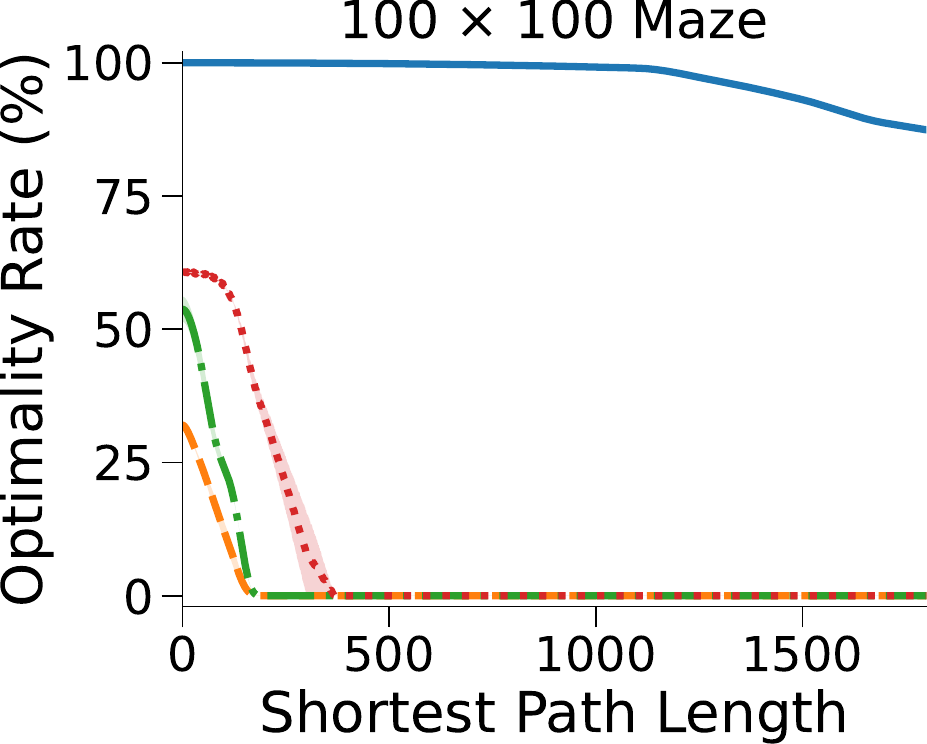}
        }
    }
    \caption{
        SRs and ORs for different algorithms as a function of the shortest path length on 2D maze navigation tasks with various sizes.
        For each algorithm, we select the best result across various depths.
        Specifically, for our DT-VIN, the optimal depth consistently corresponds to the maximum value in the range: $200$ for mazes of size $15$, $600$ for size $35$, and $5000$ for size $100$.
        For other methods, the optimal depth differs per task. In the maze of size $100$, the optimal depth for all the baselines is $600$.
        See \cref{fig__success_rate__algorithm__all_depths} and \cref{fig__optimal_rate__algorithm__all_depths} in \cref{sec_different_depths} for additional results at other depths.        
    }
\end{figure*}

\paragraph{Results and Discussion.}
\cref{fig__success_rate__algorithm__best_depth} shows the success rates (SRs) of our method and the baseline methods, as a function of the SPLs.
For each algorithm and environment configuration, we report the performance of the NN with the best depth $N$ across the ranges specified in the previous paragraph (see \cref{fig__success_rate__algorithm__all_depths} in \cref{sec_different_depths} for additional results concerning different values of $N$).

Here, DT-VIN outperforms all the other methods on all the maze navigation tasks under all the various sizes $M$ and SPLs.
Notably, on small-scale mazes with size in $M={15,35}$, DT-VIN achieves approximately 100\% SRs on all the tasks.
For the most challenging environment with $M=100$, DT-VIN performs best with the full $5000$ layers,
and it maintains an SR of approximately 100\% on short-term planning tasks with SPL ranging in $[1,200]$ and an SR of approximately 88\% on tasks with SPLs over $1200$.
Comparatively, VIN performs well on small-scale and short-term planning tasks.
However, even on a small-scale maze with size $M=15$, VIN's SRs drop to 0\% when the SPL exceeds $30$.
Moreover, when the maze size increases to $100$, VIN only achieves an SR of less than 40\%---even on short-term planning tasks with SPL within $[1,100]$.
GPPN performs well on short-term planning tasks, but it fails to generalize well on long-term planning tasks, which also decreases to an SR of 0\% as the SPL increases.
Highway VIN performs well across tasks with various SPLs on a small-scale maze with $M={15}$. However, it shows a performance decrease on larger-scale maze tasks with $M={35,100}$.
\cref{fig__optimality_rate__algorithm__best_depth} shows the optimality rates (ORs) of the algorithms, which measure the rate at which the model outputs the optimal path.
Our DT-VIN maintains consistent ORs compared to SRs.
However, some other methods---especially Highway VIN---exhibit a clear decrease in ORs, indicating that the paths generated by these methods is often sub-optimal.

\textbf{Addtional Experiments.}
Due to space constraints, we evaluate challenging cases with noisy maze observations in \Cref{sec_maze_noisy} and different transition types (agents moving to any of the eight Moore neighborhood cells) in \Cref{sec_mechanism}.
We also evaluate 3D ViZDoom navigation with first-person inputs \cite{Wydmuch2019ViZDoom}, where the model plans based on noisy maze matrix predictions (\Cref{sec_3D_navigation}).	
DT-VIN outperforms all compared methods in these cases.

\subsection{Continuous Control}\label{sec_robotic_navigation}

\begin{figure}[h!]
    \centering
    \def\height{0.25}
    \subfloat[Point Maze]{
        \includegraphics[height=\height\linewidth]{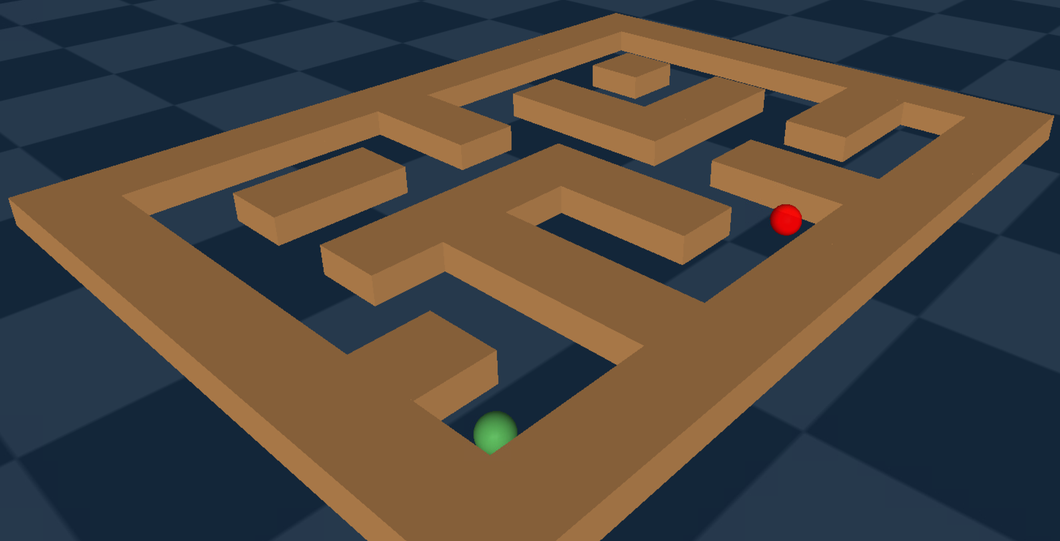}
    }
    \subfloat[Ant Maze]{
        \includegraphics[height=\height\linewidth]{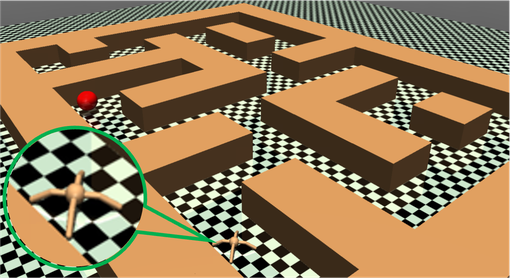}
    }
    \caption{The Point Maze and Ant Maze environment in the training dataset from D4RL \citep{gymnasium,fu2020d4rl}, where map size is $9 \times 12$.
    Please refer to \Cref{sec_app_robotic_navigation} for larger (e.g., $100 \times 100$) mazes in our testing dataset.
    }
    \label{fig_RoboticControl}
\end{figure}

We evaluate the algorithms on two continuous control tasks: Point Maze and Ant Maze ~\citep{gymnasium,fu2020d4rl}.
These tasks are more challenging than simple maze navigation: the agent must not only conduct (explicit or implicit) planning over the complex maze but also output the torques to direct the controlled agent (a 2-DOF ball or a 3D quadruped robot) toward the correct direction, as shown in \Cref{fig_RoboticControl}.
The input includes the maze's top-down view, agent and goal locations, positional values, and velocities of the agent's body parts.
The network architecture integrates a planning module structurally similar to that used in 2D maze navigation. The outputs of this module, combined with additional state information, are aggregated and passed to the policy mapping module to produce the control action.
We train the model using the D4RL offline Point/Ant Maze dataset, which features a fixed $9 \times 12$ maze layout across all tasks, varying only in start and goal positions.
However, during evaluation, the model is tested on substantially larger and more complex mazes ($35 \times 35$ and $100 \times 100$) with previously unseen layouts and start/goal configurations, significantly increasing task complexity and demanding strong generalization capabilities.
To address this challenge, we pretrain the planning module on a diverse and readily available 2D maze navigation dataset to endow the model with transferable planning skills.
Please refer to \Cref{sec_app_robotic_navigation} for more details.
\Cref{tab_continuous_control_results} shows the results of this experiment.
DT-VIN solves the mazes at a much higher rate than all the baseline methods.
Specifically, in the challenging $100 \times 100$ Ant Maze, DT-VIN achieves a notable $51\%$ higher success rate than the second-best baseline.

\begin{table}[t]
\centering
\caption{
The success rates of the methods on Point Maze and Ant Maze tasks.
}
\scalebox{0.7}{
\begin{tabular}{c|cc|cc}
\hline
 & \multicolumn{2}{c|}{\textbf{Point Maze}}  & \multicolumn{2}{c}{\textbf{Ant Maze}} \\
 & $35 \times 35$ & $100 \times 100$ & $35 \times 35$ & $100 \times 100$ \\
\hline
VIN & $67.42_{\pm 7.56}$ & $38.19_{\pm 9.73}$ & $64.85_{\pm 9.34}$ & $31.67_{\pm 11.28}$ \\
GPPN & $91.21_{\pm 1.94}$ & $69.37_{\pm 2.85}$ & $86.09_{\pm 3.46}$ & $61.78_{\pm 6.12}$ \\
Highway VIN & $89.63_{\pm 2.71}$ & $66.45_{\pm 4.83}$ & $82.34_{\pm 4.19}$ & $52.91_{\pm 8.55}$ \\
DT-VIN (ours) & $\mathbf{99.87_{\pm 0.12}}$ & $\mathbf{98.04_{\pm 1.93}}$ & $\mathbf{96.57_{\pm 2.61}}$ & $\mathbf{93.26_{\pm 3.14}}$ \\
\hline
\end{tabular}
\label{tab_continuous_control_results}
}
\end{table}

\subsection{Rover Navigation}\label{sec_rover}

\begin{figure}[th]
\captionsetup[subfloat]{font=scriptsize}
    \def\width{.3}
    \centering
    \noindent
    \subfloat[\footnotesize Terrain image]{ %
        \includegraphics[height=\width\linewidth]{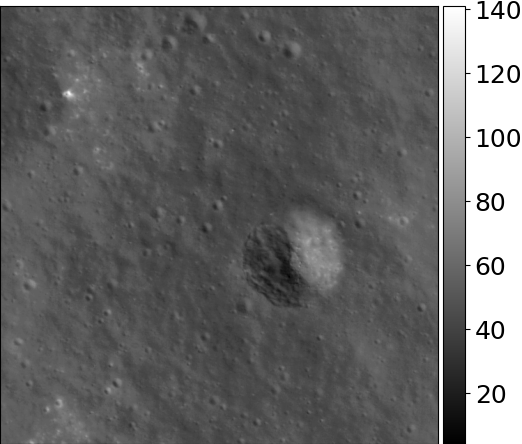}
        \label{fig_rover_image}
    }
    \subfloat[\footnotesize Elevation data]{
        \includegraphics[height=\width\linewidth]{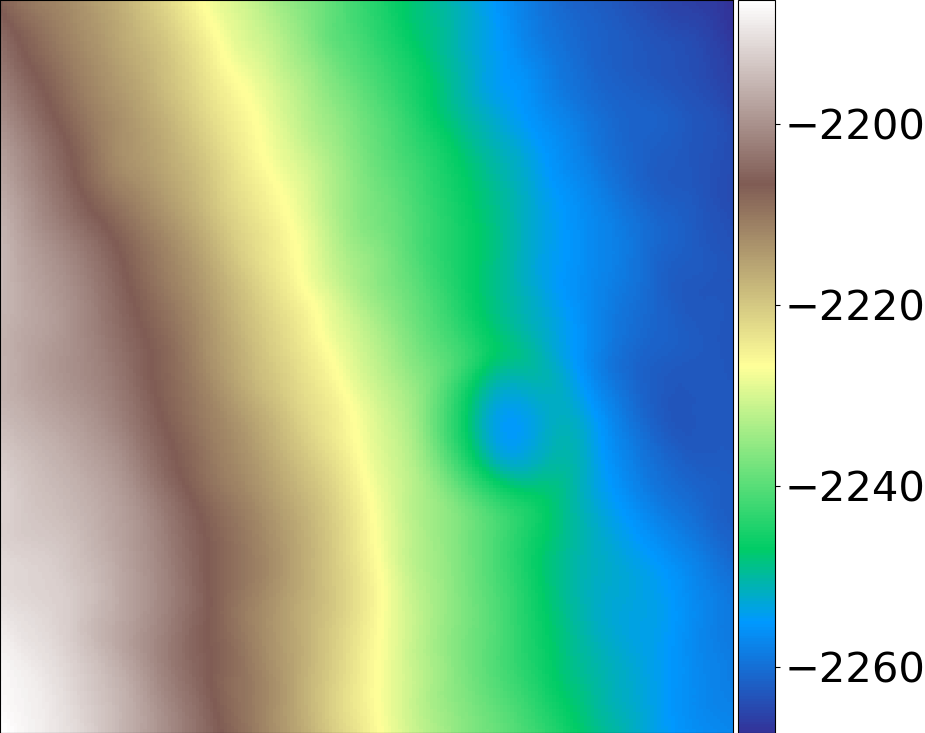}
        \label{fig_rover_elevation}
    }
    \caption{
    \subref{fig_rover_image} and \subref{fig_rover_elevation} show a patch of terrain image and elevation data from Apollo 17 landing tasks.
    }
\end{figure}

\begin{table}[t]
\centering
\caption{
 The success rates of the algorithms on rover navigation tasks with various image sizes.
\label{fig_rover_result}
}
\scalebox{0.80}{
  \begin{tabular}{c|ccc}
        \hline
        \textbf{Image Sizes} & 270$\times$270 & 450$\times$450 & 630$\times$630 \\
        \hline
         VIN          & $85.32_{\pm 0.14} $ & $82.43_{ \pm 0.74 }$ & $71.71_{\pm 3.48 }$ \\ %
        GPPN          & $85.79_{\pm 0.31}$ & $81.72_{\pm 0.22}$ & $76.31_{\pm 0.75}$ \\ %
        Highway VIN  & {$85.81 _{ \pm 0.6} $} & {$81.88 _{ \pm 0.86} $} & {$73.21 _{ \pm 0.81} $}  \\ %
        DT-VIN (ours) & \bm{$86.54 _{ \pm 0.5} $} & \bm{$82.78 _{ \pm 0.6} $} & \bm{$77.4 _{ \pm 0.98} $}  \\
        \hline
     CNN+$A^*$          & $84.42_{\pm 0.67} $ & $82.11_{ \pm 0.74 }$ & $76.19_{\pm 1.23 }$
     \\
     \noalign{\smallskip}
     \hline
    \end{tabular}
}
\end{table}

\begin{figure*}[t]
    \def\height{0.145}
    \hspace{0.1\linewidth}
    \includegraphics[width=0.60\linewidth]{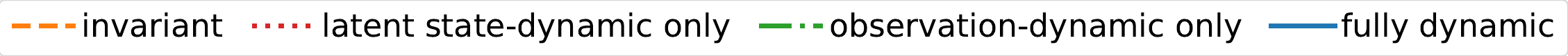}\\
    \centerline{
        \subfloat[SRs w.r.t. SPLs]{
            \includegraphics[height=\height\linewidth]{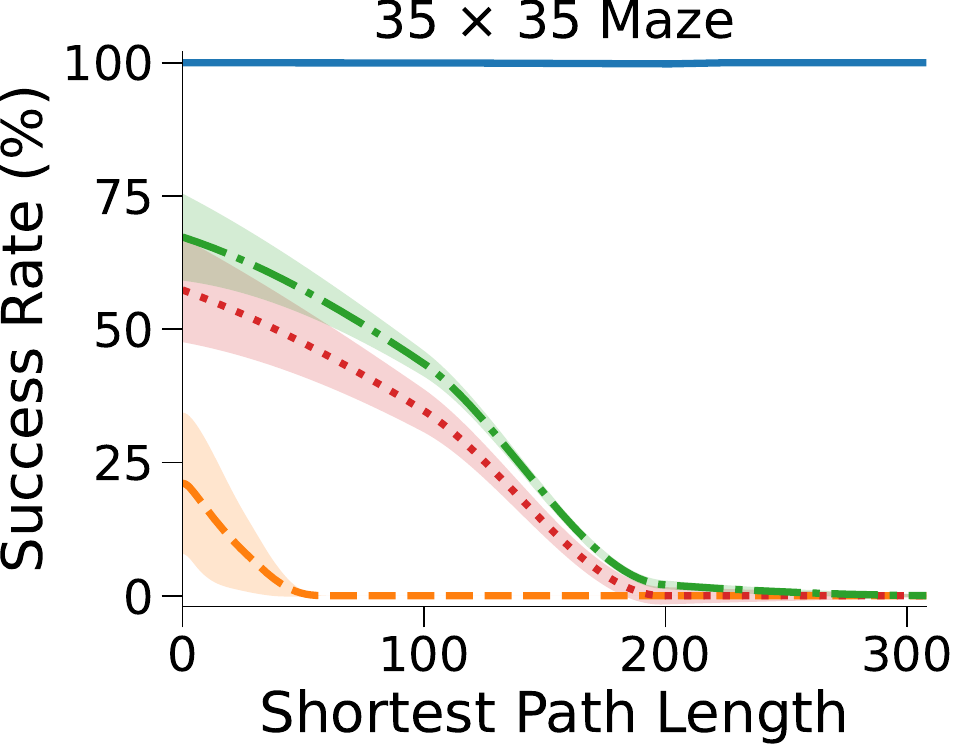}
            \label{fig_DynamicTransition_SR_SPL}
        }
        \subfloat[\scriptsize SRs w.r.t. Obstacle Counts]{
            \includegraphics[height=\height\linewidth]{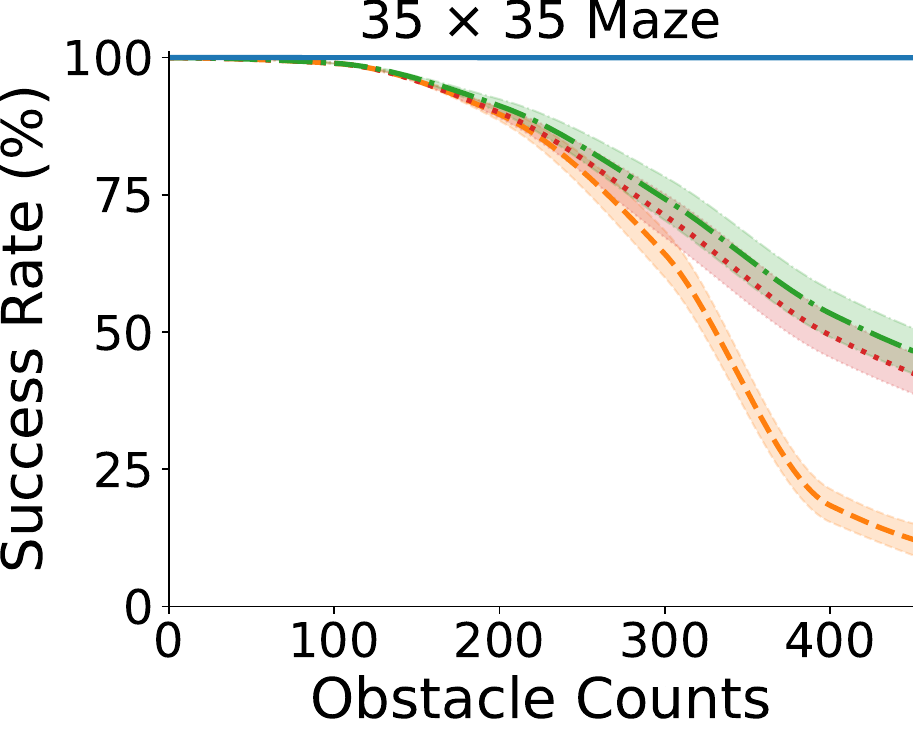}
            \label{fig_DynamicTransition_SR_WALL}
        }
        {\color{gray}\rule{0.2pt}{\height\linewidth}}
    \subfloat[Loss w.r.t. SPLs]{
            \includegraphics[height=\height\linewidth]{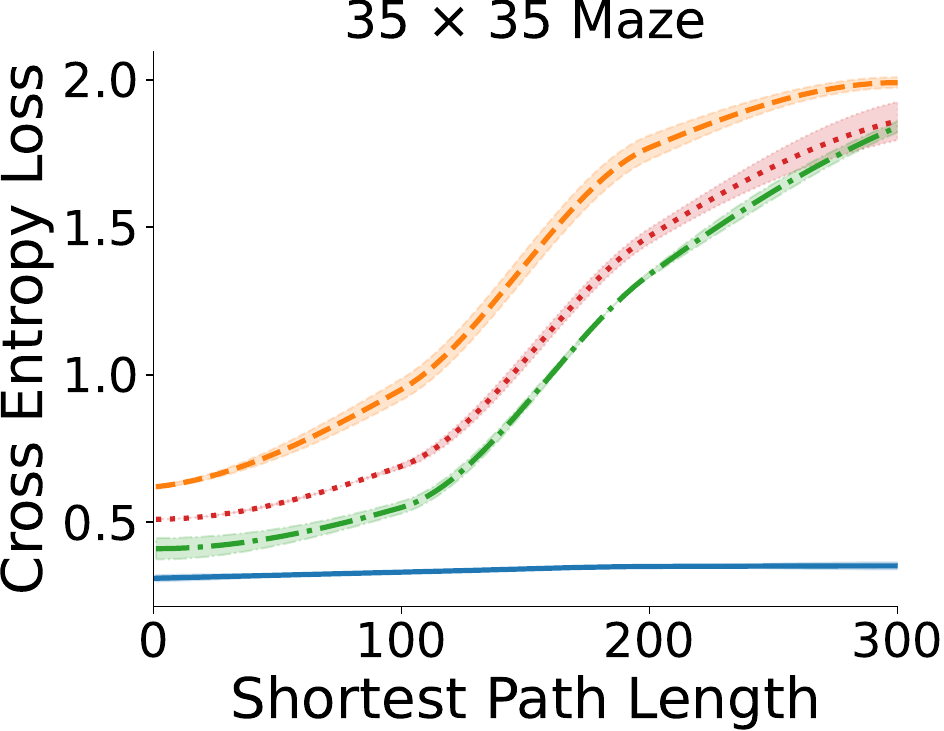}
            \label{fig_DynamicTransition_Loss_SPL}
        }
        \subfloat[\tiny Loss w.r.t. Obstacle Counts]{
            \includegraphics[height=\height\linewidth]{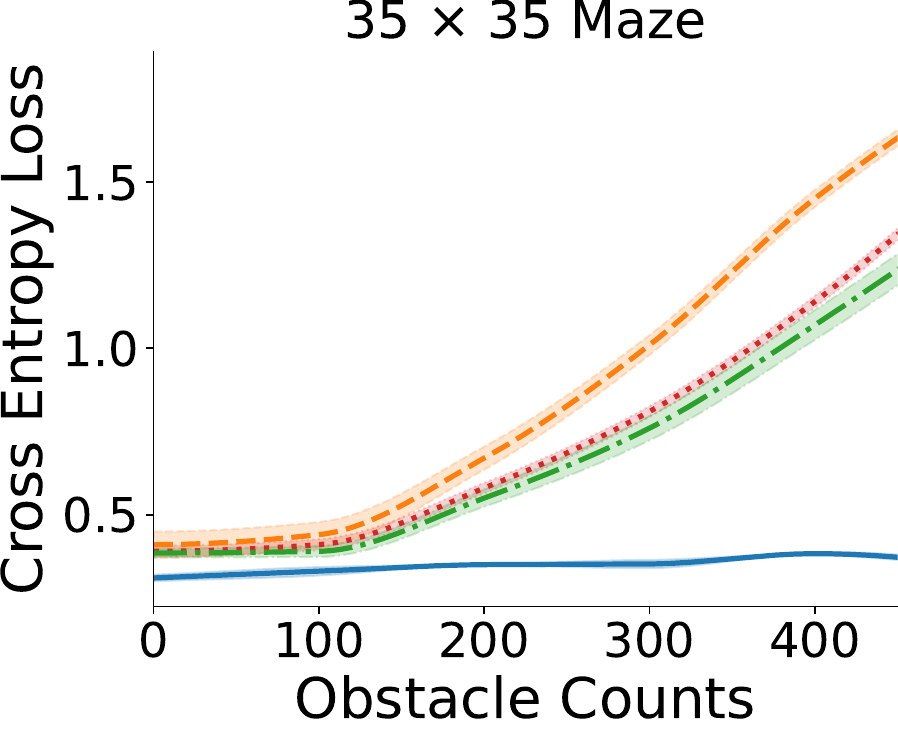}
            \label{fig_DynamicTransition_Loss_Wall}
        }
        {\color{gray}\rule{0.2pt}{\height\linewidth}}
        \subfloat[Learned Kernels]{
            \includegraphics[height=\height\linewidth]{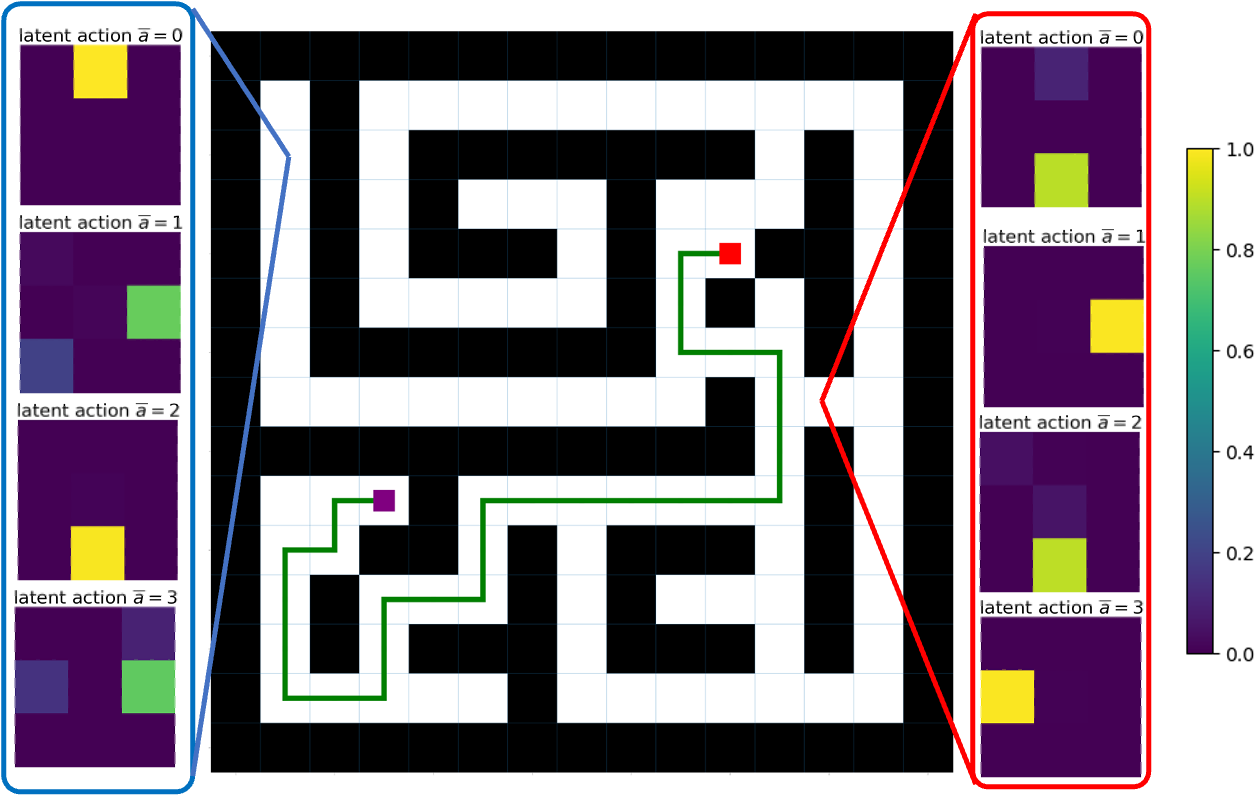}
            \label{fig_dynamic_transition_kernel_demonstration}
        }
    }
    \caption{
        Ablation studies on the \textbf{dynamic latent transition kernel}.
        \subref{fig_DynamicTransition_SR_SPL} and \subref{fig_DynamicTransition_SR_WALL} show the success rates (SRs) w.r.t. to various shortest path lengths (SPLs) and obstacle counts of the tasks, respectively.
        \subref{fig_DynamicTransition_Loss_SPL} and \subref{fig_DynamicTransition_Loss_Wall} show the losses.
        \subref{fig_dynamic_transition_kernel_demonstration} shows the learned latent transition kernels of DT-VIN.
    }
\end{figure*}

\begin{figure*}[t]
    \def\height{0.145}
    \centering
    \centerline{
        \subfloat[\scriptsize Softmax Operation: Success Rate(Left), Gradient Norm(Right)]{
            \includegraphics[height=\height\linewidth]{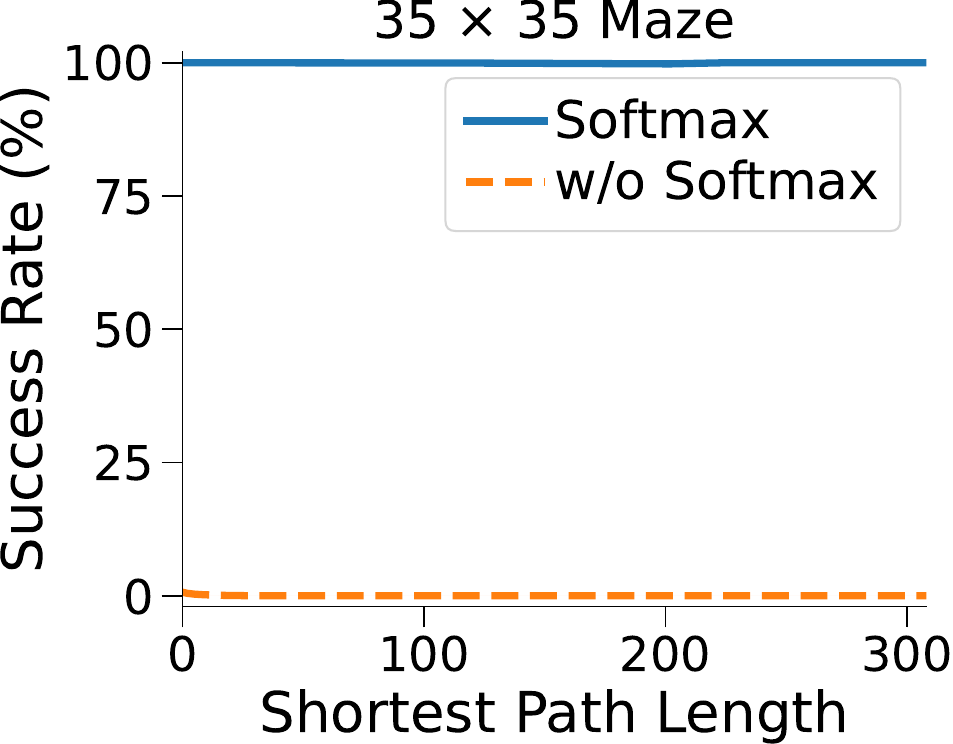}
            \includegraphics[height=\height\linewidth]{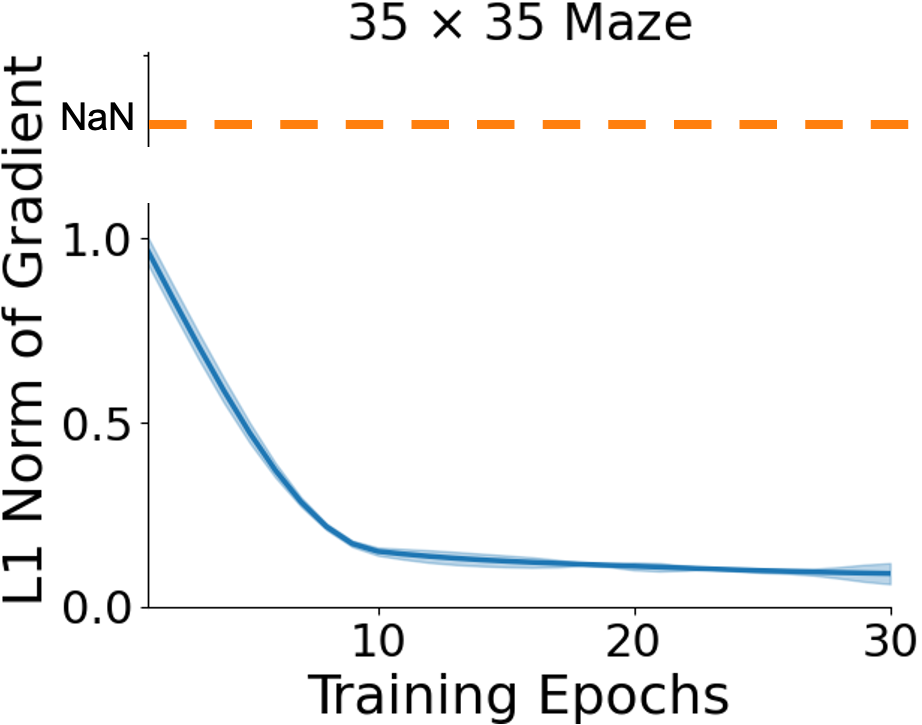}
            \label{fig_softmax}
        }
        {\color{gray}\rule{0.2pt}{\height\linewidth}}
        \subfloat[\tiny Adaptive Highway Loss: Success Rate(Left), Gradient Norm(Right)]{
            \includegraphics[height=\height\linewidth]{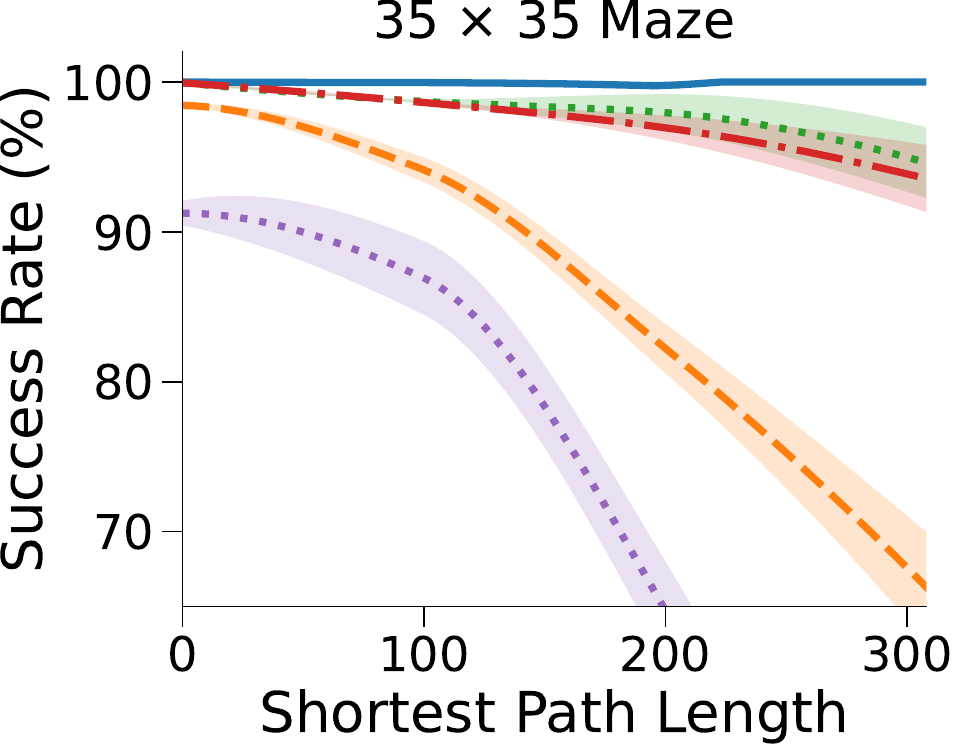}
            \includegraphics[height=\height\linewidth]{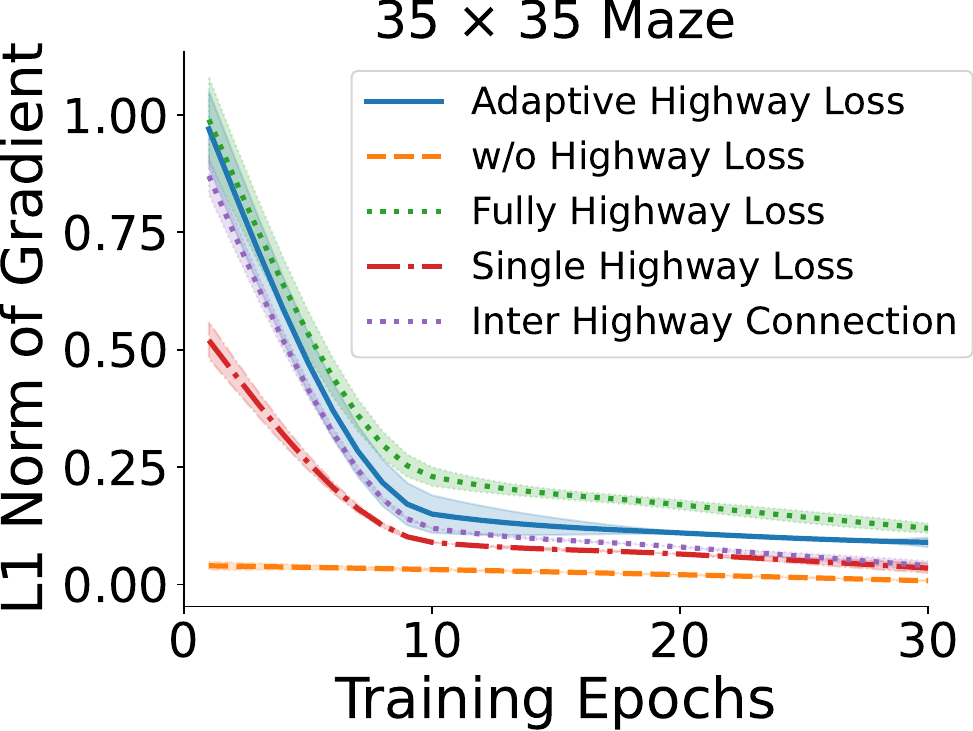}
            \label{fig_adaptive_highwayloss}
        }
        {\color{gray}\rule{0.2pt}{\height\linewidth}}
        \subfloat[Depths of Model]{
            \includegraphics[height=\height\linewidth]{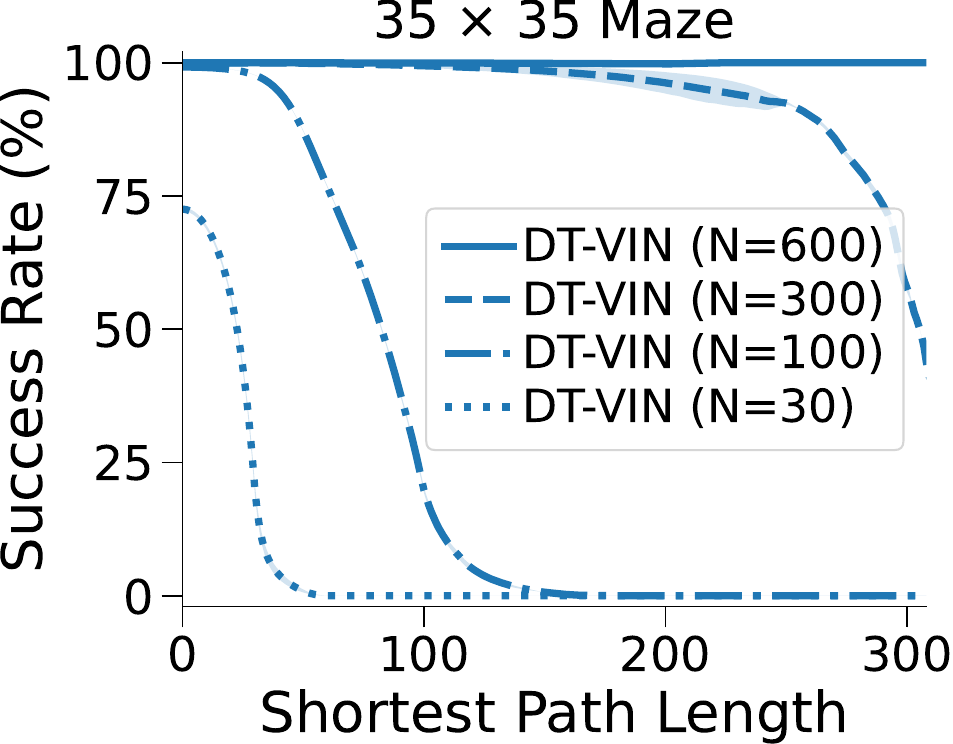}\label{fig_various_depths}
        }
    }
    \caption{
        \subref{fig_softmax} and \subref{fig_adaptive_highwayloss} show ablation on \textbf{softmax operation} and \textbf{adaptive highway loss}, respectively. The left and right sides of each sub-figure show the success rate, and the averaged L1 norm of the gradient over the first 10 layers, respectively.
       \subref{fig_various_depths} show the the success rate across various \textbf{depths of model}.
    }
\end{figure*}

We further evaluate the algorithms on the rover navigation task, where the algorithm conduct planning based on a \emph{terrain image} (e.g., \Cref{fig_rover_image}) rather than the \emph{elevation data} (e.g., \Cref{fig_rover_elevation}).
While terrain images, visual representations from aerial photographs, typically lack elevation information, they are generally more readily available or less expensive to obtain.
We evaluate the Apollo 17 landing tasks, featuring images with a resolution of 0.5 meters per pixel~\citep{apollo17orthomosaic}.
We crop the terrain image into various-sized patches, each $18 \times 18$ patch defining a cell.
The cell is considered an obstacle if the associated area exhibits an elevation angle exceeding 10 degrees. 
Analogous to supervised learning, the training phase uses external elevation data to generate the expert path, while in the testing phase, this data is inaccessible to the models and only used to assess performance metrics.
Please refer to \Cref{sec_Rover_Navigation_app} for details on the task setting and the models' architecture.

As shown in \Cref{fig_rover_result}, our DT-VIN outperforms all compared methods across various terrain image sizes.
Notably, with larger image sizes (particularly $630\times 630$), our DT-VIN outperforms VIN by more than $5\%$.
We also compare with a special baseline, CNN+$A^*$, which combines classification and classical planning algorithms. This method trains a CNN to classify whether an $18 \times 18$ image patch is an obstacle using elevation data, then uses $A^*$ to conduct planning based on this prediction.
While this unsurprisingly is able to outperform VIN, it is still outperformed by DT-VIN.

\subsection{Ablation Study}\label{sec_abaltion_study}
We perform multiple ablation studies with a $M=35$ maze and an NN with depth $N=600$ to assess the impact of DT-VIN components, including the dynamic latent transition kernel and its associated softmax operation, the adaptive highway loss and the network depth. Unless otherwise indicated, all these elements are included.

\textbf{Dynamic Latent Transition Kernel.}
We evaluate four variants of dynamic latent transition kernels:
\emph{fully invariant}, \emph{observation-dynamic only}, \emph{latent state-dynamic only}, \emph{fully dynamic} (incorporating both observation and latent state-dynamic).
\cref{fig_DynamicTransition_SR_SPL} and \subref{fig_DynamicTransition_SR_WALL} show the SRs of all these variants w.r.t. the SPLs and obstacle counts respectively.
The variant with invariant kernels performs the worst. Both observation-dynamic and latent state-dynamic components contribute to the performance of DT-VIN -- removing either results in a substantial performance drop.
It's worth noting that the performance gap between the dynamic kernel and invariant kernel grows with increasing the SPL or obstacle counts.
This improvement is due to the improved model's representational capacity by the dynamic kernel, reducing compounded errors over extended planning horizons and complex environments with increased obstacle counts, as shown in \Cref{fig_DynamicTransition_Loss_SPL} and \subref{fig_DynamicTransition_Loss_Wall}.
\Cref{fig_dynamic_transition_kernel_demonstration} gives an illustration of DT-VIN's dynamic transition kernels.

\textbf{Softmax Operation on Latent Transition Kernel.}
As shown in \cref{fig_softmax} (left), the variant without the softmax operation on the latent transition kernel fails on all the tasks.
This failure is due to exploding gradients, wherein the gradient becomes extremely large, eventually resulting in the model's parameters overflowing and becoming a NaN (Not a Number) value, as depicted in \Cref{fig_softmax} (right).

\textbf{Adaptive Highway Loss.}
We evaluate several additional variants to verify the effectiveness of adaptive highway loss. \Cref{fig_adaptive_highwayloss} (left) shows the success rate of all variants.
\textbf{(a)} The variant \emph{without highway loss} suffer a decrease in performance, as shown in \Cref{fig_adaptive_highwayloss}.
\textbf{(b)} We also evaluate a variant named \emph{fully highway loss}, which enforces a highway loss on each hidden layer without adaptive adjustment by removing the filter $\indicatorFunction[n\geq l]$ in \Cref{eq_loss_highway}.
This variant performs worse than the one with adaptive highway loss, highlighting the necessity of adaptive adjustment.
\textbf{(c)} We assess a variant called \emph{single highway loss}, which only apply highway loss to a specific layer $n$ where $n=l$ by substituting $\indicatorFunction[n\geq l]$ with $\indicatorFunction[n = l]$ in \Cref{eq_loss_highway}.
This variant performs worse than the adaptive highway loss, demonstrating the critical role of the component $n>l$ in performance.
\textbf{(d)}
Finally, we evaluate a variant named \emph{inter highway connection}, which constructs skip connections across the intermediate layers of the planning module,  akin to techniques used in Highway Nets \citep{highway2015}. 
This variant performs the worst because the skip connections over intermediate layers can disrupt the structure of value iteration.
These results are consistent with those in existing work \citep{wang2024highway}.

\textbf{Depth of Planning Module.}
\cref{fig_various_depths} shows the SRs of DT-VIN with various depths.
Here, increasing the depth dramatically improves the long-term planning ability.
For example, for tasks with an SPL of $200$, the variant with depth $N=300$ outperforms the variant with depth $N=100$.
Moreover, for tasks with an SPL of $300$, the deeper variant with depth $N=600$ excels greatly.
Other methods like VIN and GPPN do not show a clear performance improvement when the depth increases, please refer to \cref{fig__success_rate__algorithm__all_depths} in \cref{sec_different_depths} for more details.

\textbf{Additional Ablation Studies.} 
Due to space constraints, please refer to \Cref{sec_app_ablation_study} and \ref{sec_scaling_app}, which show the robustness of DT-VIN with $50\%$ of the training dataset, the choice of the length of the expert path $l$, impact of jumping hyperparameter $J$, and the scaling of computation complexity, dataset requirements, and model size with task scale.

\section{Related Work}

\paragraph{Variants of Value Iteration Networks (VINs).}
Several variants of VIN~\citep{tamar2016value} have been proposed in recent years.
Gated Path Planning Networks employ gating recurrent mechanisms to reduce the training instability and hyperparameter sensitivity seen in VINs~\citep{lee2018gated}.
To mitigate overestimation bias (which is detrimental to learning here), dVINs were proposed and use a weighted double estimator as an alternative to the maximum operator~\citep{jin2021value}.
For addressing challenges in irregular spatial graphs, Generalized VINs adopt a graph convolution operator, extending the traditional convolution operator used in VINs~\citep{niu2018generalized}.
To improve scalability, AVINs introduce an abstraction module that extracts higher-level information from the environment and the goal~\citep{schleich2019value}.
To improve scalability, \citeauthor{zhao2023scaling} proposed implicit differentiable planners that decouple forward and backward passes~\citep{zhao2023scaling}. In contrast, our work takes a complementary architectural approach, redesigning VINs for deeper and more expressive planning over larger-scale tasks.
For transfer learning, Transfer VINs address the generalization of VINs to target domains where the action space or the environment's features differ from those of the training environments~\citep{shen2020transfer}.
More recently, VIRN was proposed and employs larger convolutional kernels to plan using fewer iterations as well as self-attention to propagate information from each layer to the final output of the network~\citep{cai2022value}.
Similarly, GS-VIN also uses larger convolutional kernels but to stabilize training and also incorporates a gated summarization module that reduces the accumulated errors during value iteration~\citep{cai2023value}.
Most related to DT-VIN is other recent work that focused on developing very deep VINs for long-term planning.
Specifically, Highway VIN~\citep{wang2024highway} incorporates the theory of Highway Reinforcement Learning~\citep{wang2023highway} to create deep planning networks with up to 300 layers for long-term planning tasks.
Highway VIN modifies the planning module of VIN by introducing an exploration module that injects stochasticity in the forward pass and uses gating mechanisms to allow selective information flow through the network layers.
Our method, however, achieves even deeper planning by incorporating a dynamic transition matrix in the latent MDP and adaptively weighting each layer's connection to the final output.

\paragraph{Neural Networks with Deep Architectures.}
There is a long history of developing very deep neural networks (NNs).
For sequential data, this prominently includes the LSTM architecture and its gated residual connections, which help alleviate the ``vanishing gradient problem''~\citep{Hochreiter:97lstm,Hochreiter:91}.
For feedforward NNs, a similar gated residual connection architecture was used in Highway Networks~\citep{srivastava2015training} and later in the ResNet architecture~\citep{he2016deep}, where the gates were kept open.
Such residual connections are still ubiquitous in modern language architectures, such as the Generative Pre-trained Transformer (GPT)~\citep{achiam2023gpt}. Our method dynamically employs skip connections from select hidden layers to the final loss, utilizing a state and observation map-dependent transition kernel. This approach is more closely aligned with the computation of the true VI algorithm. Similar kernels, dependent on an input image~\citep{chen2020dynamic} or the coordinates of an image~\citep{liu2018intriguing}, have been previously used in Computer Vision.

\section{Conclusions}

Previous work introduced VIN as a differentiable neural network for planning in artificial intelligence and reinforcement learning.
While VINs have been successful at short-term small-scale planning, they start to fail quite rapidly as the planning horizon and the scale of the tasks grows.
We observed that this decay in performance is principally due to limitations in the (1)~representational capacity of their network and (2)~its depth.
To alleviate these problems, we propose several modifications to the architecture, including a dynamic transition kernel to increase the representation capacity and an adaptive highway loss function to ease the training of very deep models.
Altogether, these modifications have allowed us to train networks with $5000$ layers.
In line with previous work, we evaluate the efficacy of our proposed Dynamic Transition VINs (DT-VINs) on 2D/3D maze navigation, continuous control, and rover navigation.
We find that DT-VINs scale to longer-term and larger-scale planning problems than previous attempts.
To the best of our knowledge, DT-VINs is, at the time of publication, the current state-of-the-art planning solution for these specific environments.
We note that the upper bound for this approach (i.e., the scale of the network and, consequently, the scale of the planning ability) remains unknown.
As our experiments were limited mostly by computational cost and did not observe instability, we expect that with the growth of available computational power, our method will scale to even longer-term and larger-scale planning.
\clearpage

\clearpage

\section*{Acknowledgements}

We gratefully acknowledge the insightful comments from the ICML 2025 reviewers, as well as the constructive feedback received during earlier stages of peer review. We are especially thankful to Yanning Dai for her valuable support with the continuous control experiments.
The research reported in this publication was supported by funding from King Abdullah University of Science and Technology (KAUST) - Center of Excellence for Generative AI, under award number 5940.
We also acknowledge the support by the grant EP/Y002644/1 under the EPSRC ECR International Collaboration Grants program, funded by the International Science Partnerships Fund (ISPF) and the UK Research and Innovation. 
This work was additionally supported by the European Research Council (ERC, Advanced Grant Number 742870).
For computer time, this research used Ibex managed by the Supercomputing Core Laboratory at King Abdullah University of Science and Technology (KAUST) in Saudi Arabia.

\section*{Impact Statement}
This paper aims to advance the field of Machine Learning. While acknowledging there are many potential societal consequences of our work, we believe that none need to be specially highlighted here.

\bibliography{main,juergen}
\bibliographystyle{icml2025}

\clearpage
\appendix
\onecolumn
\section{Limitations and Future Works}

The principal limitation of our work compared to VIN and Highway VIN is the increased computational cost (see \cref{sec_scaling_app}).
This is a consequence of the scale of the network.
The past decades have seen AI dominated by the trend of scaling up systems~\citep{sutton2019bitter}, so this is not likely a long-term issue.

Future work will explore the impact of a more sophisticated transition mapping module (this work uses a single CNN layer for this purpose) in more challenging real-world applications, such as real-time robotics navigation in dynamic and unpredictable environments.
 Additionally, recent diffusion-based planners~\citep{janner2022planning, mishra2023generative} offer a complementary approach to long-horizon reasoning, which may be integrated with VIN-style architectures to improve planning capacity and temporal abstraction.

\section{Method: Additional Details}

\subsection{Loss Function}\label{sec_app_loss_function}
In imitation learning, the loss function in \cref{eq_loss_highway} can be written as
$$
  \mathcal{L}\left( \theta \right) =\frac{1}{K |\mathcal{D}| }\sum_{\left( x,y,l \right) }^{}{\sum_{ n}{
\indicatorFunctionSymbol_{ \lbrace n \geq l \rbrace }
\indicatorFunction[ n \ \mathrm{ mod }\ J =0 ]
\left(
    -\sum_{a }{
        \indicatorFunctionSymbol_{ \lbrace a = y \rbrace } \log f^{\pi} \left( \overline{V}^{(n)}(x), a \right)
    }
 \right)
}}.
$$
In RL, where the policies are learned through policy gradient, the loss function can be rewritten as
$$
\mathcal{L}\left( \theta \right) =\frac{1}{K |\mathcal{D}| }\sum_{\left( x,y,R,l \right) \in \mathcal{D}}^{}{\sum_{1\le n\le N}{
\indicatorFunctionSymbol_{ \lbrace n \geq l \rbrace }
\indicatorFunction[ n \ \mathrm{ mod }\ J =0 ]
\Big(
       - R \log f^{\pi} \left( \overline{V}^{(n)}(x), y \right)
 \Big),
}}$$
where $y$ is the excuted action, and $R$ the cumulative future reward.

\subsection{Translation Equivariance}

Although our dynamic latent transition kernel is input-dependent and breaks the conventional weight-sharing scheme, it still preserves translation equivariance—a desirable property in spatial planning tasks. In other words, shifting the input (e.g., the maze image) leads to a corresponding shift in the output. While this may seem counterintuitive at first, it is in fact a direct consequence of our design: the kernel $\overline{\mathsf{T}}^{\rm dyn} = f^{\overline{\mathsf{T}}}(x)$ is produced by a CNN, which is itself translation equivariant. As a result, the downstream value iteration process retains this property (see \Crefnop{eq_DT_VIN}).

\section{Maze Navigation: Additional Experimental Details and Results}\label{sec_maze_navigation}

\subsection{2D Maze Navigation: Experimental Details}\label{sec_experiment_detail_app_2D_maze}

\begin{figure}[h]
    \centering
    \def\wAOJDOJOWJOW{0.32}
    \centerline{
        \subfloat[$15 \times 15 $ Maze]{
        \includegraphics[width=\wAOJDOJOWJOW\linewidth]{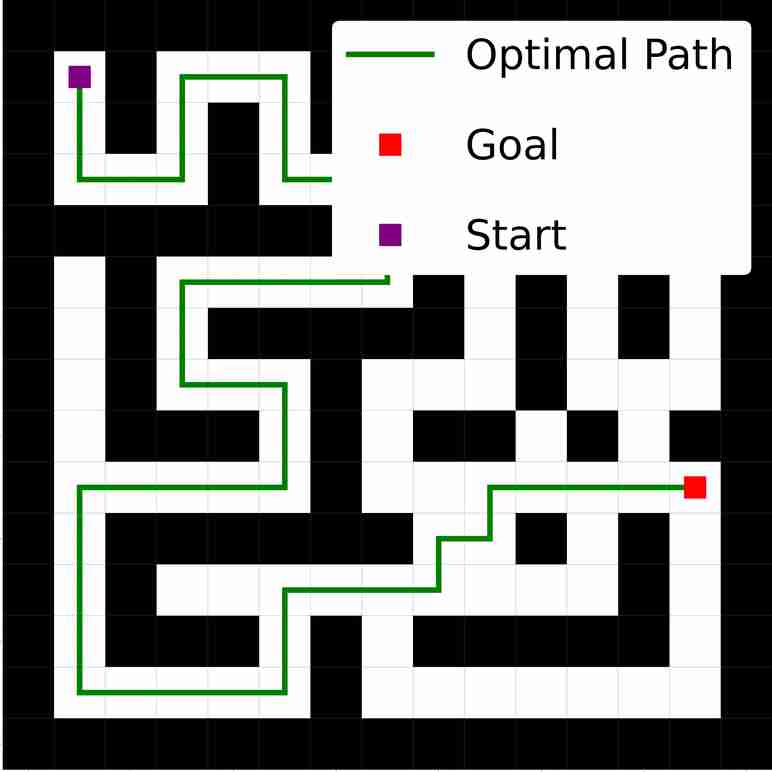}
        }
        \subfloat[$35 \times 35 $ Maze]{
        \includegraphics[width=\wAOJDOJOWJOW\linewidth]{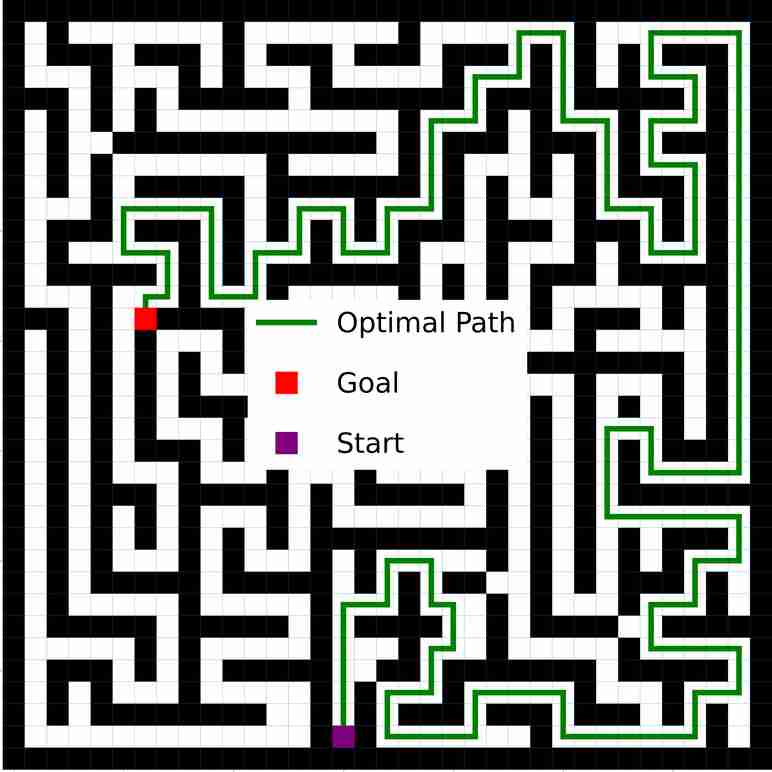}
        }
        \subfloat[$100 \times 100 $ Maze]{
            \includegraphics[width=\wAOJDOJOWJOW\linewidth]{figs/main/m100}
        }
    }
    \caption{
        Examples of the 2D maze navigation tasks.
        \label{fig_maze_various_size}
    }
\end{figure}

\Cref{fig_maze_various_size} shows some visualizations of some of the different 2D maze navigation tasks we experiment with.
Our experimental setup follows the guidelines established in the GPPN paper~\citep{lee2018gated}.
For these tasks, the datasets for training, validation, and testing comprise $25000$, $5000$, and $5000$ mazes, respectively.
Each maze features a goal position, with all reachable positions selected as potential starting points. Note that this setting, as done by GPPN, produces a distribution of mazes with non-uniform SPLs, which is skewed towards shorter SPLs.
\Cref{table_hyperparamters_2d_maze} shows the hyperparameters used by our method.
Note that, while DT-VIN consistently uses $3$ for the size of the latent transition kernel $F$ and $4$ for the size of the latent action space $|\latentASpace|$, other methods instead used their best-performing sizes from between $3$ and $5$, and between $4$ and $150$, respectively.

\begin{table}[ht]
\centering
\caption{2D Maze Navigation Hyperparameters
\label{table_hyperparamters_2d_maze}
}
\begin{tabular}{lc}
\toprule
Hyperparameter                     & Value                       \\
\midrule
Transition Mapping Module                  & A 1-layer CNN with $3\times 3$ kernel    \\
Reward Mapping Module                  & A 1-layer CNN with $1\times 1$ kernel      \\
Policy Mapping Module                  & A 1-layer FCN      \\
Latent Transition Kernel Size ($F$)                   & 3      \\
Latent Action Space Size ($|\latentASpace|$)                   & 4      \\
Optimizer                          & RMSprop                        \\
Learning Rate                      & 1e-3                        \\
Batch Size & 32 \\
\hline
Depth of Planning Module                   &
\multicolumn{1}{l}{\begin{tabular}[c]{@{}l@{}}
$15 \times 15 $ maze: $200$  \\
$35 \times 35 $ maze: $600$  \\
$100 \times 100 $ maze: $5000$
\end{tabular}}
\\
\bottomrule
\end{tabular}
\end{table}

\begin{table}[p]
\centering
\caption{Rover Navigation Hyperparameters
\label{table_hyperparamters_rover}
}
\begin{tabular}{lc}
\toprule
Hyperparameter                     & Value                       \\
\midrule
Transition Mapping Module                  & A $10$-layer CNN    \\
Reward Mapping Module                  &
\parbox{6cm}{
\centering
A $10$-layer CNN (sharing the first $8$ layers \\
with Transition Mapping Module)
}
\\
Policy Mapping Module                  & A 1-layer FCN      \\
Latent Transition Kernel Size ($F$)                   & 3      \\
Latent Action Space Size ($|\latentASpace|$)                   & 4      \\
Optimizer                          & RMSprop                        \\
Learning Rate                      & 1e-3                        \\
Batch Size & 32 \\
\hline
Depth of Planning Module                   &
\multicolumn{1}{l}{\begin{tabular}[c]{@{}l@{}}
$270 \times 270 $ : $50$  \\
$450 \times 450 $ : $100$  \\
$630 \times 630 $: $200$  \\
\end{tabular}}
\\
\bottomrule
\end{tabular}
\end{table}

\begin{table}[p]
\centering
\caption{3D ViZDoom Preprocessing Network
\label{appendix::table::ViZDoom_setting}}
\begin{tabular}{lc}
\toprule
Hyperparameter                     & Value                       \\ \midrule
Batch Size ($B$)                   & 32                          \\
Image Directions ($D$)             & 4                           \\
Image Channels ($C$)               & 3                           \\
Image Width ($W$)                  & 24                          \\
Image Height ($H$)                 & 32                          \\
Input Size                         & $(B, M, M, D, W, H, C)$     \\
Layer 1 (Convolution)              & $(3, 32, 8, 4, 1)$          \\
Layer 2 (Convolution)              & $(32, 64, 4, 2, 1)$         \\
Layer 3 (Linear)                   & $(384, 256)$                \\
Layer 4 (Convolution)              & $(1024, 64, 3, 1, 1)$       \\
Layer 5 (Convolution)              & $(64, 1, 3, 1, 1)$          \\
Output Size                        & $(B, M, M)$                 \\
Optimizer                          & Adam                        \\
Learning Rate                      & 1e-3                        \\
Betas                              & $(0.9, 0.999)$              \\
\bottomrule
\end{tabular}
\end{table}

\begin{table*}[t]
\centering
\caption{ The success rates for each method under tasks with different ranges of shortest path length.
For each algorithm, we choose the best result from a range of depths.
Specifically, for our DT-VIN, the optimal depth consistently corresponds to the maximum value in the range: $600$ for size $35$, and $5000$ for size $100$. For other compared methods, the optimal depth differs depending on the task. In the maze of size $100$, the optimal depth for all the baselines is $600$.
    For additional results, see \cref{fig__success_rate__algorithm__all_depths} in \cref{sec_different_depths}.
\label{table_success_rate}
}
\scalebox{0.65}{\begin{tabular}{c|ccc|ccc}
\hline
Maze Size            & \multicolumn{3}{c|}{$35 \times 35$}                                                & \multicolumn{3}{c}{$100 \times 100$}                                              \\ \hline
Ranges of Shortest Path Lengths & {[}1,100{]}                & {[}100, 200{]}            & {[}200, 300{]}            & {[}1,600{]}               & {[}600, 1200{]}           & {[}1200, 1800{]}          \\ \hline
VIN~\citep{tamar2016value}                  & $68.41_{\pm 6.25}$           & $0.0_{\pm 0.00}$           & $0.00_{\pm 0.00}$           & $45.05_{\pm 0.04}$          & $0.00_{\pm 0.00}$           & $0.00_{\pm 0.00}$           \\
GPPN~\citep{lee2018gated}                 & $95.71_{\pm 0.33}$           & $0.39_{\pm 0.27}$           & $0.00_{\pm 0.00}$           & $75.72_{\pm 0.64}$          & $0.00_{\pm 0.00}$           & $0.00_{\pm 0.00}$           \\
Highway VIN~\citep{wang2024highway}          & $90.67_{\pm 3.92}$           & $65.50_{\pm 5.59}$          & $54.40_{\pm 10.2}$          & $69.12_{\pm 0.02}$          & $0.00_{\pm 0.00}$           & $0.00_{\pm 0.00}$           \\
DT-VIN (ours)              & \bm{$100.00_{\pm 0.00}$} & \bm{$99.99_{\pm 0.01}$} & \bm{$99.77_{\pm 0.23}$} & \bm{$99.98_{\pm 0.00}$} & \bm{$99.56_{\pm 0.20}$} & \bm{$88.65_{\pm 4.76}$} \\ \hline
\end{tabular}
}
\end{table*}

\subsection{2D Maze Navigation with Different Depths of Models}\label{sec_different_depths}
\begin{figure*}[th]
    \centering
    \def\widthAJEJJDJ{0.3}
    \centerline{
  \subfloat[DT-VIN (ours) \vspace{1em}]{
    \label{fig_DTVIN_alldepths}
    \includegraphics[width=\widthAJEJJDJ\linewidth]{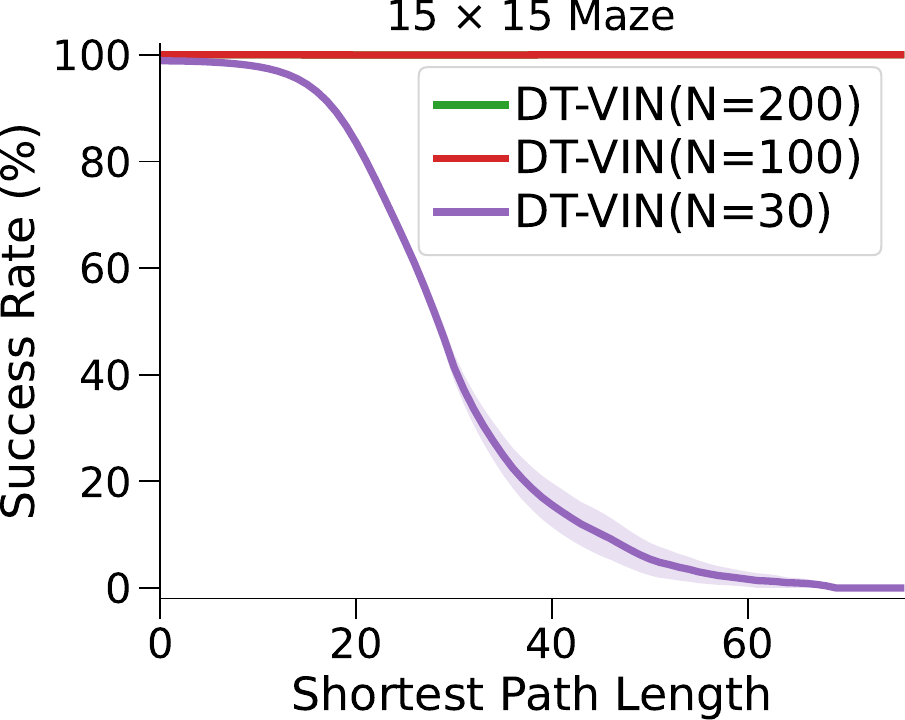}
    \includegraphics[width=\widthAJEJJDJ\linewidth]{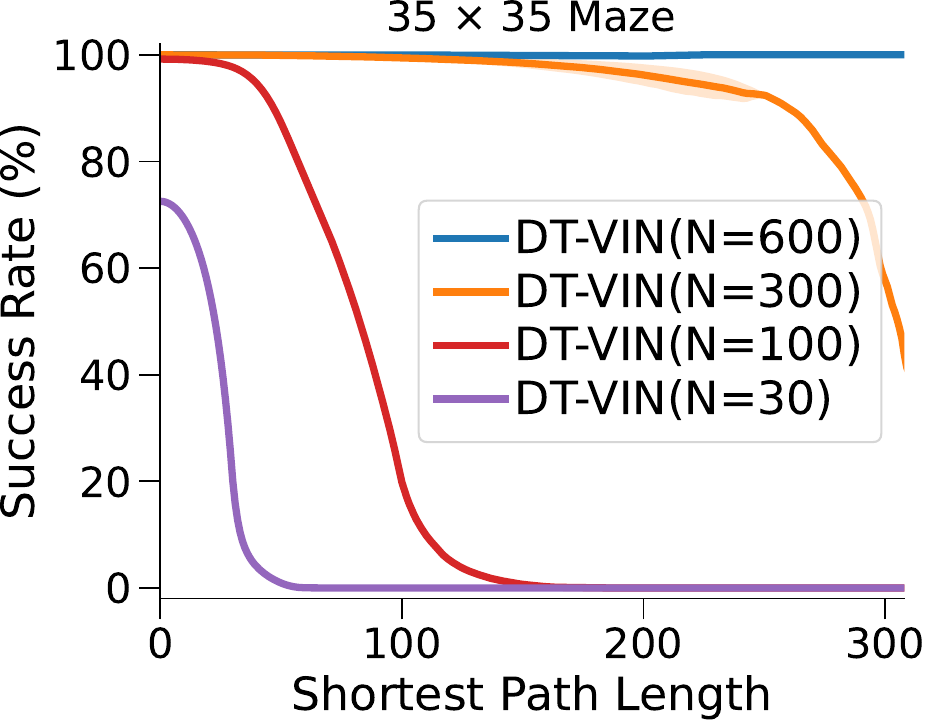}
    \includegraphics[width=\widthAJEJJDJ\linewidth]{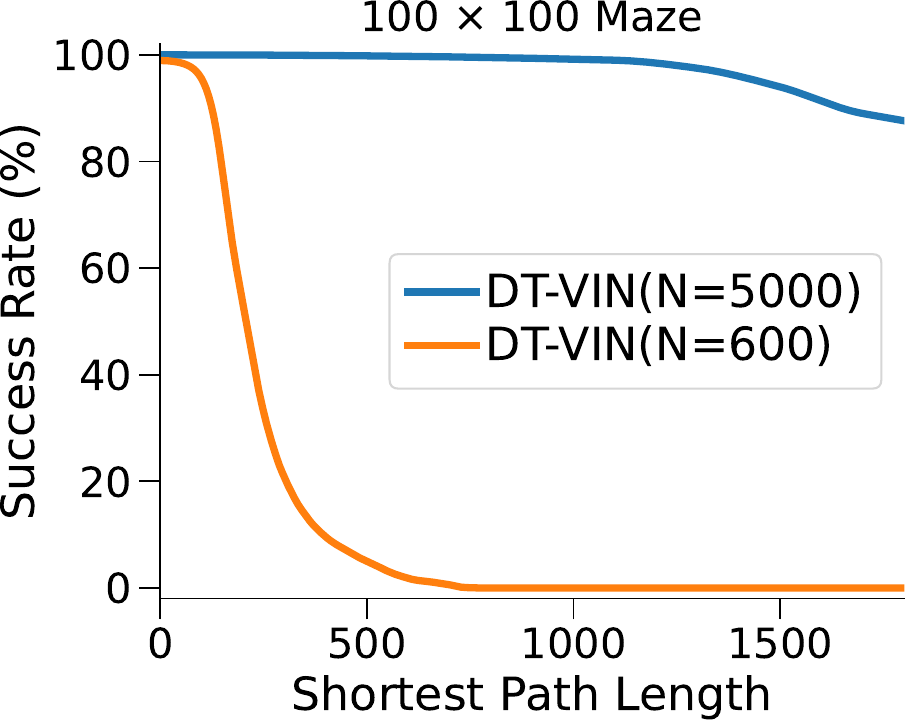}
 }
 }
 \centerline{
  \subfloat[VIN \vspace{1em}]{
    \includegraphics[width=\widthAJEJJDJ\linewidth]{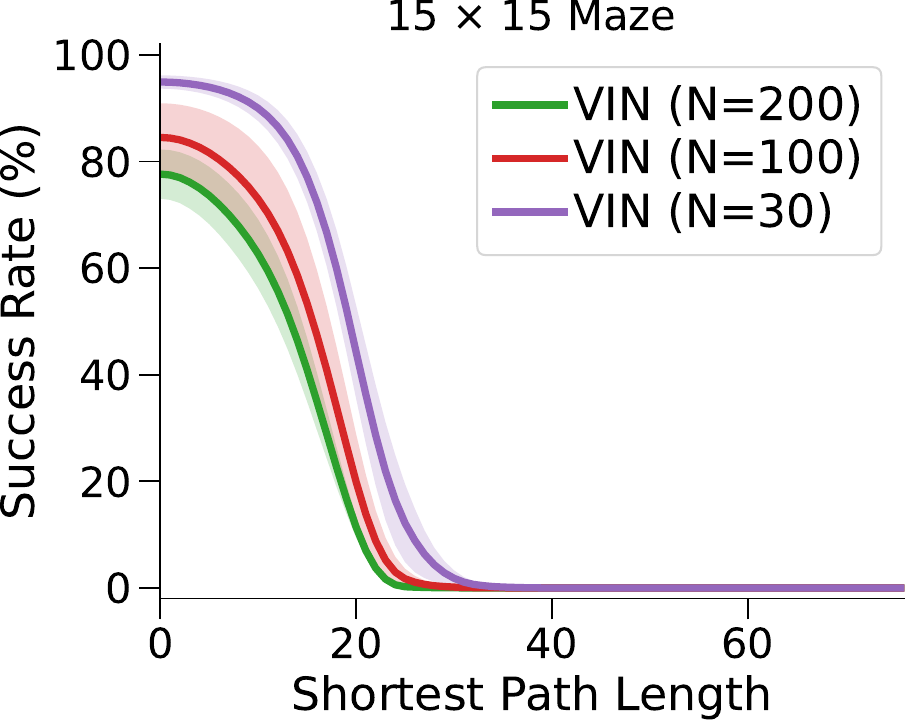}
    \includegraphics[width=\widthAJEJJDJ\linewidth]{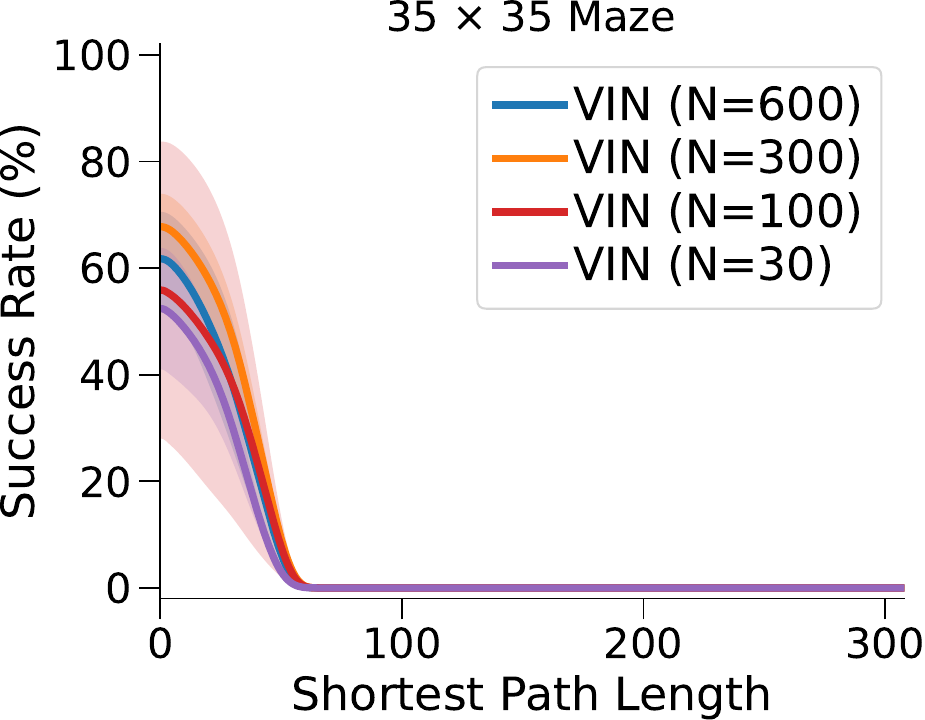}
    \includegraphics[width=\widthAJEJJDJ\linewidth]{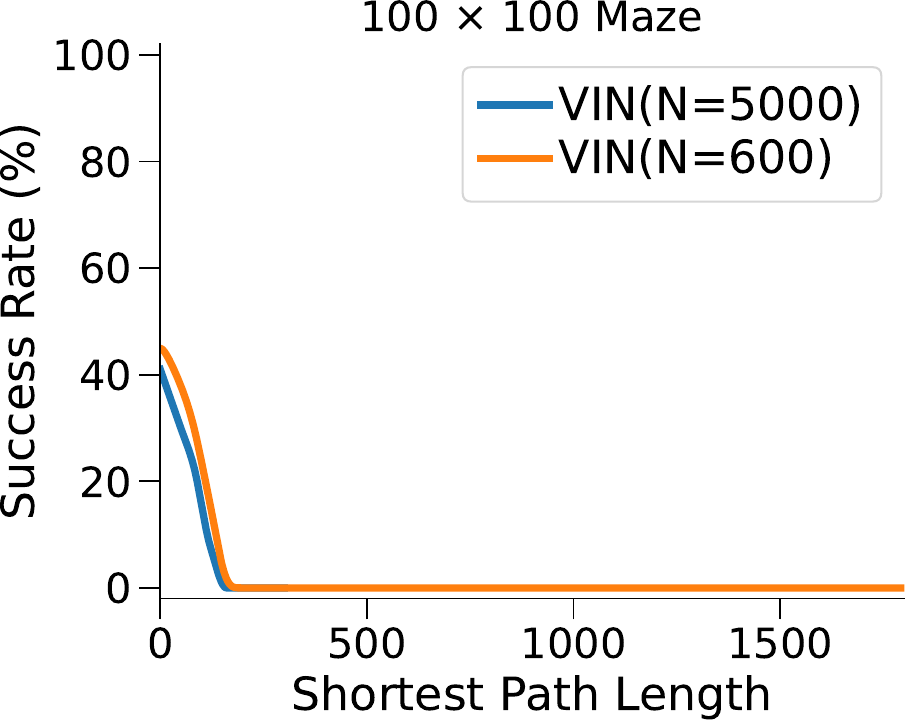}
 }
 }
 \centerline{
  \subfloat[GPPN]{
    \includegraphics[width=\widthAJEJJDJ\linewidth]{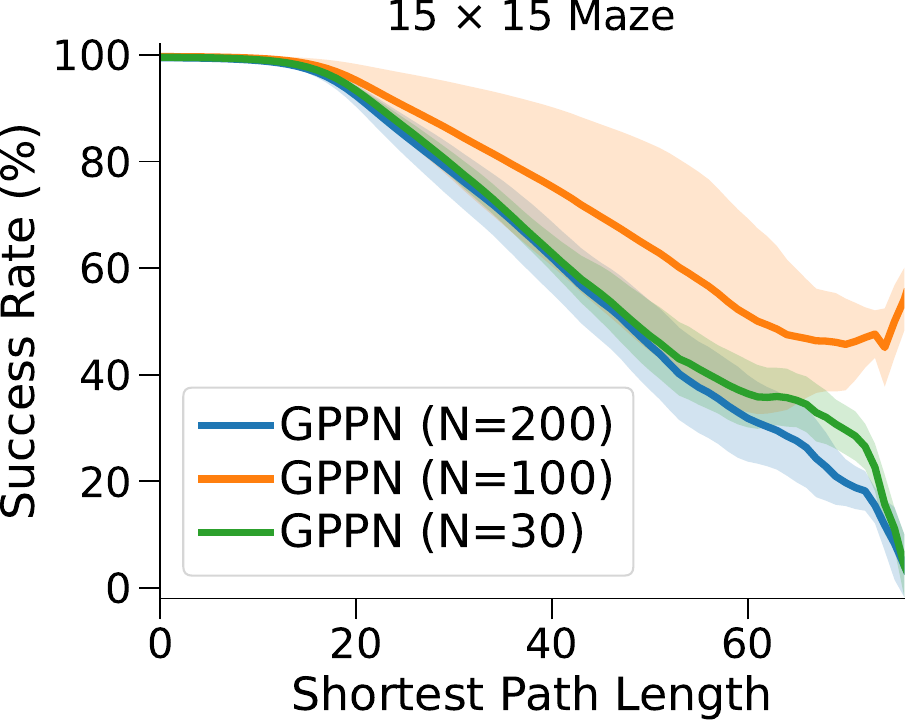}
    \includegraphics[width=\widthAJEJJDJ\linewidth]{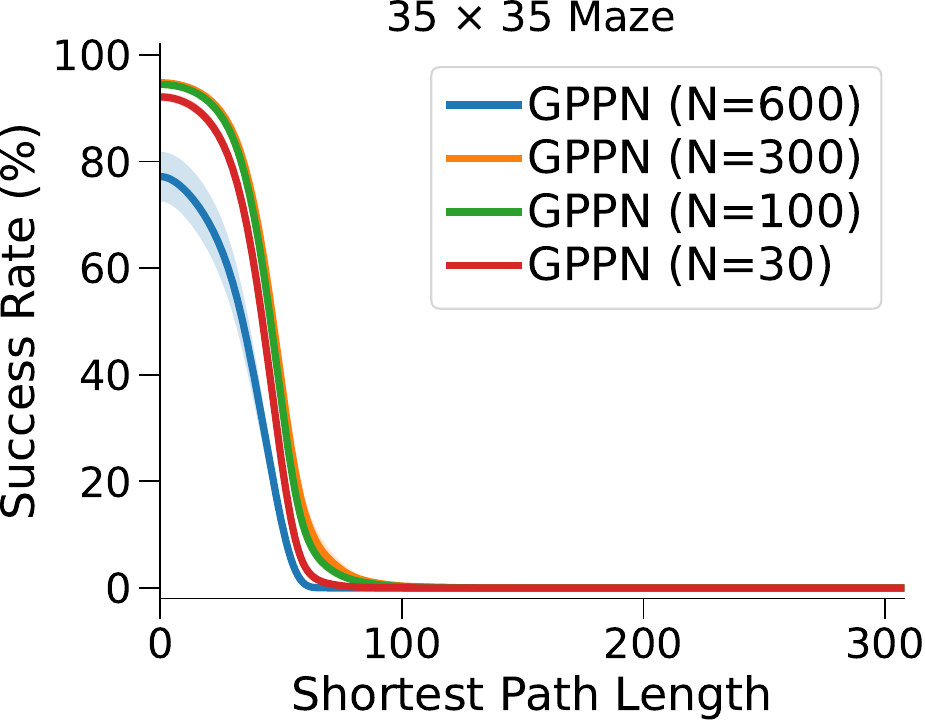}
    \includegraphics[width=\widthAJEJJDJ\linewidth]{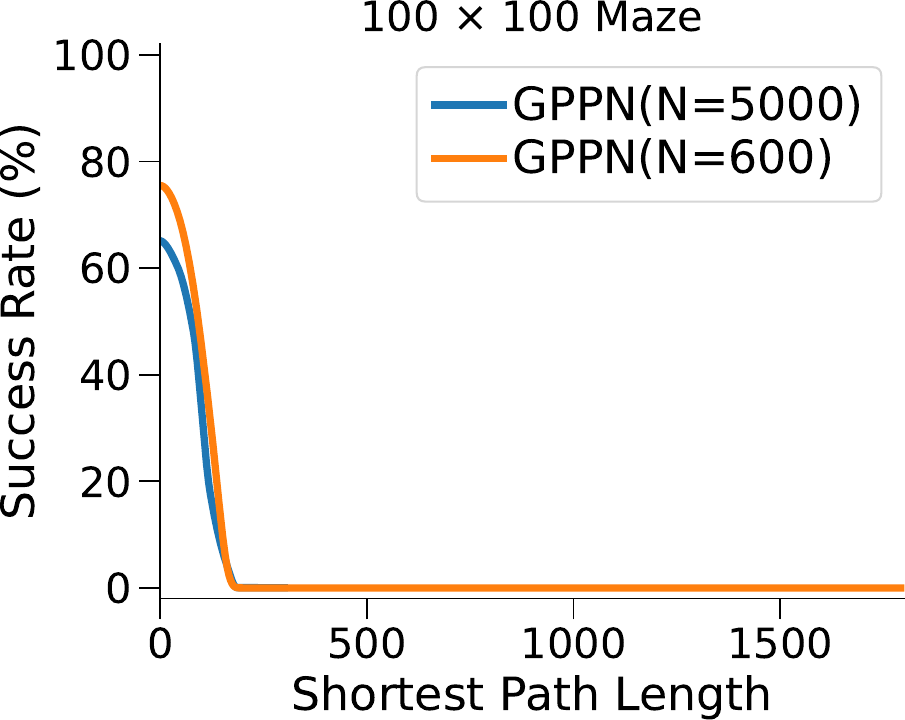}
}
 }

    \caption{
    The success rate of each method as a function of shortest path length and network depth. The green and red curves overlap in the first plot of \subref{fig_DTVIN_alldepths}.
    }
    \label{fig__success_rate__algorithm__all_depths}
\end{figure*}

\begin{figure*}[ht]
    \centering
    \def\widthAJEJJDJ{0.3}
    \centerline{
  \subfloat[DT-VIN (ours) \vspace{1em}]{
    \includegraphics[width=\widthAJEJJDJ\linewidth]{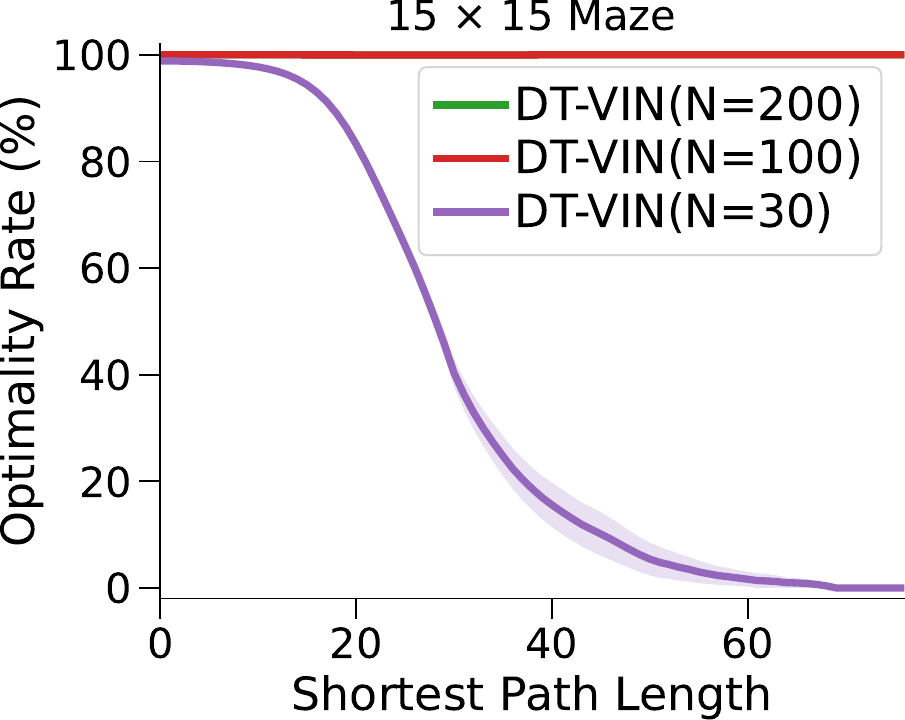}
    \includegraphics[width=\widthAJEJJDJ\linewidth]{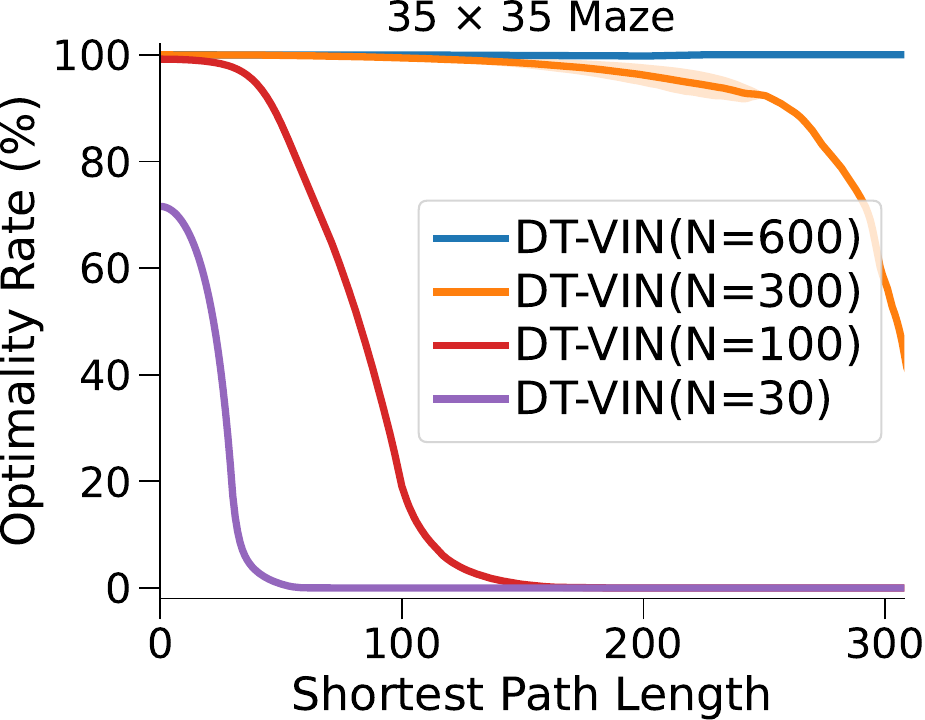}
    \includegraphics[width=\widthAJEJJDJ\linewidth]{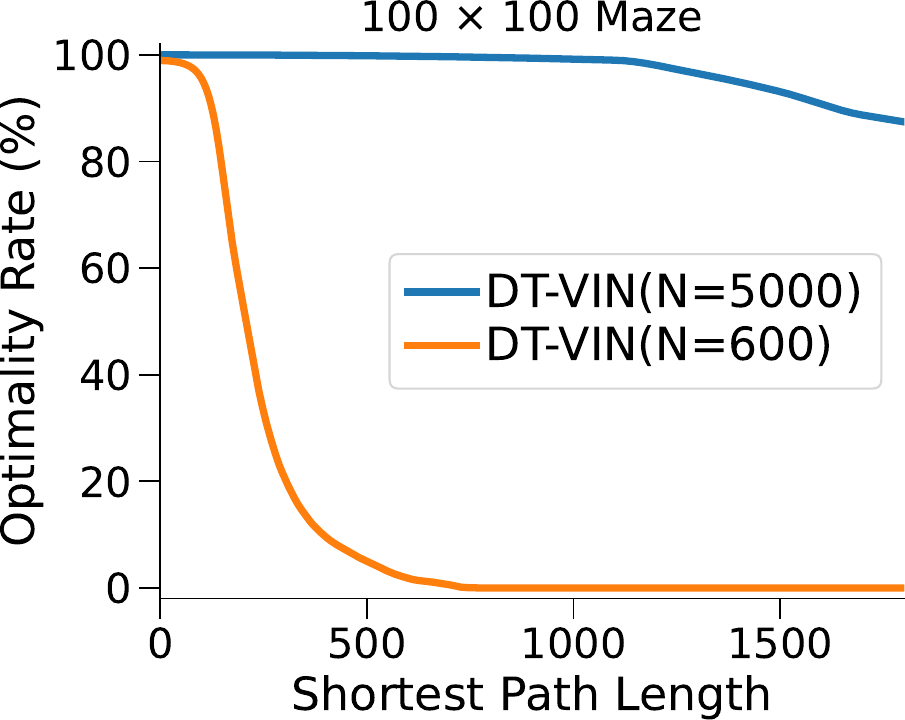}
 }
 }
 \centerline{
  \subfloat[VIN \vspace{1em}]{
    \includegraphics[width=\widthAJEJJDJ\linewidth]{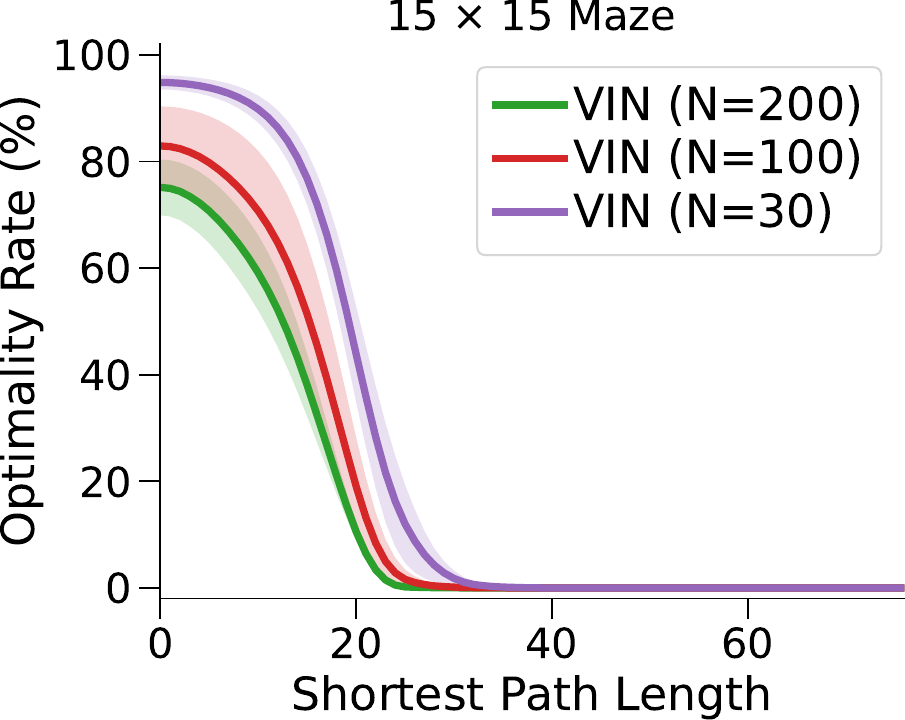}
    \includegraphics[width=\widthAJEJJDJ\linewidth]{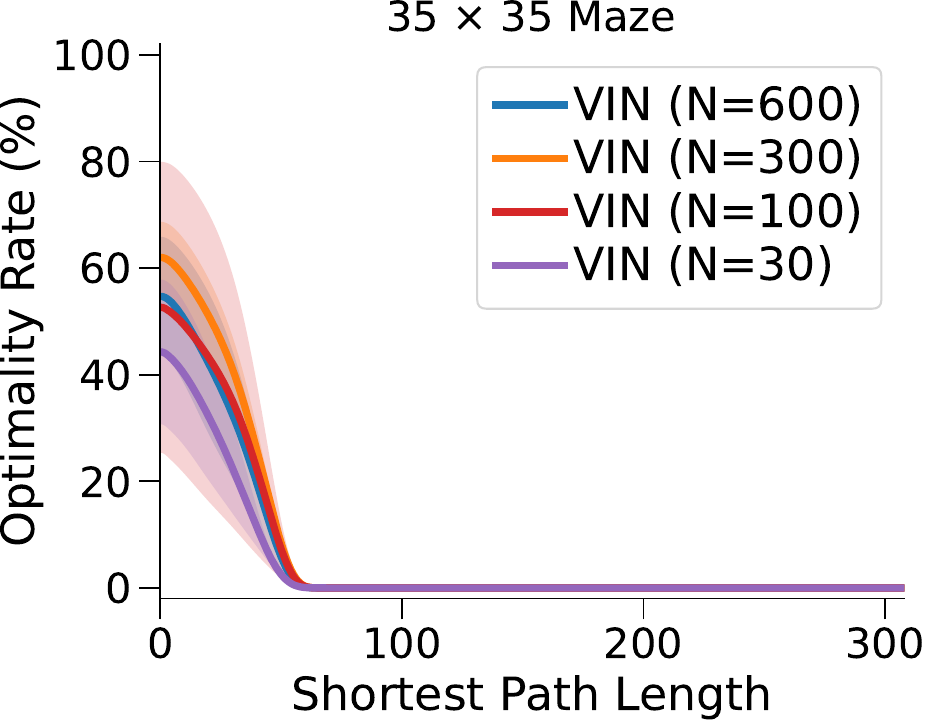}
    \includegraphics[width=\widthAJEJJDJ\linewidth]{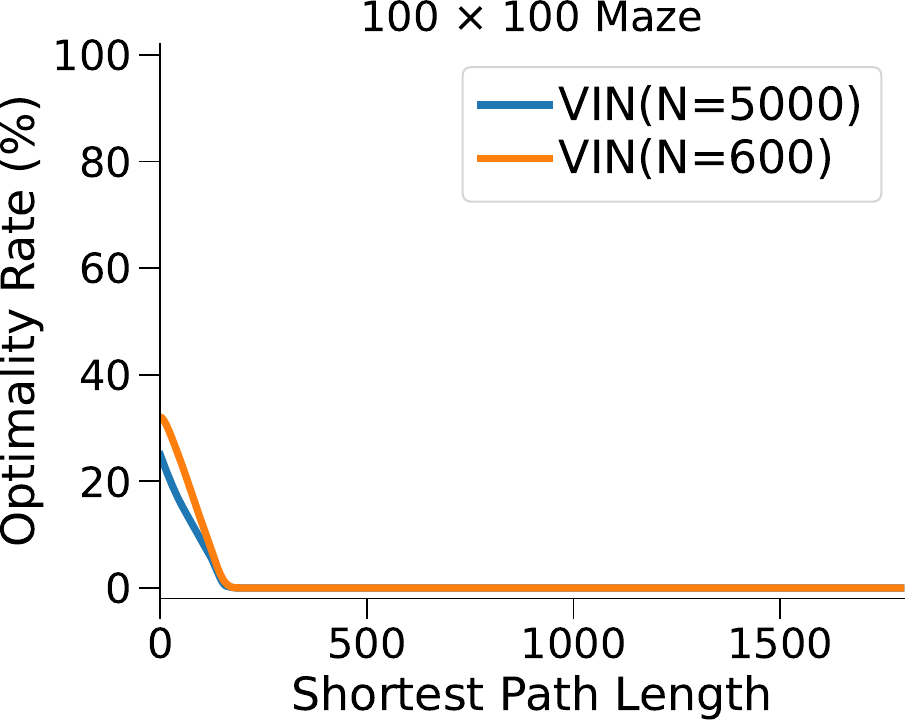}
 }
 }
 \centerline{
  \subfloat[GPPN]{
    \includegraphics[width=\widthAJEJJDJ\linewidth]{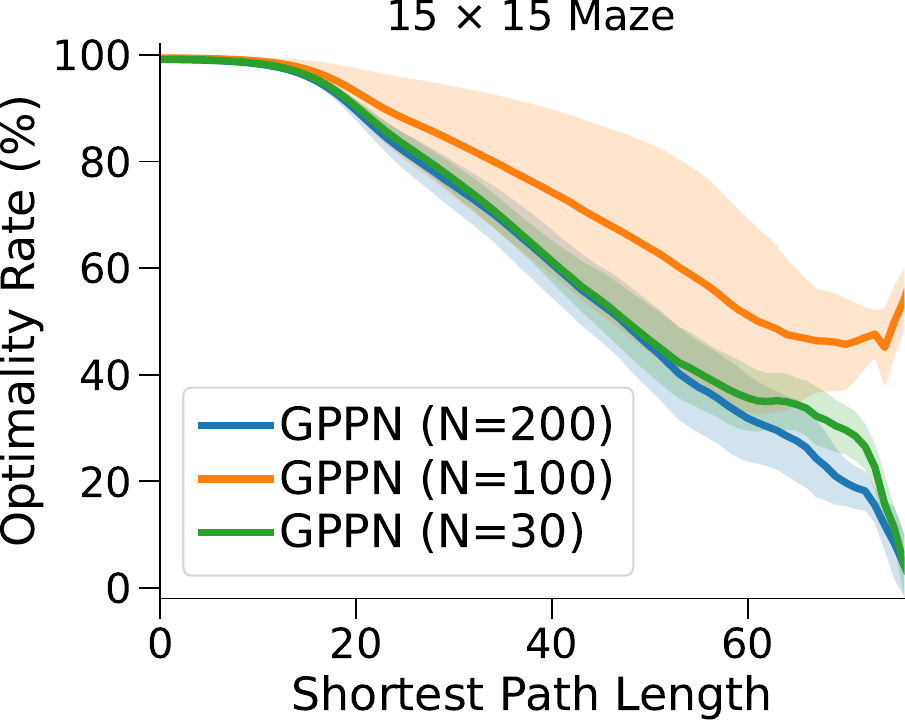}
    \includegraphics[width=\widthAJEJJDJ\linewidth]{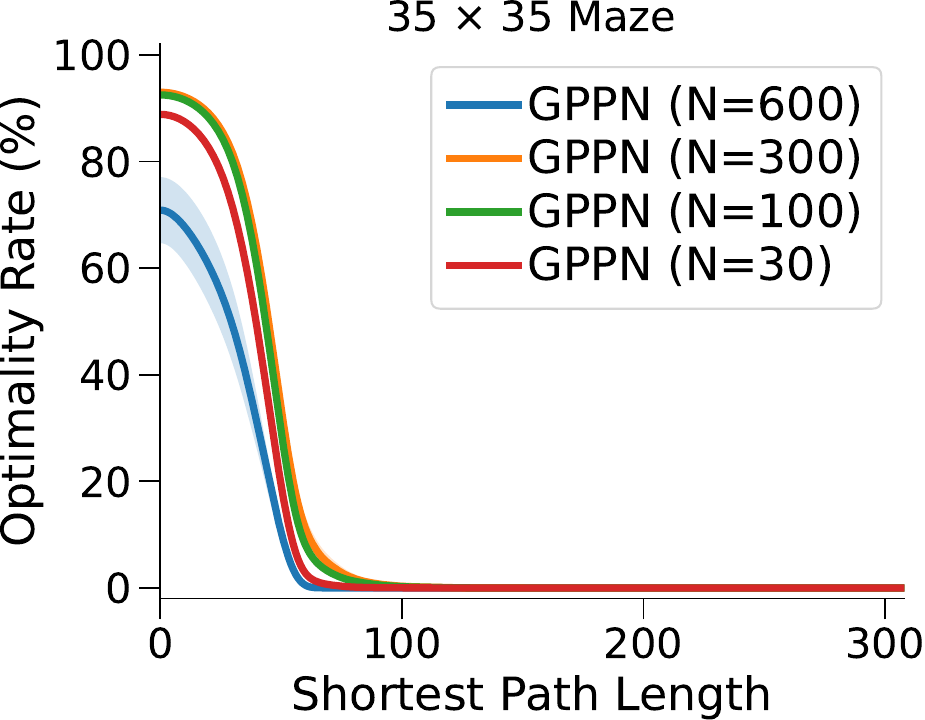}
    \includegraphics[width=\widthAJEJJDJ\linewidth]{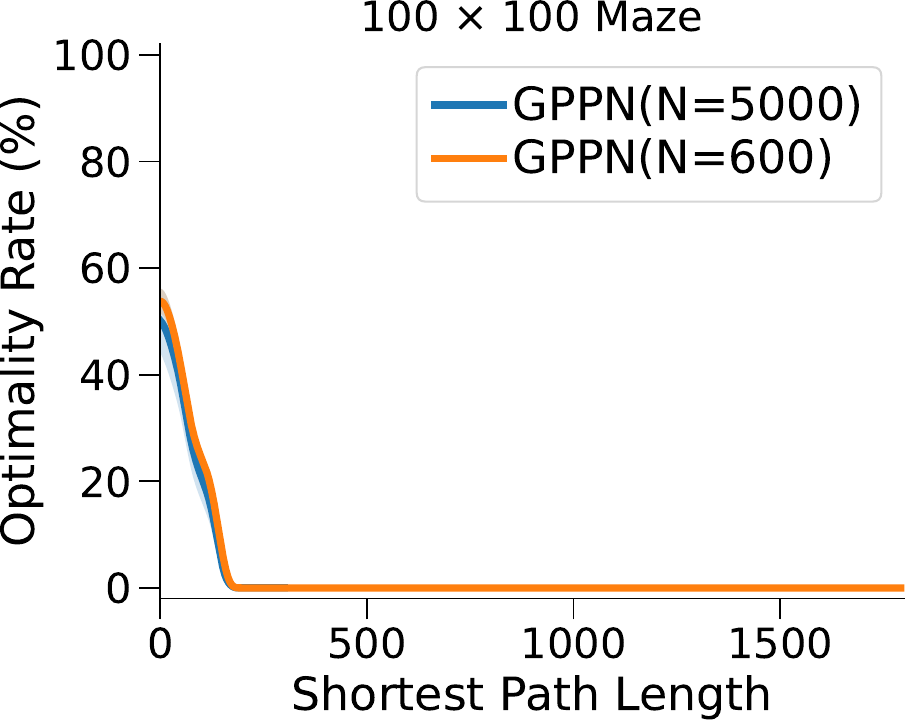}
}
 }
    \caption{
    The optimality rate of each method as a function of shortest path length and network depth. The green and red curves overlap in the top-left plot.
    }
    \label{fig__optimal_rate__algorithm__all_depths}
\end{figure*}

\cref{fig__success_rate__algorithm__all_depths} shows the success rate of all the algorithms on the $15 \times 15$, $35 \times 35$, $100 \times 100$ mazes as a function of the shortest path length and the depth of the network.
Similarly, \cref{fig__optimal_rate__algorithm__all_depths} shows the corresponding optimality rates.

\subsection{2D Maze Navigation with Different Transition Types}\label{sec_mechanism}

Following the GPPN paper \citep{lee2018gated},
we run additional experiments using different transition type: the \emph{Differential Drive} transition, where the agent can move forward along its orientation or rotate 90 degrees left or right, and the \emph{MOORE} transition, where the agent can relocate to any of the eight adjacent cells that comprise its Moore neighborhood.
As shown in \Cref{tab_success_rate_diffdrive} and \ref{tab_success_rate_moore}, DT-VIN consistently outperforms all the compared methods regardless of the transition type used.

\begin{table}[th]
\centering
\caption{The success rate (\%) for each method in $35\times 35$ 2D maze navigation with \emph{Differential Drive} transition type, where the agent can move forward along its orientation or rotate $90^\circ$ left or right.}
\begin{tabular}{cccc}
\hline
{Shortest Path Length} & {[1,150]}                      & {[150,300]}                    & {[300,500]}                   \\ \hline
VIN                           & $68.44_{\pm 3.12}$           & $0.03_{\pm 0.01}$            & $0.00_{\pm 0.00}$           \\
GPPN                          & $83.1_{\pm 1.23}$            & $0.31_{\pm 0.01}$            & $0.0_{\pm 0.0}$             \\
Highway VIN                   & $87.1_{\pm 3.73}$            & $57.1_{\pm 3.98}$            & $49.1_{\pm 8.73}$           \\
DT-VIN (ours)                 & $\mathbf{100.00_{\pm 0.00}}$ & $\mathbf{100.00_{\pm 0.00}}$ & $\mathbf{99.99_{\pm 0.01}}$ \\ \hline
\end{tabular}
\label{tab_success_rate_diffdrive}
\end{table}

\begin{table}[th]
\centering
\caption{The success rate for each method in $35\times 35$ 2D maze navigation with \emph{Moore} transition type, where the agent can relocate to any of the eight adjacent cells that comprise its Moore neighborhood.}
\begin{tabular}{cccc}
\toprule
{Shortest Path Length} & {[1,100]}                     & {[100,200]}                  & {[200,250]}                  \\ \midrule
VIN                           & $66.44_{\pm 3.21}$          & $0.00_{\pm 0.00}$          & $0.00_{\pm 0.00}$          \\
GPPN                          & $89.94_{\pm 1.31}$          & $0.04_{\pm 0.01}$          & $0.00_{\pm 0.00}$          \\
Highway VIN                   & $83.14_{\pm 2.21}$          & $37.1_{\pm 1.98}$          & $25.1_{\pm 3.28}$          \\
DT-VIN (ours)                 & $\mathbf{100.0_{\pm 0.00}}$ & $\mathbf{98.9_{\pm 0.72}}$ & $\mathbf{96.7_{\pm 1.23}}$ \\ \bottomrule
\end{tabular}
\label{tab_success_rate_moore}
\end{table}

\subsection{2D Maze Navigation with Controlled Noisy Maze}\label{sec_maze_noisy}

We evaluated the methods under controlled noise.
To emulate the prediction noise, we add Gaussian noise to the original maze and enforce the value within $(0,1)$ by $clip(maze+\epsilon, 0, 1)$, where $\epsilon$ is a noise sampled from Gaussian distribution. The results of this are below. DT-VIN seems more robust in handling noise than Highway VIN and VIN. GPPN shows robustness in the short term setting, but is not able to solve the task when long term planning is required.

\begin{table}[ht]
\centering
\caption{The success rate for each method in $35 \times 35 $ 2D maze navigation, with maze noise $\epsilon \sim \mathcal{N}(0,0.01)$.}
\begin{tabular}{lccc}
\toprule
Shortest Path Length & [1,100]            & [100,200]          & [200,300]          \\
\midrule
VIN                  & $66.41_{\pm 7.25}$ & $0.00_{\pm 0.00}$  & $0.00_{\pm 0.00}$  \\
GPPN                 & $91.71_{\pm 1.33}$ & $0.21_{\pm 0.27}$  & $0.00_{\pm 0.00}$  \\
Highway VIN          & $86.67_{\pm 4.92}$ & $57.50_{\pm 6.59}$ & $46.40_{\pm 11.2}$ \\
DT-VIN (ours)        & $\mathbf{99.21_{\pm 0.00}}$ & $\mathbf{98.17_{\pm 0.01}}$ & $\mathbf{97.77_{\pm 0.23}}$ \\
\bottomrule
\end{tabular}
\label{table_success_rate_noise_01}
\end{table}

\begin{table}[ht]
\centering
\caption{The success rate for each method in $35 \times 35 $ 2D maze navigation, with maze noise $\epsilon \sim \mathcal{N}(0,0.04)$.}
\begin{tabular}{lccc}
\toprule
Shortest Path Length & [1,100]            & [100,200]          & [200,300]          \\
\midrule
VIN                  & $59.11_{\pm 7.94}$ & $0.00_{\pm 0.00}$  & $0.00_{\pm 0.00}$  \\
GPPN                 & $87.43_{\pm 1.34}$ & $0.15_{\pm 0.48}$  & $0.00_{\pm 0.00}$  \\
Highway VIN          & $81.76_{\pm 4.74}$ & $49.5_{\pm 6.05}$  & $35.7_{\pm 11.3}$  \\
DT-VIN (ours)        & $\mathbf{95.17_{\pm 0.00}}$ & $\mathbf{92.29_{\pm 0.57}}$ & $\mathbf{91.99_{\pm 0.27}}$ \\
\bottomrule
\end{tabular}
\label{table_success_rate_noise_04}
\end{table}

\subsection{3D ViZDoom Navigation}\label{sec_3D_navigation}

\begin{figure}[ht]
    \def\height{0.3}
    \centering
    \subfloat[3D ViZDoom]{
        \includegraphics[height=\height\linewidth]{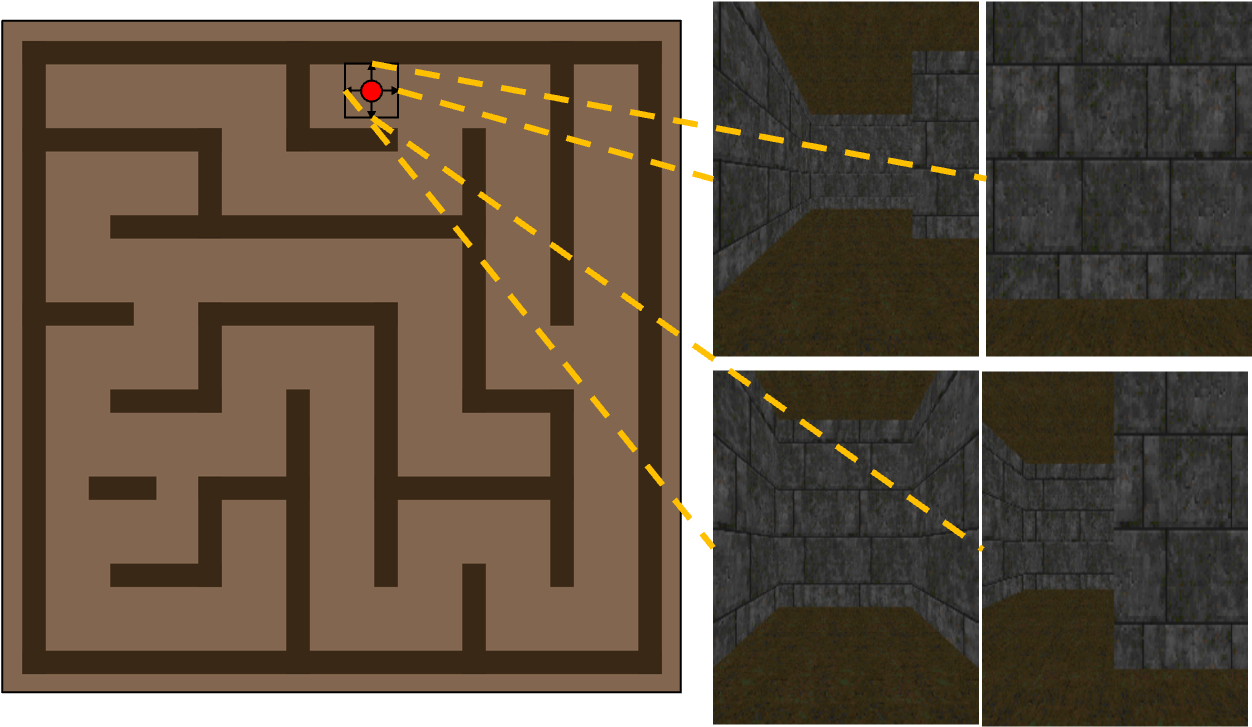}
        \label{fig_3D_ViZDoom}
    }
    \subfloat[Performance ]{
        \includegraphics[height=\height\linewidth]{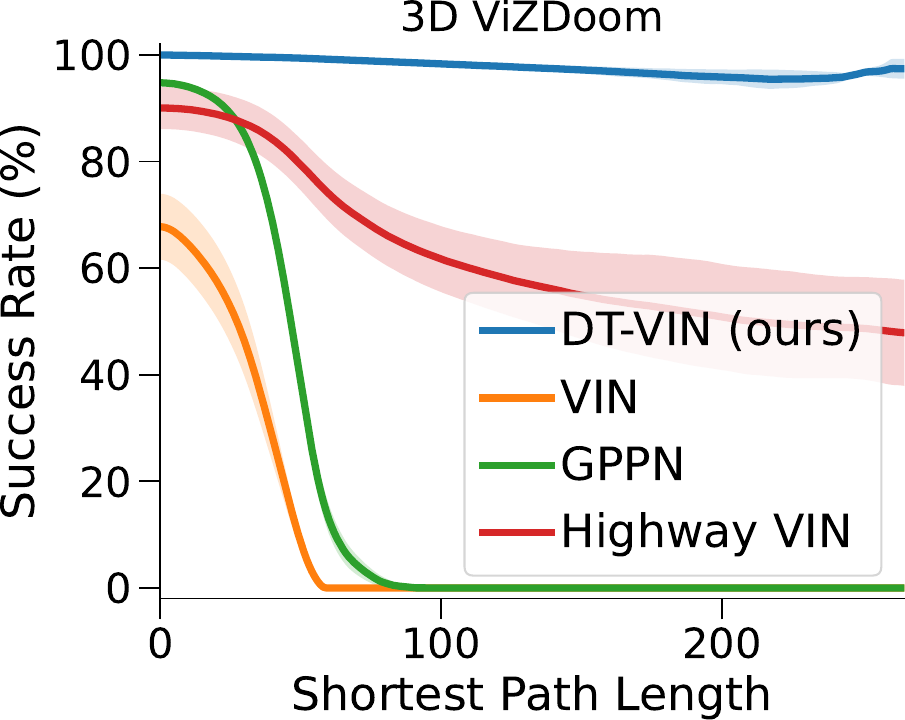}
        \label{fig_success_rate_3D_ViZDoom}
    }
    \caption{
        \subref{fig_3D_ViZDoom} an example of a ViZDoom 3D maze and the first-person view of the environment with each of the corresponding four orientations.
        \subref{fig_success_rate_3D_ViZDoom} the success rates of the algorithms over various SPLs.
    }
\end{figure}
Following the methodology of the GPPN paper, we test our method on 3D ViZDoom~\citep{Wydmuch2019ViZDoom} environments.
Here, instead of directly using the top-down 2D maze as in the previous experiments, we use the observation consists of RGB images capturing the first-person perspective of the environment, as illustrated in \cref{fig_3D_ViZDoom}.
Then, a CNN is trained to predict the maze map from the first-person observation.
The map is then given as input to the planning model, using the same architecture and hyperparameters as the 2D maze environments.
For each algorithm, we select the best result across the various network depths $N={30,100,300,600}$.
We find that the optimal depth for DT-VIN is $600$, for GPPN is $300$, for VIN is $300$, and for Highway VIN is $300$.
We evaluated the algorithm on 3D ViZDoom mazes with grid $35 \times 35$, where each cell in the grid corresponds to a $64 \times 64$ map unit area, the standard spatial measurement in the game engine.
\cref{fig_success_rate_3D_ViZDoom} shows the SRs.
Predictably, the performance of all the baselines decreases compared to the 2D maze environments due to the additional noise introduced by the predictions.
Here, DT-VIN outperforms all the methods compared to the task over all the various SPLs.

To be in line with previous work, we use a state representation preprocessing stage for the 3D ViZDoom environment similar to that used in the GPPN paper and others~\citep{lee2018gated, lample2017playing}.
In 3D ViZDoom, a maze is designed on a grid of  $M \times M$  cells. Each cell in this grid corresponds to an area of  $64 \times 64 $ map units within the 3D ViZDoom environment. The map unit is the basic measure of space used in the ViZDoom game engine to define distances and sizes.
Specifically, for each cell in the $M \times M$ 3D maze, the RGB first-person views for each of the four cardinal directions are given as state to a preprocessing network (see \cref{fig_3D_ViZDoom}).
This network then encodes this state and produces an $M \times M$ binary maze matrix.
The hyperparameters and exact specification of the network are given in \cref{appendix::table::ViZDoom_setting}.

\section{Continuous Control: Additional Experimental Details and Results}\label{sec_app_robotic_navigation}

\begin{figure}[ht]
    \centering
    \def\wAOJDOJOWJOW{0.5}
    \centerline{
        \subfloat[$35 \times 35 $ Maze]{
        \includegraphics[width=\wAOJDOJOWJOW\linewidth]{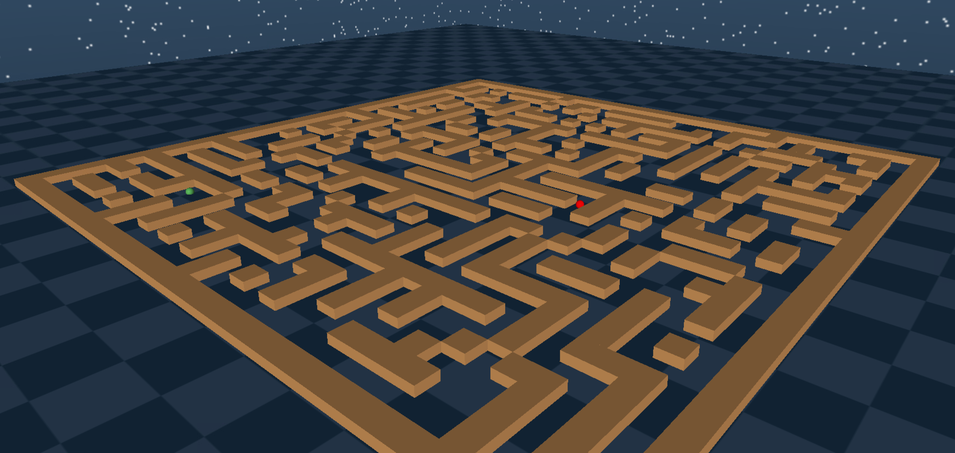}
        }
    }
    \centerline{
        \subfloat[$100 \times 100 $ Maze]{
            \includegraphics[width=\wAOJDOJOWJOW\linewidth]{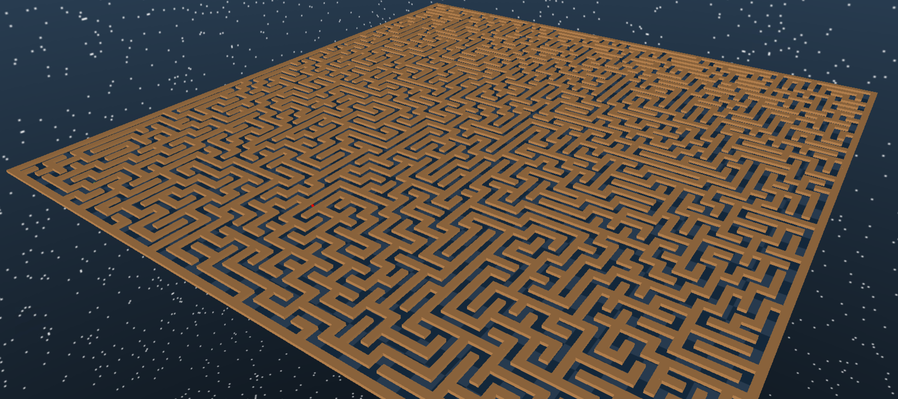}
        }
    }
    \caption{
        Some examples of the Point Maze tasks.
        \label{fig_robotic_various_size}
    }
\end{figure}

In this section, we present experimental details of continuous control, including Point Maze and Ant Maze.
\Cref{fig_robotic_various_size} demonstrates Point Maze with larger sizes $35\times 35$ and $100 \times 100$.
For continuous control tasks, the input includes a $4M \times 4M$ top-down view of the maze, agent and goal locations, positional values, and velocities of the agent's body parts.
Network architecture comprises 2-layer CNNs for reward and transition mapping, and a 3-layer fully connected network for policy mapping, which takes outputs from the planning module and additional state information to produce controlled actions.
The planning module in this architecture overlaps with that in the 2D maze navigation task. Therefore, we use pre-trained parameters from 2D maze navigation tasks.
\Cref{table_hyperparamters_robotic_control} lists the hyperparameters of 
DT-VIN for the continuous control tasks.
We train using PointMaze-Large and AntMaze-Large-Play datasets from D4RL \cite{fu2020d4rl} and pre-train on the $100 \times 100$ 2D maze.

\begin{table}[ht]
\centering
\caption{Continuous Control Hyperparameters
\label{table_hyperparamters_robotic_control}
}
\begin{tabular}{lc}
\toprule
Hyperparameter                     & Value                       \\
\midrule
Transition Mapping Module                  & A 2-layer CNN with $3\times 3$ kernel    \\
Reward Mapping Module                  & \parbox{6cm}{
\centering
A $2$-layer CNN (sharing the first layers \\
with Transition Mapping Module)
}      \\
Policy Mapping Module                  & A 3-layer FCN      \\
Latent Transition Kernel Size ($F$)                   & 3      \\
Latent Action Space Size ($|\latentASpace|$)                   & 4      \\
Optimizer                          & RMSprop                        \\
Learning Rate                      & 1e-3                        \\
Batch Size & 32 \\
\hline
Depth of Planning Module                   &
\multicolumn{1}{l}{\begin{tabular}[c]{@{}l@{}}
$35 \times 35 $ maze: $600$  \\
$100 \times 100 $ maze: $5000$
\end{tabular}}
\\
\bottomrule
\end{tabular}
\end{table}

\section{Rover Navigation: Additional Experimental Details and Results} \label{sec_Rover_Navigation_app}

\Cref{table_hyperparamters_rover} shows the hyperparameters of DT-VIN for the rover navigation tasks.
For the transition and reward mapping modules, we employ $10$-layer CNNs, with the first $8$ layers shared between them.

\section{Ablation Studies: Additional Experimental Details and Results}\label{sec_app_ablation_study}

\subsection{Ablation on the Jumping Hyperparameter $J$}

To reduce the computational complexity of highway loss, we apply adaptive highway loss only to layers $n$ satisfying the condition $n\ \mathrm{mod}\ J=0$, where $J$ is a hyperparameter set to 10 in our experiments. 
Here, the main idea is to build the highway connections at an interval $J$ --- for example, every 10 neural network layers, as in our setting where $J=10$.
Using this, the number of the loss terms will reduce to only $1/J$ of the original one.
\Cref{tab:performance_j} shows the magnitude of the computational speedup as a consequence of this implementation detail.

\begin{table}[h!]
\centering
\caption{Training time and success rate (\%) across different ranges of SPLs for DT-VIN with different $J$ values.}
\begin{tabular}{ccccc}
\toprule
& Wall-Clock Time (hours) & [1,100] & [100,200] & [200,300] \\ \midrule
$J=1$ & 37 & $\mathbf{100.00_{\pm 0.00}}$ & $\mathbf{99.99_{\pm 0.01}}$ & $\mathbf{99.78_{\pm 0.21}}$ \\
$J=10$ & 12.1 & $\mathbf{100.00_{\pm 0.00}}$ & $\mathbf{99.99_{\pm 0.01}}$ & $99.77_{\pm 0.23}$ \\
$J=50$ & 7.1 & $100.00_{\pm 0.00}$ & $99.98_{\pm 0.02}$ & $99.69_{\pm 0.27}$ \\ \bottomrule
\end{tabular}
\label{tab:performance_j}
\end{table}

\subsection{Ablation on the Choice of $l$}

The knowledge of the length $l$ of the expert path naturally exists in the imitation learning case.
However, for the case where such information is unknown, one can use either the length of non-expert data or some heuristic methods to estimate $l$ when the actual $l$ is completely unknown, e.g., using the distance between the start and the goal position.

To measure the effect of overestimation/underestimation, we experiment with various estimated values of the length of the shortest path $\widehat{l}$, which are $0, l/2, l, 2l, N$ (where $l$ is the actual length of the shortest path, $N$ is the depth of the planning module).
Second, to evaluate the case when the estimation of $l$ has variance, we use $l \cdot \max(\epsilon, 0)$ as the estimation, with $\epsilon$ sampled from a Gaussian distribution $\mathcal{N}(1, 1)$.
Third, we also assess two additional variants for estimating $l$:
(a) One variant that utilizes the length of non-expert trajectories for $l$;
(b) Another variant that estimates the shortest path length heuristically using the L1 distance between the start  $(x_s,y_s)$ and the goal $(x_g, y_g)$, i.e., $D=|x_s-x_g|+|y_s-y_g|$.

As indicated in Table \ref{tab:success_rates_l_hat}, both overestimation and underestimation lead to a performance degradation of no more than $7\%$. Additionally, we find that leveraging non-expert data or the heuristic L1 distance only yields a nearly $3\%$ degradation in performance, and performs better than the case when the optimal length is extremely overestimated/underestimated. These results imply that employing the information from non-expert data or heuristic estimation could be taken as an alternative when the optimal length is not available.

\begin{table}[h!]
\centering
\caption{Ablation study for using various estimated lengths of optimal paths for adaptive highway loss, under $35 \times 35 $ ViZDoom navigation.
The best results are highlighted.}
\label{tab:success_rates_l_hat}
\begin{tabular}{lccc}
\toprule
\textbf{Shortest Path Length} & \textbf{[1,100]} & \textbf{[100, 200]} & \textbf{[200,300]} \\
\midrule
$\widehat{l}=0$ (connected to all hidden layers) & $99.49_{\pm 0.35}$ & $94.51_{\pm 0.77}$ & $89.9_{\pm 3.76}$ \\
$\widehat{l}=l/2$ & $99.62_{\pm 0.91}$ & $96.21_{\pm 0.44}$ & $91.24_{\pm 1.68}$ \\
${\widehat{l}=l}$ & $\mathbf{99.67_{\pm 0.22}}$ & $\mathbf{97.92_{\pm 0.11}}$ & $\mathbf{96.41_{\pm 0.37}}$ \\
$\widehat{l}=2*l$ & $99.61_{\pm 0.18}$ & $96.29_{\pm 0.48}$ & $93.12_{\pm 0.73}$ \\
$\widehat{l}=N$ (connected to only last layer) & $99.52_{\pm 0.29}$ & $95.52_{\pm 0.86}$ & $91.12_{\pm 1.64}$ \\
$\widehat{l}=l \cdot \max(\epsilon,0), \epsilon \sim \mathcal{N}(1,1)$ & $99.62_{\pm 0.50}$ & $96.19_{\pm 0.15}$ & $93.21_{\pm 0.92}$ \\
$\widehat{l}=len($non-expert path$)$ & $99.62_{\pm 0.12}$ & $97.01_{\pm 0.69}$ & $93.31_{\pm 0.31}$ \\
$\widehat{l}=D$ (L1 distance) & $99.64_{\pm 0.49}$ & $96.92_{\pm 0.05}$ & $93.52_{\pm 0.87}$ \\
\bottomrule
\end{tabular}
\end{table}

\subsection{Ablation on the Size of Training Dataset}

Even in situations where data is rare, DT-VIN still outperforms compared methods.
As shown in Table \ref{tab_success_rate_reduced}, with only $50\%$ of the original dataset, DT-VIN greatly outperforms existing methods.
We also highlight the changes compared to the performance with a full-sized dataset in Table \ref{tab_success_rate_reduced}, where DT-VIN results in less than a $0.2\%$ degradation for tasks within the range [1, 100], while the best-performing comparison method, GPPN, incurs a degradation of nearly $12\%$.

\begin{table}[h!]
\centering
\caption{
The success rates and the changes in success rates for each method, using a dataset reduced to 50\% of the original size, are presented. These changes are compared to the results from the full-sized dataset, with more negative values indicating worse performance.
\label{tab_success_rate_reduced}
}
\begin{tabular}{cccc|ccc}
\toprule
& \multicolumn{3}{c|}{Success Rate} & \multicolumn{3}{c}{Changes}\\
Shortest Path Length & [1,100] & [100,200] & [200,300] & [1,100] & [100,200] & [200,300]  \\
\midrule
VIN                  & $32.41_{\pm 4.25}$ & $0.00_{\pm 0.00}$ & $0.00_{\pm 0.00}$ & $-36.00_{\pm 3.12}$ & $0.00_{\pm 0.00}$ & $0.00_{\pm 0.00}$ \\
GPPN                 & $83.11_{\pm 1.33}$ & $0.01_{\pm 0.01}$ & $0.00_{\pm 0.00}$ & $-12.60_{\pm 1.29}$ & $-0.38_{\pm 0.11}$ & $0.00_{\pm 0.00}$ \\
Highway VIN          & $45.41_{\pm 4.13}$ & $37.41_{\pm 3.25}$ & $21.41_{\pm 6.98}$ & $-45.26_{\pm 3.48}$ & $-28.09_{\pm 2.98}$ & $-32.99_{\pm 3.11}$ \\
DT-VIN (ours)        & $\mathbf{99.96_{\pm 0.01}}$ & $\mathbf{99.8_{\pm 0.12}}$ & $\mathbf{96.01_{\pm 0.32}}$ & $\mathbf{-0.04_{\pm 0.01}}$ & $\mathbf{-0.19_{\pm 0.04}}$ & $\mathbf{-3.76_{\pm 0.31}}$ \\
\bottomrule
\end{tabular}
\end{table}

\section{Scaling Experiments}\label{sec_scaling_app}

In this section, we examine how the methods scale in terms of computational complexity, dataset requirements, and model size relative to task scale.

\subsection{Required Computational Resources}
As we have discussed in \cref{sec_method_dynamic_transition}, our approaches only require $|\latentASpace| \times F^4 $ parameters, where we set $|\latentASpace|=4$ and $F=3$ in our experiments.
Table \ref{tab_compuational_complexity_100} shows the memory consumption and training time on NVIDIA A100 GPUs for DT-VIN and the compared methods when using 5000 layers and training for $90$ epochs on $100 \times 100$ maze.
As shown in the table below, DT-VIN incurs only slightly higher GPU memory and time costs compared to VIN, while being far more memory-efficient than GPPN and Highway VIN.
These results are generally consistent with those observed in the $35 \times 35$ 2D maze in Table \ref{tab_compuational_complexity}.

\begin{table}[h!]
\centering
\caption{
Computational complexity during training of each method, employing 600 layers and trained over 30 epochs, evaluated in a $35 \times 35$ 2D maze navigation.
\label{tab_compuational_complexity}
}
\begin{tabular}{lccc}
\toprule
Method         & GPU Memory (GB) & Wall-Clock Time (h) & GPU Hours (h) \\
\midrule
VIN      & 4.2   & 8.4 & 8.4  \\
GPPN     & 182   & 4.2 &  12.6  \\
Highway VIN & 41.3  & 14.3 & 14.3  \\
DT-VIN & 7.2  & 12.3 & 12.3  \\
\bottomrule
\end{tabular}
\end{table}

\begin{table}[h!]
\centering

\caption{ Computational complexity during training of each method using 5000 layers and training for 90 epochs, evaluated on a $100 \times 100$ 2D maze navigation.}
\label{tab_compuational_complexity_100}
\begin{adjustbox}{max width=\textwidth}
\begin{tabular}{lccc}
\hline
{Method} & {GPU Memory (GB)} & Wall-Clock Time (h) & GPU Hours (h) \\ \hline
VIN             & 35                       & 36         & 36                    \\
GPPN            & 710                      & 31             &  310               \\
Highway VIN     & 111                      & 112               & 224             \\
DT-VIN (ours)   & 61.2                      & 97                & 97             \\ \hline
\end{tabular}
\end{adjustbox}
\end{table}

\subsection{Required Size of Model}
In our experiments, the depth of the network required to solve the problem is close to linear with the number of planning steps required by the problem.
For maze size $M=15, 25, 35$, we test DT-VIN models at increasing depths in increments of 100 until the optimal performance is achieved.
For instance, for mazes of size $25\times25$, we assess depths of $100,200,300,400$.
For maze size $M=100$, we assess depths of $4000, 5000, 6000$.
As Table \ref{tab_minimal_depths} illustrates, the depth of the smallest network that can solve the task increases slightly more than linearly with the required planning steps.  Therefore, it might be feasible to continue increasing the network depth as the problems become more complex.

\begin{table}[h!]
\centering
\caption{
Minimal depths of DT-VIN model across various maze sizes.
\label{tab_minimal_depths}
}
\begin{tabular}{ccc}
\toprule
Maze Size & Longest Length of Optimal Path & Minimal Depth of DT-VIN \\
\midrule
15 & 80 & 100 \\
25 & 200 & 300 \\
35 & 300 & 500 \\
100 & 1800 & 5000 \\
\bottomrule
\end{tabular}
\end{table}

\subsection{Required Size of Training Data}

We evaluate the models on the mazes of size $15 \times 15$, $25 \times 25$, and $35 \times 35$, on increasingly larger numbers of expert steps (a step here is one transition from a cell to another cell).
Specifically, for each maze size, we look for the smallest $n$ such that, with $5000 \times n$ different mazes (with each maze having a number of expert trajectories and larger mazes having more expert trajectories), DT-VIN can achieve a $\geq 98\%$ success rate on the longest length planning problem for each maze. The results of this experiment are below (and we will add this to a camera-ready version of the paper). While this is a bit of a coarse result, comparing the length column to the expert data column, we see a growth rate that looks a more than linear, but still indicates a manageable degree of complexity increase. We would like to note that using more mazes in the training set with fewer expert trajectories for each maze might further increase the sample efficiency of our method and the baselines. Designing an appropriate curriculum for this could be an exciting area of future work.

The scale of the dataset needs to scale up with the complexity of the problem rather than the model depth.
Under the same scale of the problem, we didn't find that increasing model depth requires additional data. As shown in Table \ref{tab_success_rate}, without expanding the dataset, increasing the model depth does not reduce the performance.

\begin{table}[ht]
\centering
\caption{Minimal required size of training data for DT-VIN across various maze sizes.}
\begin{tabular}{cccc}
\toprule
Maze Size & Longest Length of Optimal Path & Required Number of Mazes & Required Amount of Expert Steps \\
\midrule
15 & 80 & 15K & 4M \\
25 & 200 & 15K & 24M \\
35 & 300 & 10K & 45M \\
\bottomrule
\end{tabular}
\label{tab:maze_requirements}
\end{table}

\begin{table}[ht]
\centering
\caption{
The success rate of DT-VIN across various model depths $N$, maintaining the same size as the original dataset.
\label{tab_success_rate}
}
\begin{tabular}{cccc}
\toprule
Shortest Path Length & [1,100] & [100,200] & [200,300] \\
\midrule
$N=300$  & $99.99_{\pm 0.01}$ & $99.81_{\pm 0.13}$ & $92.11_{\pm 1.31}$ \\
$N=600$  & $100.00_{\pm 0.00}$ & $99.99_{\pm 0.01}$ & $99.77_{\pm 0.23}$ \\
$N=1200$ & $100.00_{\pm 0.00}$ & $99.99_{\pm 0.01}$ & $99.81_{\pm 0.11}$ \\
\bottomrule
\end{tabular}
\end{table}

\end{document}

%% file: lib_my/A_ALL.tex
\newif\ifshowtmp
\showtmptrue

\ifshowtmp
    \usepackage[textsize=tiny]{todonotes}

    \newcommand{\todom}[1]{{ \todo{#1} }}
    \newcommand{\tocheck}[1]{{\textcolor{red}{(TO CHECK: #1)}}}

	\newcommand{\T}[1]{\textcolor{usccardinal}{#1}}

    \makeatletter
    \def\w#1 {\T{#1}~}
    \makeatother

\else
    \newcommand{\todom}[1]{}
    \newcommand{\tocheck}[1]{}
	
	\newcommand{\T}{}
	\newcommand{\todo}[1]{}

    \newcommand{\w}{}
\fi
\newcommand{\todoLessImportant}[1]{}
\newcommand{\removeForSavingSpace}[1]{}

\newif\ifPPT
\PPTtrue

\newif\ifshowstruct
\showstructtrue
\showstructfalse %

\ifshowstruct
	\newcommand{\tstructz}[1]{[#1] }
\else
	\newcommand{\tstructz}[1]{}
\fi

\label{packages}
\usepackage{mfirstuc}
\usepackage{graphicx}%
\usepackage{subcaption}
\usepackage{caption}

\usepackage{multirow}
\usepackage{multicol}
\usepackage{array}
\usepackage{booktabs}%

\usepackage{amsmath}
\usepackage{amssymb, graphicx, url, mathtools}%
\usepackage[overload]{empheq}

\usepackage{algorithmic,algorithm}   %

\usepackage{footnote}
\makesavenoteenv{tabular}
\makesavenoteenv{table}
\usepackage{pdfpages}
\usepackage{textcomp}
\usepackage{chngcntr}
\usepackage{verbatim}

\usepackage{xparse,calc}
\usepackage{afterpage}
\usepackage{comment}

\label{Command}

\counterwithin*{question}{section}

\makeatletter
\def\renewtheorem#1{%
  \expandafter\let\csname#1\endcsname\relax
  \expandafter\let\csname c@#1\endcsname\relax
  \gdef\renewtheorem@envname{#1}
  \renewtheorem@secpar
}
\def\renewtheorem@secpar{\@ifnextchar[{\renewtheorem@numberedlike}{\renewtheorem@nonumberedlike}}
\def\renewtheorem@numberedlike[#1]#2{\newtheorem{\renewtheorem@envname}[#1]{#2}}
\def\renewtheorem@nonumberedlike#1{  
\def\renewtheorem@caption{#1}
\edef\renewtheorem@nowithin{\noexpand\newtheorem{\renewtheorem@envname}{\renewtheorem@caption}}
\renewtheorem@thirdpar
}
\def\renewtheorem@thirdpar{\@ifnextchar[{\renewtheorem@within}{\renewtheorem@nowithin}}
\def\renewtheorem@within[#1]{\renewtheorem@nowithin[#1]}
\makeatother

\renewtheorem{theorem}{Theorem}
\renewtheorem{remark}{Remark}
\renewtheorem{definition}{Definition}
\renewtheorem{example}{Example}

\crefname{mdpexample}{MDP Example}{MDP Examples}

\crefname{todo}{TODO}{TODO}

\makeatletter
\newcommand{\dummylabel}[2]{\def\@currentlabel{#2}\label[td]{#1}}
\makeatother

\usepackage{thmtools} 
\usepackage{thm-restate}

\usepackage{xstring}

\newcolumntype{+}{>{\global\let\currentrowstyle\relax}}
\newcolumntype{Y}{>{\currentrowstyle}}\label{key}
\newcolumntype{-}{>{\currentrowstyle}}

\newcommand{\breakingcomma}{
  \begingroup\lccode`~=`,
  \lowercase{\endgroup\expandafter\def\expandafter~\expandafter{~\penalty0 }}
}

\DeclareFontFamily{U}{mathx}{\hyphenchar\font45}
\DeclareFontShape{U}{mathx}{m}{n}{<-> mathx10}{}
\DeclareSymbolFont{mathx}{U}{mathx}{m}{n}
\DeclareMathAccent{\widebar}{0}{mathx}{"73}

\renewcommand{\algorithmicrequire}{\textbf{Input:}}
\renewcommand{\algorithmicensure}{\textbf{Output:}}

\label{Settings}

\usepackage{makecell}

\Crefname{equation}{Eq.}{Eq.}

\newcommand{\Crefnop}[1]{\nameCref{#1} \ref{#1}}

\newcommand{\reftodo}[1]{\todo{cref}}

\newcommand{\Creftodo}[1]{\todo{cref}}
\newcommand{\creftodo}[1]{\todo{cref}}
\newcommand{\citetodo}[1]{\todo{cite}}

\newcommand{\Creftd}[1]{\todo{cref}}
\newcommand{\creftd}[1]{\todo{cref}}
\newcommand{\citetd}[1]{\todo{cite}}

\usepackage[export]{adjustbox}

\usepackage{xcolor}
\definecolor{bluemy}{HTML}{1f77b4}

\usepackage{enumitem}

    \usepackage{xr}
    \makeatletter
    \newcommand*{\addFileDependency}[1]{
      \typeout{(#1)}
      \@addtofilelist{#1}
      \IfFileExists{#1}{}{\typeout{No file #1.}}
    }
    \makeatother

\NewDocumentCommand{\indicatorFunction}{ O{} }{ {\mathbbm{1} }_{\{#1\}} } 
\NewDocumentCommand{\indicatorFunctionSymbol}{  }{ {\mathbbm{1} } }

\makeatletter
\def\widebreve{\mathpalette\wide@breve}
\def\wide@breve#1#2{\sbox\z@{$#1#2$}%
     \mathop{\vbox{\m@th\ialign{##\crcr
\kern0.08em\brevefill#1{0.8\wd\z@}\crcr\noalign{\nointerlineskip}%
                    $\hss#1#2\hss$\crcr}}}\limits}
\def\brevefill#1#2{$\m@th\sbox\tw@{$#1($}%
  \hss\resizebox{#2}{\wd\tw@}{\rotatebox[origin=c]{90}{\upshape(}}\hss$}
\makeatletter

\usepackage{bbm}

\usepackage{accents}

\usepackage[makeroom]{cancel}

\usepackage{pifont}
\newcommand{\cmark}{\ding{51}}  %
\newcommand{\xmark}{\ding{55}}  %

\definecolor{blueMy}{HTML}{1f77b4}
\definecolor{orangeMy}{HTML}{ff7f0e}
\definecolor{greenMy}{HTML}{2ca02c}
\definecolor{redMy}{HTML}{d62728}
\definecolor{purpleMy}{HTML}{9467bd}
\definecolor{brownMy}{HTML}{8c564b}
\definecolor{usccardinal}{rgb}{0.6, 0.0, 0.0}

\usepackage{makecell}

%% file: lib_my/A_CUR_HighwayVIN.tex
\NewDocumentCommand{\pooling}{ O{} O{} }{ \mathbb{P} }

\NewDocumentCommand{\conv}{ O{} O{} }{ \mathbb{C} }

\NewDocumentCommand{\stack}{ O{} O{} }{ {#1} \oplus {#2} }

\NewDocumentCommand{\linear}{ O{} O{} }{\mathbb{L} }

\def\policyIndex{ n_{p} }

\NewDocumentCommand{\policy}{ O{ ( n +n_b )} O{\policyIndex} }{ {\overline{\pi}}^{#1}_{#2} }

\NewDocumentCommand{\V}{ O{} O{} }{ \overline V^{(#1)}_{#2}  }
\NewDocumentCommand{\Q}{ O{} O{} }{ \overline Q^{(#1)}_{#2}  }
\NewDocumentCommand{\PiM}{ O{} O{} }{ \overline \Pi^{(#1)}_{#2}  }

\NewDocumentCommand{\characteristic}{ O{} }{\emph{(#1)}}

\NewDocumentCommand{\nstepReturn}{O{Q} O{n}   }{ {G}_{#1}^{#2} }
\NewDocumentCommand{\nstepReturnAvg}{O{Q} O{n}   }{ {\bar G}_{#1}^{#2} }

\def\lanpaper/{[Lan, 2020]}
\def\youngpaper/{[Young, 2019]}

\NewDocumentCommand{\K}{ O{k} }{ _{#1} }

\definecolor{expectationPolicyStep}{named}{black}
\definecolor{maxStepOneN}{named}{red}
\definecolor{maxStep}{named}{maxStepOneN}
\definecolor{maxPolicyStep}{named}{blue}
\definecolor{softmaxPolicyStep}{named}{black}
\definecolor{stepFirst}{named}{pink}
\NewDocumentCommand{\EPolicyStep}{ O{expectationPolicyStep}  }
{{
    \color{#1}
    \E_{ \pi \sim \policyDist,  n  \sim \stepDist }
}}

\NewDocumentCommand{\dataset}{ O{m} O{s,a} }{ {\cal D}_{#2}^{\ifthenelse{\equal{#1}{}}{}{(#1)}} }
\NewDocumentCommand{\traj}{ O{\pi} O{s_t,a_t} }{ {\tau}_{#2}^{#1} }
\NewDocumentCommand{\trajnew}{ O{\pi} O{s_t,a_t} O{n} }{ {\tau}_{#2}^{#3} \sim {#1} }

\def\step/{lookahead depth}
\def\steps/{lookahead depths}
\def\stepText/{lookahead depth}
\def\stepsText/{lookahead depths}

\def\stepSet{{\cal N}}
\def\stepSetMath/{ $\stepSet$ } 

\def\stepSetText/{set of \steps/}
\def\stepSetTextMath/{set of \steps/ $\stepSet$}
\def\StepSetText/{Set of \steps/}

\def\policyText/{lookahead policy}
\def\policiesText/{lookahead policies}
\def\policySetText/{set of lookahead policies}
\def\PolicySetText/{Set of lookahead policies}
\def\policyDistText/{selection distribution of behavioral policies }

\NewDocumentCommand{\policyDist}{ O{} O{\policySet} }{ {P}^{#1}_{#2} }

\NewDocumentCommand{\stepDist}{ O{} O{\stepSet} }{ {P}^{#1}_{#2} }

\NewDocumentCommand{\policySet}{ }{ {\Pi} }

\def\policySetMath/ { $\policySet$ }

\def\policySetTextMath/{\policySetText/ $\policySet$}
\def\policyDistTextMath/{\policyDistText/ $\policyDist$}

\NewDocumentCommand{\dist}{ O{} O{} }{ \mathcal{P}^{#1}_{#2} }

\NewDocumentCommand{\bellmanOperator}{ O{} O{} }{ \mathcal{B}^{#1}_{#2} }

\NewDocumentCommand{\bellmanOptOperator}{ O{} O{} }{ \mathcal{B}^{#1}_{#2} }
\NewDocumentCommand{\BOOperator}{ O{} O{} }{ \mathcal{B}^{#1}_{#2} }
\NewDocumentCommand{\BOOperatorEmbed}{ O{} O{} }{ \mathcal{B}^{#1}_{#2} }
\NewDocumentCommand{\BEOperator}{ O{\pi} O{} }{ \mathcal{B}^{#1}_{#2} }
\NewDocumentCommand{\BEOperatorEmbed}{ O{\overline{\pi}} O{} }{ \mathcal{B}^{#1}_{#2} }
\NewDocumentCommand{\multistepBOOperator}{ O{\policyDist} O{\stepDist} }{\overline{\mathcal{B}}^{#1}_{#2} }
\def\multistepBOOperatorText/{Multi-step BO Operator}
\def\multistepBOOperatorTextMath/{Multi-step BO Operator $\multistepBOOperator[][]$}

\def\multistepBEOperatorText/{Multi-step IS-based BE Operator}
\NewDocumentCommand{\multistepBEOperator}{ O{\policyDist} O{\stepDist} O{} }{ \overset{ {\tiny #3} }{ \widebreve{ \mathcal{B} } }^{#1}_{#2} }
\def\multistepBEOperatorTextMath/{\multistepBEOperatorText/ $\multistepBEOperator[][]$}

\def\stStepSet/{\text{s.t. }}

\def\highwayPrefix/{Highway}
\def\highwayPrefixFull/{Highway}

\def\highwayQLearning/{\highwayPrefix/ Q-Learning}
\def\highwayDQN/{\highwayPrefix/ DQN}

\def\softHighwayDQN/{Soft \highwayPrefix/ DQN}
\def\nstepHighwayDQN/{$n$-step \highwayPrefix/ DQN}

\def\highwayQLearningFull/{\highwayPrefixFull/ Q-Learning}
\def\highwayDQNFull/{\highwayPrefixFull/ DQN}

\def\highwayValueIteration/{\highwayPrefixFull/ Value Iteration}

\def\highwayOperatorText/{\highwayPrefix/ Operator}

\def\highwayGenOperatorText/{\highwayPrefix/ Generalized Operator}

\NewDocumentCommand{\highwayOperator}{ O{\policyDist} O{\stepDist} }{   {{{\mathcal{G}}}}^{ {#1} }_{ {#2} }   }

\NewDocumentCommand{\highwayOperatorQ}{ O{\policyDist} O{\stepDist} }{   {{\mathcal{G}}}^{ {#1} }_{ {#2} }   }
\NewDocumentCommand{\wronghighwayOperator}{ O{\policyDist} O{\stepDist} }{   {\bcancel{{\mathcal{G}}} }^{ {#1} }_{ {#2} }   }

\NewDocumentCommand{\highwayOptOperator}{ O{\policySet} O{\stepSet} }{   {\accentset{\mbox{\large\bfseries .}}{\mathcal{G}}}^{ {#1} }_{ {#2} }   }
\def\highwayOptOperatorText/{Highway Optimality Operator}
\def\highwayOptOperatorTextMath/{Highway Optimality Operator $\highwayOptOperator[][]$}
\def\highwayOperatorTextMath/{\highwayOperatorText/ $\highwayOperator[][]$}
\def\highwayGenOperatorTextMath/{\highwayGenOperatorText/ $\highwayOperator[][]$}

\def\highwaySoftmaxOperatorText/{\highwayPrefix/ Softmax Operator}
\NewDocumentCommand{\highwaySoftmaxOperator}{ O{\stepSet, \alpha} O{\policySet,\widetilde\alpha} }{   {{\mathcal{G}}}_{ {#1} }^{ {#2} }   }
\def\highwaySoftmaxOperatorTextMath/{\highwaySoftmaxOperatorText/ $\highwaySoftmaxOperatorMath$}

\def\highwayEquation/{\highwayPrefix/ Equation}
\def\highwayEquationFull/{\highwayPrefixFull/ Bellman Optimality Equation}

\def\QQQQ{}

\def\highwayQQPrefix/{\highwayPrefix/ $Q$}
\def\highwayQQPrefixFull/{\highwayPrefixFull/ $Q$}
\def\highwayQQOperatorText/{\highwayQQPrefix/ Operator}
\def\highwayQQOperatorTextFull/{\highwayQQPrefixFull/ Bellman Optimality Operator}
\def\highwayQQOperatorTextMath/{\highwayQQOperatorText/ $\highwayQQOperator$}

\def\highwayQQEquation/{\highwayQQPrefix/ Equation}
\def\highwayQQEquationFull/{\highwayQQPrefixFull/ Bellman Optimality Equation}

\def\greedyReturn/{\highwayPrefix/ return}
\def\GreedyReturn/{\HighwayPrefix Return}

\NewDocumentCommand{\distance}{ O{} }{ \mathbf{d}_{#1}^Q(s,a) } 
\NewDocumentCommand{\distanceAll}{ O{} }{ d_{#1}^{Q} } 

\NewDocumentCommand{\assumptionAn}{ O{n} O{\widehat\Pi} O{\pi^*} }{Assumption A$(#2,#1)$}

\NewDocumentCommand{\stateSetNStep}{ O{s} O{\pi} O{n} }{ \mathcal{S}_{#1, #3}^{#2} }

\NewDocumentCommand{\ourExpOpt}{ O{\widehat{\Pi}} O{\mathcal{N}} }{ \overline{\mathcal{G}}_{#1}^{#2} }
\def\ourExpOptText/{Expectation Highway Operator}

\def\oneStepOperatorText/{BO operator}
\NewDocumentCommand{\oneStepOperatorMath}{ O{}  }{ \multiStepOperatorMath[#1][] }
\NewDocumentCommand{\oneStepOperator}{ O{}  }{ \multiStepOperatorMath[#1][] }
\def\oneStepOperatorTextMath/{\oneStepOperatorText/ $\oneStepOperatorMath$}

\def\bellmanOptText/{Bellman Optimality Operator}
\NewDocumentCommand{\bellmanOpt}{ O{}  }{ \mathcal{B}^{#1} }
\def\bellmanExpText/{Bellman Expectation Operator}
\NewDocumentCommand{\bellmanExp}{ O{#1}  }{ \bellmanOpt[#1] }
\NewDocumentCommand{\bellmanIS}{ O{} O{}  }{{\breve{\mathcal{B}}}^{#1}_{#2} }

\def\oneStepQQOperatorText/{Bellman Optimality Operator}
\NewDocumentCommand{\oneStepQQOperator}{ O{}  }{ \multiStepQQOperator[#1][] }
\NewDocumentCommand{\oneStepQQOperatorMath}{ O{}  }{ \multiStepQQOperator[#1][] }
\def\oneStepQQOperatorTextMath/{\oneStepQQOperatorText/ $\oneStepQQOperatorMath$}

\NewDocumentCommand{\multiStepOnPolicyOperator}{ O{N} O{\pi} }{ {\mathcal{B}}^{{#2}}_{#1} }
\NewDocumentCommand{\multiStepOnPolicyOperatorMath}{ O{N} O{\pi} }{ \multiStepOnPolicyOperator[#1][#2] }

\def\multiStepOnPolicyOperatorText/{Multi-Step Bellman Expectation Operator}
\def\multiStepOnPolicyOperatorTextMath/{ \multiStepOnPolicyOperatorText/ $\multiStepOnPolicyOperatorMath$ }

\NewDocumentCommand{\multiStepOnPolicyQQOperatorMath}{ O{N} O{\pi} }{ \QQQQ{\mathcal{B}}^{{#2}}_{#1} }

\NewDocumentCommand{\multiStepOffPolicyOperatorMath}{ O{N} O{\pi} O{\policyDist} }{ {\mathcal{B}}^{{#2}}_{#1} }
\NewDocumentCommand{\multiStepOffPolicyQQOperatorMath}{ O{N} O{\pi} O{\policySet} }{ \QQQQ{\mathcal{B}}^{{#2,#3}}_{#1} }

\NewDocumentCommand{\nreturnWeight}{ O{n} O{\pi}  }{ w_{#1}^{#2} }

\def\multiStepOperatorText/{Multi-step Bellman Optimality operator}
\NewDocumentCommand{\multiStepOperator}{ O{N} O{\policySet} }{ \mathcal{B}^{{#1}}_{#2} }
\NewDocumentCommand{\multiStepOperatorMath}{ O{N} O{\policySet} }{ \multiStepOperator[#1][#2] }
\def\multiStepOperatorTextMath/{\multiStepOperatorText/ $\multiStepOperatorMath$}

\def\multiStepQQOperatorText/{\multiStepOperatorText/}
\NewDocumentCommand{\multiStepQQOperator}{ O{N} O{\policyDist} }{ \QQQQ{\mathcal{B}}^{{#1}}_{#2} }
\NewDocumentCommand{\multiStepQQOperatorMath}{ O{N} O{\pi} }{ \multiStepQQOperator[#2][#1] }
\def\multiStepQQOperatorTextMath/{\multiStepOperatorText/ $\multiStepQQOperatorMath$}

\def\highwayPolicyIteration/{\highwayPrefixFull/ Policy Iteration}

\def\HighwayEvaluatePrefix/{Highway}
\def\highwayEvaluatePrefix/{Highway}
\def\highwayEvaluatePrefixFull/{Highway}
\def\highwayEvaluateValueIteration/{\highwayPrefixFull/ Value Iteration}
\def\highwayEvaluateEquationFull/{\highwayPrefixFull/ Recombined Equation}

\def\highwayEvaluateEquation/{\highwayPrefix/ Recombined Equation}
\def\highwayEvaluateOperatorText/{\highwayPrefix/ Recombined Operator}
\def\highwayEvaluateOperatorTextFull/{\highwayPrefixFull/ Recombined Operator}
\NewDocumentCommand{\highwayEvaluateOperatorMath}{ O{N} O{\policySet} }{ \mathcal{G}^{ {#1} }_{ {#2} } }

\def\recombinedPolicySetText/{set of recombined behavioral policies}

\NewDocumentCommand{\highwayRecombinationOperatorMath}{ O{N} }{ \mathcal{G}^{{#1}} }

\definecolor{QPiFixedPoint}{named}{usccardinal}

\def\nrandomseed/{5}

%% file: lib_my/A_CUR.tex
\def\latentSSpace{ \overline{\mathcal{S}} }
\def\latentASpace{ \overline{\mathcal{A}} }

\def\latentS{ \overline{s} }

\def\latentReward{ \overline{\mathsf{R}} }
\def\latentTransition{ \overline{\mathsf{T}} }

%% file: lib_my/math_commands.tex
\usepackage{amsmath,amsfonts,bm}

\def\eqref#1{equation~\ref{#1}}

\def\1{\bm{1}}

\DeclareMathAlphabet{\mathsfit}{\encodingdefault}{\sfdefault}{m}{sl}
\SetMathAlphabet{\mathsfit}{bold}{\encodingdefault}{\sfdefault}{bx}{n}

\newcommand{\E}{\mathbb{E}}

\newcommand{\R}{\mathbb{R}}

\NewDocumentCommand{\rangeInt}{ O{a} O{b} }{ [ #1..#2 ] }

%% file: main.bbl
\begin{thebibliography}{48}
\providecommand{\natexlab}[1]{#1}
\providecommand{\url}[1]{\texttt{#1}}
\expandafter\ifx\csname urlstyle\endcsname\relax
  \providecommand{\doi}[1]{doi: #1}\else
  \providecommand{\doi}{doi: \begingroup \urlstyle{rm}\Url}\fi

\bibitem[Achiam et~al.(2023)Achiam, Adler, Agarwal, Ahmad, Akkaya, Aleman,
  Almeida, Altenschmidt, Altman, Anadkat, et~al.]{achiam2023gpt}
Achiam, J., Adler, S., Agarwal, S., Ahmad, L., Akkaya, I., Aleman, F.~L.,
  Almeida, D., Altenschmidt, J., Altman, S., Anadkat, S., et~al.
\newblock Gpt-4 technical report.
\newblock \emph{arXiv preprint arXiv:2303.08774}, 2023.

\bibitem[Agarwal et~al.(2019)Agarwal, Jiang, Kakade, and
  Sun]{agarwal2019reinforcement}
Agarwal, A., Jiang, N., Kakade, S.~M., and Sun, W.
\newblock Reinforcement learning: Theory and algorithms.
\newblock \emph{CS Dept., UW Seattle, Seattle, WA, USA, Tech. Rep}, 32, 2019.

\bibitem[Bain \& Sammut(1995)Bain and Sammut]{bain1995framework}
Bain, M. and Sammut, C.
\newblock A framework for behavioural cloning.
\newblock In \emph{Machine Intelligence 15}, pp.\  103--129, 1995.

\bibitem[Bellman(1957)]{bellman1957markovian}
Bellman, R.~E.
\newblock A markovian decision process.
\newblock \emph{Journal of Mathematics and Mechanics}, 6\penalty0 (5):\penalty0
  679--684, 1957.
\newblock URL \url{http://www.jstor.org/stable/24900506}.

\bibitem[Berlin(2018)]{apollo17orthomosaic}
Berlin, T.~U.
\newblock Moon apollo 17 lroc nac landing site orthomosaic 50cm, 2018.
\newblock URL
  \url{https://astrogeology.usgs.gov/search/map/moon_apollo_17_lroc_nac_landing_site_orthomosaic_50cm}.
\newblock Accessed: 2024-10-02.

\bibitem[Cai et~al.(2022)Cai, Li, Mao, and Tei]{cai2022value}
Cai, J., Li, J., Mao, Z., and Tei, K.
\newblock Value iteration residual network with self-attention.
\newblock In \emph{International Conference on Intelligent Systems Design and
  Applications}, pp.\  16--24. Springer, 2022.

\bibitem[Cai et~al.(2023)Cai, Li, Zhang, and Tei]{cai2023value}
Cai, J., Li, J., Zhang, M., and Tei, K.
\newblock Value iteration networks with gated summarization module.
\newblock \emph{IEEE Access}, 2023.

\bibitem[Chen et~al.(2020)Chen, Dai, Liu, Chen, Yuan, and Liu]{chen2020dynamic}
Chen, Y., Dai, X., Liu, M., Chen, D., Yuan, L., and Liu, Z.
\newblock Dynamic convolution: Attention over convolution kernels.
\newblock In \emph{Proceedings of the IEEE/CVF conference on computer vision
  and pattern recognition}, pp.\  11030--11039, 2020.

\bibitem[Chen et~al.(2017)Chen, Everett, Liu, and How]{chen2017socially}
Chen, Y.~F., Everett, M., Liu, M., and How, J.~P.
\newblock Socially aware motion planning with deep reinforcement learning.
\newblock In \emph{2017 IEEE/RSJ International Conference on Intelligent Robots
  and Systems (IROS)}, pp.\  1343--1350. IEEE, 2017.

\bibitem[Eysenbach et~al.(2019)Eysenbach, Salakhutdinov, and
  Levine]{eysenbach2019search}
Eysenbach, B., Salakhutdinov, R.~R., and Levine, S.
\newblock Search on the replay buffer: Bridging planning and reinforcement
  learning.
\newblock \emph{Advances in neural information processing systems}, 32, 2019.

\bibitem[Foundation(2022)]{gymnasium}
Foundation, F.
\newblock Gymnasium: A toolkit for developing and comparing reinforcement
  learning environments.
\newblock \url{https://github.com/Farama-Foundation/Gymnasium}, 2022.

\bibitem[Fu et~al.(2020)Fu, Kumar, Nachum, Tucker, and Levine]{fu2020d4rl}
Fu, J., Kumar, A., Nachum, O., Tucker, G., and Levine, S.
\newblock D4rl: Datasets for deep data-driven reinforcement learning, 2020.

\bibitem[Hafner et~al.(2020)Hafner, Lillicrap, Ba, and
  Norouzi]{hafner2020dream}
Hafner, D., Lillicrap, T.~P., Ba, J., and Norouzi, M.
\newblock Dream to control: Learning behaviors by latent imagination.
\newblock \emph{Proceedings of the 8th International Conference on Learning
  Representations}, 2020.
\newblock URL \url{https://openreview.net/forum?id=S1lOTC4tDS}.

\bibitem[Hafner et~al.(2021)Hafner, Lillicrap, Norouzi, and
  Ba]{hafner2021mastering}
Hafner, D., Lillicrap, T.~P., Norouzi, M., and Ba, J.
\newblock Mastering atari with discrete world models.
\newblock \emph{Proceedings of the 9th International Conference on Learning
  Representations}, 2021.
\newblock URL \url{https://openreview.net/forum?id=0oabwyZbOu}.

\bibitem[Hafner et~al.(2023)Hafner, Pasukonis, Ba, and
  Lillicrap]{hafner2023mastering}
Hafner, D., Pasukonis, J., Ba, J., and Lillicrap, T.~P.
\newblock \emph{Mastering Diverse Domains through World Models}.
\newblock ar{X}iv, 2023.
\newblock URL \url{https://arxiv.org/abs/2301.04104}.

\bibitem[Hart et~al.(1968)Hart, Nilsson, and Raphael]{hart1968formal}
Hart, P.~E., Nilsson, N.~J., and Raphael, B.
\newblock A formal basis for the heuristic determination of minimum cost paths.
\newblock \emph{{IEEE} Transactions on Systems Science and Cybernetics},
  4\penalty0 (2):\penalty0 100--107, 1968.
\newblock \doi{10.1109/TSSC.1968.300136}.

\bibitem[He et~al.(2016)He, Zhang, Ren, and Sun]{he2016deep}
He, K., Zhang, X., Ren, S., and Sun, J.
\newblock Deep residual learning for image recognition.
\newblock In \emph{Proceedings of the IEEE conference on computer vision and
  pattern recognition}, pp.\  770--778, 2016.

\bibitem[Hochreiter(1991)]{Hochreiter:91}
Hochreiter, S.
\newblock {Untersuchungen zu dynamischen neuronalen Netzen. Diploma thesis,
  Institut f\"{u}r Informatik, Lehrstuhl Prof. Brauer, Technische
  Universit\"{a}t M\"{u}nchen}, 1991.
\newblock Advisor: J. Schmidhuber.

\bibitem[Hochreiter \& Schmidhuber(1997)Hochreiter and
  Schmidhuber]{Hochreiter:97lstm}
Hochreiter, S. and Schmidhuber, J.
\newblock {Long Short-Term Memory}.
\newblock \emph{Neural Computation}, 9\penalty0 (8):\penalty0 1735--1780, 1997.

\bibitem[Janner et~al.(2022)Janner, Du, Tenenbaum, and
  Levine]{janner2022planning}
Janner, M., Du, Y., Tenenbaum, J., and Levine, S.
\newblock Planning with diffusion for flexible behavior synthesis.
\newblock In \emph{International Conference on Machine Learning}, pp.\
  9902--9915. PMLR, 2022.

\bibitem[Jin et~al.(2021)Jin, Lan, Wang, and Yu]{jin2021value}
Jin, X., Lan, W., Wang, T., and Yu, P.
\newblock Value iteration networks with double estimator for planetary rover
  path planning.
\newblock \emph{Sensors}, 21\penalty0 (24):\penalty0 8418, 2021.

\bibitem[Lample \& Chaplot(2017)Lample and Chaplot]{lample2017playing}
Lample, G. and Chaplot, D.~S.
\newblock Playing fps games with deep reinforcement learning.
\newblock In \emph{Proceedings of the AAAI Conference on Artificial
  Intelligence}, volume~31, 2017.

\bibitem[Lee et~al.(2018)Lee, Parisotto, Chaplot, Xing, and
  Salakhutdinov]{lee2018gated}
Lee, L., Parisotto, E., Chaplot, D.~S., Xing, E., and Salakhutdinov, R.
\newblock Gated path planning networks.
\newblock In \emph{International Conference on Machine Learning}, pp.\
  2947--2955. PMLR, 2018.

\bibitem[Li et~al.(2021)Li, Yang, Song, and Jiang]{li2021dynamic}
Li, W., Yang, B., Song, G., and Jiang, X.
\newblock Dynamic value iteration networks for the planning of rapidly changing
  uav swarms.
\newblock \emph{Frontiers of Information Technology \& Electronic Engineering},
  22\penalty0 (5):\penalty0 687--696, 2021.

\bibitem[Liu et~al.(2018)Liu, Lehman, Molino, Petroski~Such, Frank, Sergeev,
  and Yosinski]{liu2018intriguing}
Liu, R., Lehman, J., Molino, P., Petroski~Such, F., Frank, E., Sergeev, A., and
  Yosinski, J.
\newblock An intriguing failing of convolutional neural networks and the
  coordconv solution.
\newblock \emph{Advances in neural information processing systems}, 31, 2018.

\bibitem[Mishra et~al.(2023)Mishra, Xue, Chen, and Xu]{mishra2023generative}
Mishra, U.~A., Xue, S., Chen, Y., and Xu, D.
\newblock Generative skill chaining: Long-horizon skill planning with diffusion
  models.
\newblock In \emph{Conference on Robot Learning}, pp.\  2905--2925. PMLR, 2023.

\bibitem[Niu et~al.(2018)Niu, Chen, Guo, Targonski, Smith, and
  Kova{\v{c}}evi{\'c}]{niu2018generalized}
Niu, S., Chen, S., Guo, H., Targonski, C., Smith, M., and Kova{\v{c}}evi{\'c},
  J.
\newblock Generalized value iteration networks: Life beyond lattices.
\newblock In \emph{Proceedings of the AAAI Conference on Artificial
  Intelligence}, volume~32, 2018.

\bibitem[Pflueger et~al.(2019)Pflueger, Agha, and Sukhatme]{pflueger2019rover}
Pflueger, M., Agha, A., and Sukhatme, G.~S.
\newblock Rover-irl: Inverse reinforcement learning with soft value iteration
  networks for planetary rover path planning.
\newblock \emph{IEEE Robotics and Automation Letters}, 4\penalty0 (2):\penalty0
  1387--1394, 2019.

\bibitem[Puterman(2014)]{puterman2014markov}
Puterman, M.~L.
\newblock \emph{Markov decision processes: discrete stochastic dynamic
  programming}.
\newblock John Wiley \& Sons, 2014.

\bibitem[Ross et~al.(2011)Ross, Gordon, and Bagnell]{ross2011reduction}
Ross, S., Gordon, G., and Bagnell, D.
\newblock A reduction of imitation learning and structured prediction to
  no-regret online learning.
\newblock In \emph{Proceedings of the fourteenth international conference on
  artificial intelligence and statistics}, pp.\  627--635. JMLR Workshop and
  Conference Proceedings, 2011.

\bibitem[Schaal(1996)]{schaal1996learning}
Schaal, S.
\newblock Learning from demonstration.
\newblock \emph{Advances in neural information processing systems}, 9, 1996.

\bibitem[Schleich et~al.(2019)Schleich, Klamt, and Behnke]{schleich2019value}
Schleich, D., Klamt, T., and Behnke, S.
\newblock Value iteration networks on multiple levels of abstraction.
\newblock \emph{arXiv preprint arXiv:1905.11068}, 2019.

\bibitem[Schmidhuber(1990{\natexlab{a}})]{Schmidhuber:90diffgenau}
Schmidhuber, J.
\newblock Making the world differentiable: On using fully recurrent
  self-supervised neural networks for dynamic reinforcement learning and
  planning in non-stationary environments.
\newblock Technical Report FKI-126-90 (revised), Institut f\"{u}r Informatik,
  Technische Universit\"{a}t M\"{u}nchen, November 1990{\natexlab{a}}.
\newblock (Revised and extended version of an earlier report from February.).

\bibitem[Schmidhuber(1990{\natexlab{b}})]{Schmidhuber:90sandiego}
Schmidhuber, J.
\newblock An on-line algorithm for dynamic reinforcement learning and planning
  in reactive environments.
\newblock In \emph{Proc. IEEE/INNS International Joint Conference on Neural
  Networks, San Diego}, volume~2, pp.\  253--258, 1990{\natexlab{b}}.

\bibitem[Schrittwieser et~al.(2020)Schrittwieser, Antonoglou, Hubert, Simonyan,
  Sifre, Schmitt, Guez, Lockhart, Hassabis, Graepel,
  et~al.]{schrittwieser2020mastering}
Schrittwieser, J., Antonoglou, I., Hubert, T., Simonyan, K., Sifre, L.,
  Schmitt, S., Guez, A., Lockhart, E., Hassabis, D., Graepel, T., et~al.
\newblock Mastering atari, go, chess and shogi by planning with a learned
  model.
\newblock \emph{Nature}, 588\penalty0 (7839):\penalty0 604--609, 2020.

\bibitem[Shen et~al.(2020)Shen, Zhuo, Xu, Zhong, and Pan]{shen2020transfer}
Shen, J., Zhuo, H.~H., Xu, J., Zhong, B., and Pan, S.
\newblock Transfer value iteration networks.
\newblock In \emph{Proceedings of the AAAI Conference on Artificial
  Intelligence}, volume~34, pp.\  5676--5683, 2020.

\bibitem[Silver et~al.(2017)Silver, van Hasselt, Hessel, Schaul, Guez, Harley,
  Dulac{-}Arnold, Reichert, Rabinowitz, Barreto, and
  Degris]{silver2017predictron}
Silver, D., van Hasselt, H., Hessel, M., Schaul, T., Guez, A., Harley, T.,
  Dulac{-}Arnold, G., Reichert, D.~P., Rabinowitz, N.~C., Barreto, A., and
  Degris, T.
\newblock The predictron: End-to-end learning and planning.
\newblock \emph{Proceedings of the 34th International Conference on Machine
  Learning}, 70:\penalty0 3191--3199, 2017.
\newblock URL \url{http://proceedings.mlr.press/v70/silver17a/silver17a.pdf}.

\bibitem[Song et~al.(2023)Song, Wu, Washington, Sadler, Chao, and
  Su]{song2023llm}
Song, C.~H., Wu, J., Washington, C., Sadler, B.~M., Chao, W.-L., and Su, Y.
\newblock Llm-planner: Few-shot grounded planning for embodied agents with
  large language models.
\newblock In \emph{Proceedings of the IEEE/CVF International Conference on
  Computer Vision}, pp.\  2998--3009, 2023.

\bibitem[Srivastava et~al.(2015{\natexlab{a}})Srivastava, Greff, and
  Schmidhuber]{highway2015}
Srivastava, R.~K., Greff, K., and Schmidhuber, J.
\newblock Highway networks.
\newblock \emph{arXiv preprint arXiv:1505.00387}, 2015{\natexlab{a}}.

\bibitem[Srivastava et~al.(2015{\natexlab{b}})Srivastava, Greff, and
  Schmidhuber]{srivastava2015training}
Srivastava, R.~K., Greff, K., and Schmidhuber, J.
\newblock Training very deep networks.
\newblock \emph{Advances in neural information processing systems}, 28,
  2015{\natexlab{b}}.

\bibitem[Sutton(1991)]{sutton1991dyna}
Sutton, R.~S.
\newblock Dyna, an integrated architecture for learning, planning, and
  reacting.
\newblock \emph{{ACM} {SIGART} Bulletin}, 2\penalty0 (4):\penalty0 160--163,
  1991.
\newblock \doi{10.1145/122344.122377}.

\bibitem[Sutton(2019)]{sutton2019bitter}
Sutton, R.~S.
\newblock \emph{The Bitter Lesson}.
\newblock 2019.
\newblock URL \url{http://incompleteideas.net/IncIdeas/BitterLesson.html}.

\bibitem[Tamar et~al.(2016)Tamar, Levine, Abbeel, Wu, and
  Thomas]{tamar2016value}
Tamar, A., Levine, S., Abbeel, P., Wu, Y., and Thomas, G.
\newblock Value iteration networks.
\newblock \emph{Advances in Neural Information Processing Systems},
  29:\penalty0 2146--2154, 2016.

\bibitem[Wang et~al.(2024{\natexlab{a}})Wang, Li, Faccio, Wu, and
  Schmidhuber]{wang2024highway}
Wang, Y., Li, W., Faccio, F., Wu, Q., and Schmidhuber, J.
\newblock Highway value iteration networks.
\newblock \emph{Proceedings of the 41st International Conference on Machine
  Learning}, 2024{\natexlab{a}}.
\newblock URL \url{https://arxiv.org/abs/2406.03485}.

\bibitem[Wang et~al.(2024{\natexlab{b}})Wang, Liu, Strupl, Faccio, Wu, Tan, and
  Schmidhuber]{wang2023highway}
Wang, Y., Liu, H., Strupl, M., Faccio, F., Wu, Q., Tan, X., and Schmidhuber, J.
\newblock \emph{Highway Reinforcement Learning}.
\newblock ar{X}iv, 2024{\natexlab{b}}.
\newblock URL \url{https://arxiv.org/abs/2405.18289}.

\bibitem[W{\"o}hlke et~al.(2021)W{\"o}hlke, Schmitt, and van
  Hoof]{wohlke2021hierarchies}
W{\"o}hlke, J., Schmitt, F., and van Hoof, H.
\newblock Hierarchies of planning and reinforcement learning for robot
  navigation.
\newblock In \emph{2021 IEEE International Conference on Robotics and
  Automation (ICRA)}, pp.\  10682--10688. IEEE, 2021.

\bibitem[Wydmuch et~al.(2019)Wydmuch, Kempka, and
  Ja\'skowski]{Wydmuch2019ViZDoom}
Wydmuch, M., Kempka, M., and Ja\'skowski, W.
\newblock {ViZDoom} {C}ompetitions: {P}laying {D}oom from {P}ixels.
\newblock \emph{IEEE Transactions on Games}, 11\penalty0 (3):\penalty0
  248--259, 2019.
\newblock \doi{10.1109/TG.2018.2877047}.

\bibitem[Zhao et~al.(2023)Zhao, Xu, and Wong]{zhao2023scaling}
Zhao, L., Xu, H., and Wong, L.~L.
\newblock Scaling up and stabilizing differentiable planning with implicit
  differentiation.
\newblock In \emph{The Eleventh International Conference on Learning
  Representations}, 2023.
\newblock URL \url{https://openreview.net/forum?id=PYbe4MoHf32}.

\end{thebibliography}
